\documentclass{article}

\usepackage{tikz,lipsum,lmodern}
\usepackage[most]{tcolorbox}
\tcbuselibrary{breakable}
\usepackage{ulem}

\definecolor{myanalyzer}{RGB}{111,66,193} %
\definecolor{myplanner}{RGB}{0,123,255} %
\definecolor{myimplementer}{RGB}{253,126,20} %

\usepackage[12pt]{extsizes}
\usepackage[utf8]{inputenc}
\usepackage{amsmath}
\usepackage{amssymb}
\usepackage{amsthm}
\usepackage{amsfonts}
\usepackage[numbers]{ai4science}
\makeatletter
\@ifundefined{ver@natbib.sty}{%
  \@namedef{ver@natbib.sty}{}}{}
\makeatother
\usepackage{graphicx}
\usepackage{lipsum}
\usepackage{enumitem}
\usepackage{minted}
\usepackage{caption}
\usepackage{mdframed}
\usepackage{tabularx}
\usepackage{pifont}
\newcommand{\cmark}{\ding{51}}
\newcommand{\xmark}{\ding{55}}
\newcommand{\authorblock}[1]{%
  \begin{center}
  {\authorfontsize #1}
  \end{center}
}
\newcommand{\authorrow}[1]{#1\\[0.3em]}

\newcommand{\aumark}[1]{\textsuperscript{#1}}
\newcommand{\emark}{\raisebox{-0.5ex}{*}}

\newcommand{\authorfontsize}{\fontsize{11}{13}\selectfont}

\usepackage{url}
\usepackage{subfigure}
\usepackage[subfigure]{tocloft}
\usepackage{bibunits}
\usepackage[dvipsnames,table]{xcolor}
\usepackage{hyperref}
\usepackage{mathtools}
\usepackage{relsize}
\usepackage{multirow}
\usepackage{booktabs}
\usepackage{tabularx}
\usepackage{bbding}
\usepackage{listings}
\usepackage{xspace}
\usepackage[ruled,vlined]{algorithm2e}
\usepackage{multirow}
\usepackage[p,osf]{cochineal}
\usepackage[scale=.95,type1]{cabin}
\usepackage[zerostyle=c,scaled=.94]{newtxtt}
\usepackage[cal=boondoxo]{mathalfa}
\usepackage{microtype}
\usepackage{multirow}
\usepackage{adjustbox}
\usepackage{etoolbox}
\usepackage{lmodern}
\usepackage{cleveref}
\usepackage{titletoc}

\usepackage[font=small,labelfont=bf]{caption} 

\crefname{appendix}{supplementary information}{supplementary information}
\Crefname{appendix}{Supplementary Information}{Supplementary Information}

\crefname{suppsection}{supplementary information}{supplementary information}
\Crefname{suppsection}{Supplementary Information}{Supplementary Information}

\crefname{suppfig}{supplementary figure}{supplementary figures}
\Crefname{suppfig}{Supplementary Figure}{Supplementary Figures}

\crefname{supptab}{supplementary table}{supplementary tables}
\Crefname{supptab}{Supplementary Table}{Supplementary Tables}

\usepackage{bibunits}
\defaultbibliographystyle{unsrtnat}
\defaultbibliography{ref}

\usepackage{csquotes}

\usepackage{geometry}
\definecolor{myurlcolor}{HTML}{123463}
\definecolor{dc_color}{RGB}{230, 245, 244}
\definecolor{ds_color}{RGB}{195, 230, 227}
\definecolor{ms_color}{RGB}{150, 214, 209}
\hypersetup{
    colorlinks=true,
    linkcolor=black,
    filecolor=black,      
    urlcolor=black,
    citecolor=black
}

\hypersetup{
   breaklinks=true,   
   colorlinks=true,  
   linkcolor=Bittersweet,
   citecolor=Periwinkle,
   urlcolor=gray
}
\definecolor{red}{HTML}{ca0020}
\definecolor{lightred}{HTML}{f4a582}
\definecolor{lightblue}{HTML}{92c5de}
\definecolor{green}{HTML}{008837}
\definecolor{blue}{HTML}{2c7bb6}

\apptocmd{\thebibliography}{\raggedright}{}{}

\makeatletter
\patchcmd{\@maketitle}{\LARGE \@title}{\fontsize{30}{19.2}\selectfont\@title}{}{}
\makeatother
\usepackage{amsmath,amsfonts,bm}

\def\eqref#1{equation~\ref{#1}}

\DeclareMathAlphabet{\mathsfit}{\encodingdefault}{\sfdefault}{m}{sl}
\SetMathAlphabet{\mathsfit}{bold}{\encodingdefault}{\sfdefault}{bx}{n}

\newcommand{\namens}{SAGA}
\newcommand{\name}{\namens\xspace}

\begin{document}

\title{\textbf{Accelerating Scientific Discovery with Autonomous Goal-evolving Agents}}
\date{} 

\maketitle
\vspace{-6em}  

\authorblock{
  \authorrow{
  Yuanqi Du\aumark{1,\emark,${\dagger}$},
  Botao Yu\aumark{2,\emark},
  Tianyu Liu\aumark{3,\emark},
  Tony Shen\aumark{4,\emark},
  Junwu Chen\aumark{5,\emark},
  Jan G. Rittig\aumark{5,\emark}, Kunyang Sun\aumark{6,\emark},
  Yikun Zhang\aumark{7,8,\emark},
  }
  \authorrow{
  Aarti Krishnan\aumark{8,9,10,11},
  Yu Zhang\aumark{8,9,10},
  Daniel Rosen\aumark{8,12},
  Rosali Pirone\aumark{8},
  Zhangde Song\aumark{13},
  Bo Zhou\aumark{14},
  Yingze Wang\aumark{6},
  }
  \authorrow{
  Cassandra Masschelein\aumark{5},
  Haorui Wang\aumark{15},
  Haojun Jia\aumark{13},
  Chao Zhang\aumark{15},
  Hongyu Zhao\aumark{3},
  Martin Ester\aumark{4},
  Nir Hacohen\aumark{8,16},
  }
  \authorrow{
  Teresa Head-Gordon\aumark{6,${\dagger}$},
  Carla P.~Gomes\aumark{1,${\dagger}$},
  Huan Sun\aumark{2,${\dagger}$},
  Chenru Duan\aumark{13,$^{\dagger}$},
  Philippe Schwaller\aumark{5,${\dagger}$},
  Wengong Jin\aumark{7,8,${\dagger}$}
  }
}

\begingroup
\renewcommand{\thefootnote}{}
\footnotetext{\scriptsize
  $^{1}$Cornell University, Ithaca, NY, USA;
  $^{2}$The Ohio State University, Columbus, OH, USA;
  $^{3}$Yale University, New Haven, CT, USA;
  $^{4}$Simon Fraser University, Burnaby, BC, Canada;
  $^{5}$École Polytechnique Fédérale de Lausanne, Lausanne, Switzerland;
  $^{6}$University of California Berkeley, Berkeley, CA, USA;
  $^{7}$Northeastern University, Boston, MA, USA;
  $^{8}$Broad Institute of MIT and Harvard, Cambridge, MA, USA;
  $^{9}$Massachusetts Institute of Technology, Cambridge, MA, USA;
  $^{10}$Wyss Institute for Biologically Inspired Engineering, Harvard University, Boston, MA, USA;
  $^{11}$Whitehead Institute for Biomedical Research, Cambridge, MA, USA;
  $^{12}$Brigham and Women's Hospital and Dana-Farber Cancer Institute, Boston, MA, USA;
  $^{13}$Deep Principle, Hangzhou, Zhejiang, China;
  $^{14}$University of Illinois Chicago, Chicago, IL, USA;
  $^{15}$Georgia Institute of Technology, Atlanta, GA, USA;
  $^{16}$Massachusetts General Hospital, Krantz Family Center for Cancer Research, Boston, MA, USA;
  $^{\star}$These authors contribute equally
  $^{\dagger}$Correspondence to: yd392@cornell.edu, thg@berkeley.edu, gomes@cs.cornell.edu, sun.397@osu.edu, duanchenru@gmail.com, philippe.schwaller@epfl.ch, w.jin@northeastern.edu
}
\endgroup

\begin{bibunit}

\begin{abstract}
\noindent There has been unprecedented interest in developing agents that expand the boundary of scientific discovery, primarily by optimizing quantitative objective functions specified by scientists. However, for grand challenges in science, these objectives may only be imperfect proxies. We argue that automating objective function design is a central, yet unmet need for scientific discovery agents. In this work, we introduce the Scientific Autonomous Goal-evolving Agent (\name) to address this challenge.
\name employs a bi-level architecture in which an outer loop of LLM agents analyzes optimization outcomes, proposes new objectives, and converts them into computable scoring functions, while an inner loop performs solution optimization under the current objectives. This bi-level design enables systematic exploration of the space of objectives and their trade-offs, rather than treating them as fixed inputs. We demonstrate the framework through a wide range of design applications, including antibiotics, nanobodies, functional DNA sequences, inorganic materials, and chemical processes. Notably, our experimental validation identifies a structurally novel hit with promising potency and safety profiles for \textit{E. coli} in the antibiotic design task, and three \textit{de novo} PD-L1 binders in the nanobody design task. These results suggest that automating objective formulation can substantially improve the effectiveness of scientific discovery agents. Our code is available at \url{https://github.com/btyu/SAGA} under the MIT license.

\end{abstract}

\section{Introduction}

\begin{figure}[htbp]
    \centering
    \includegraphics[trim = 0 0 0 0, clip, width=0.9\linewidth]{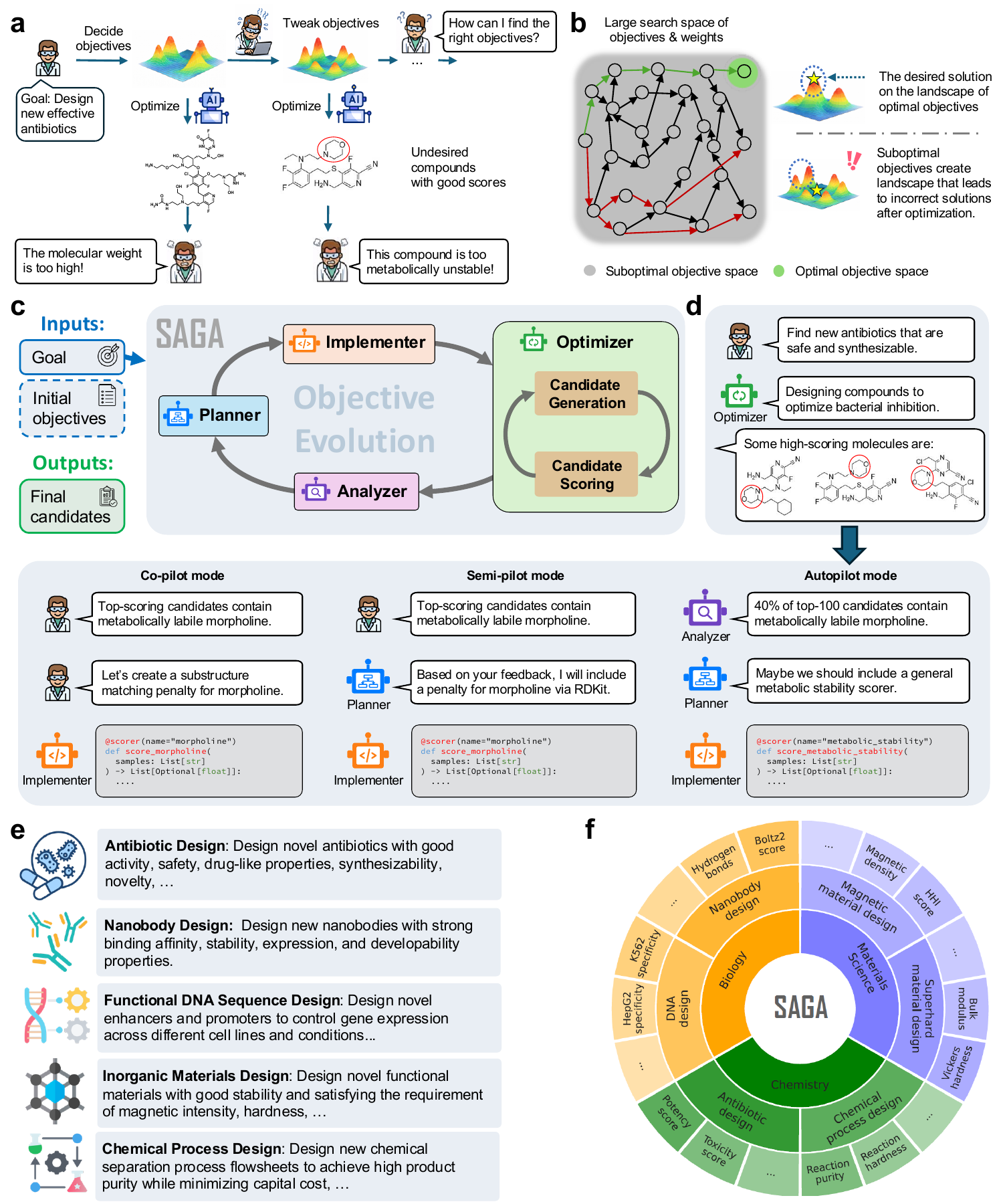}
    \hfill
    \caption{\textit{The \name framework and the examples of scientific applications.} (a) The current computational workflow with fixed objectives suffer from incomplete \textit{a priori} information about the design space, where optimization agents exploit the approximation error of objectives and propose undesirable hypotheses with good scores. (b) Finding ``optimal'' objectives is difficult due to the large search space of objectives and their relative weights. (c) We propose the \name framework to automatically discover objectives and candidate hypotheses through a bi-level procedure. (d) \name operates at three different levels of automation, allowing scientists to steer the objective discovery process in various ways. (e) We apply \name to scientific design tasks related to biology, chemistry, and materials science. (f) The \name framework is capable of implementing different objective across disciplines.}
    \label{fig:framework_combined}
\end{figure}

Scientific discovery has been driven by human ingenuity through iterations of hypothesis, experimentation, and observation, but is increasingly bottlenecked by the vast space of hypotheses to explore and the high cost of experimental validation~\citep{wang2023scientific}. 
Recent advances in artificial intelligence (AI) agents based on large language models (LLMs) offer promising approaches to address these bottlenecks and accelerate scientific discovery~\citep{m2024augmenting, huang2025biomni, boiko2023autonomous, yuksekgonul2025optimizing, novikov2025alphaevolve, swanson2025virtual, agarwalautodiscovery2025, gottweis2025towards, mitchener2025kosmos, lu2024ai,song2025evaluating}.
Leveraging massive pretrained knowledge and general capabilities for information collection and reasoning, these AI agents can efficiently navigate large hypothesis spaces and reduce experimental costs by automating key aspects of the research process.
For example, pipeline automation agents~\citep{m2024augmenting,huang2025biomni} streamline specialized data analysis workflows, reducing the manual effort required for routine experimental processes.
AI Scientist agents~\citep{Zheng2023,lu2024ai,yamada2025ai,swanson2025virtual,gottweis2025towards,mitchener2025kosmos,cascade} tackle the exploration challenge by autonomously generating and evaluating novel hypotheses (e.g., the relationship between a certain mutation and a certain disease) through integrated literature search, data analysis, and academic writing capabilities.

Our work embarks on a different and more ambitious goal in scientific discovery: building agents to discover new hypotheses to complex scientific design challenges, such as better therapeutic molecules and new functional materials. This problem is uniquely challenging due to the ``creativity'' and ``novelty'' required and the infinite combinatorial search space for hypotheses. Previous work has sought to address these challenges by developing optimization models that automatically find hypotheses maximizing a manually defined set of quantitative objectives, such as drug efficacy, protein expression, and material stability. These approaches, ranging from traditional generative models to more recent LLM-based methods, have demonstrated the ability to efficiently optimize against fixed objectives in domains including drug design~\citep{wangefficient,Cavanagh2024,loeffler2024reinvent}, algorithm discovery~\citep{novikov2025alphaevolve}, and materials design~\citep{mattergen,Liu2025}.

However, these optimization models operate under a critical assumption: that the right set of objectives is known upfront. In practice, this assumption seldom holds true \textit{a priori}. Just as scientific discovery requires iterations of hypothesis, experimentation, and observation, determining the appropriate objectives for a discovery task is itself an iterative search process. Scientists must constantly tweak objectives based on intermediate results, domain knowledge, and practical constraints that emerge during exploration (\Cref{fig:framework_combined}(a)). This iterative refinement is particularly crucial in experimental disciplines such as drug discovery, materials design, and protein engineering, where experimental success does not correlate well with computational proxies~\citep{van2025search,anstine2023generative}. Without this evolving process, the discovery suffers from incomplete knowledge about the design space: optimization algorithms exploit gaps between models and reality, producing solutions that maximize predicted scores while missing important practical considerations that experts would recognize. The search space for objectives and their relative weights is itself combinatorially large (\Cref{fig:framework_combined}(b)), making it extremely difficult to specify the right objectives from the outset. As a result, while existing optimization models can solve the low-level optimization problem efficiently, scientific discovery remains bottlenecked by the high-level objective search process that relies on manual trial-and-error.

In this work, we introduce \name as our first concrete step toward automating this iterative objective evolving process. \name is designed to navigate the combinatorial search space of objectives by integrating high-level objective planning in the outer loop with low-level optimization in the inner loop (\Cref{fig:framework_combined}(c)). The outer loop comprises four agentic modules: a planner that proposes new objectives based on the task goal and current progress, an implementer that converts proposed objectives into executable scoring functions, an optimizer that searches for candidate hypotheses maximizing the specified objectives, and an analyzer that examines the optimization results and identifies areas for improvement. Within the optimizer module, an inner loop employs any optimization strategies (e.g., genetic algorithms or reinforcement learning-based search) to iteratively evolve candidate hypotheses toward the current objectives. Importantly, \name is a flexible framework supporting different levels of human involvement. It offers three modes (\Cref{fig:framework_combined}(d)): (1) co-pilot mode, where scientists collaborate with both the planner and analyzer to reflect on results and determine new objectives; (2) semi-pilot mode, where scientists provide feedback only to the analyzer; and (3) autopilot mode, where both analysis and planning are fully automated. This design allows scientists to interact with \name in ways that best suit their expertise and preferences.

We demonstrate \name as a generalist scientific discovery agentic framework with success across multiple scientific domains, from chemistry and biology to materials science (\Cref{fig:framework_combined}(e)-(f)).
In antibiotic design, \name identifies a novel hit with experimentally validated antibacterial activity for \textit{E. coli}, no cytotoxicity in human cell lines, and a Tanimoto distance greater than 0.7 to all known antibiotics.
In nanobody design, experimental testing confirms three \textit{de novo} PD-L1 binders with $K_D$ ranging from 300nM to 400nM, and the composite scoring function autonomously evolved by \name significantly separates binders from non-binders (p = 0.03) where no single \textit{in silico} metric alone does.
In functional DNA sequence design, \name proposes high-quality cell-type-specific enhancers for the HepG2 cell line, with nearly 50\% improvement over the best baseline.
In inorganic materials design, \name designs permanent magnets with low supply chain risk and superhard materials for precision cutting, with properties validated by Density Functional Theory (DFT) calculations. 
Lastly, \name demonstrates success in automating the design of chemical process flowsheets from scratch.
In summary, these results highlight the broad applicability of \name in many disciplines and the value of adaptive objective design in scientific discovery agents.

\section{Results}

\subsection{\name for Antibiotic Design}
\begin{figure}
    \centering
    \includegraphics[ width=0.95\linewidth]{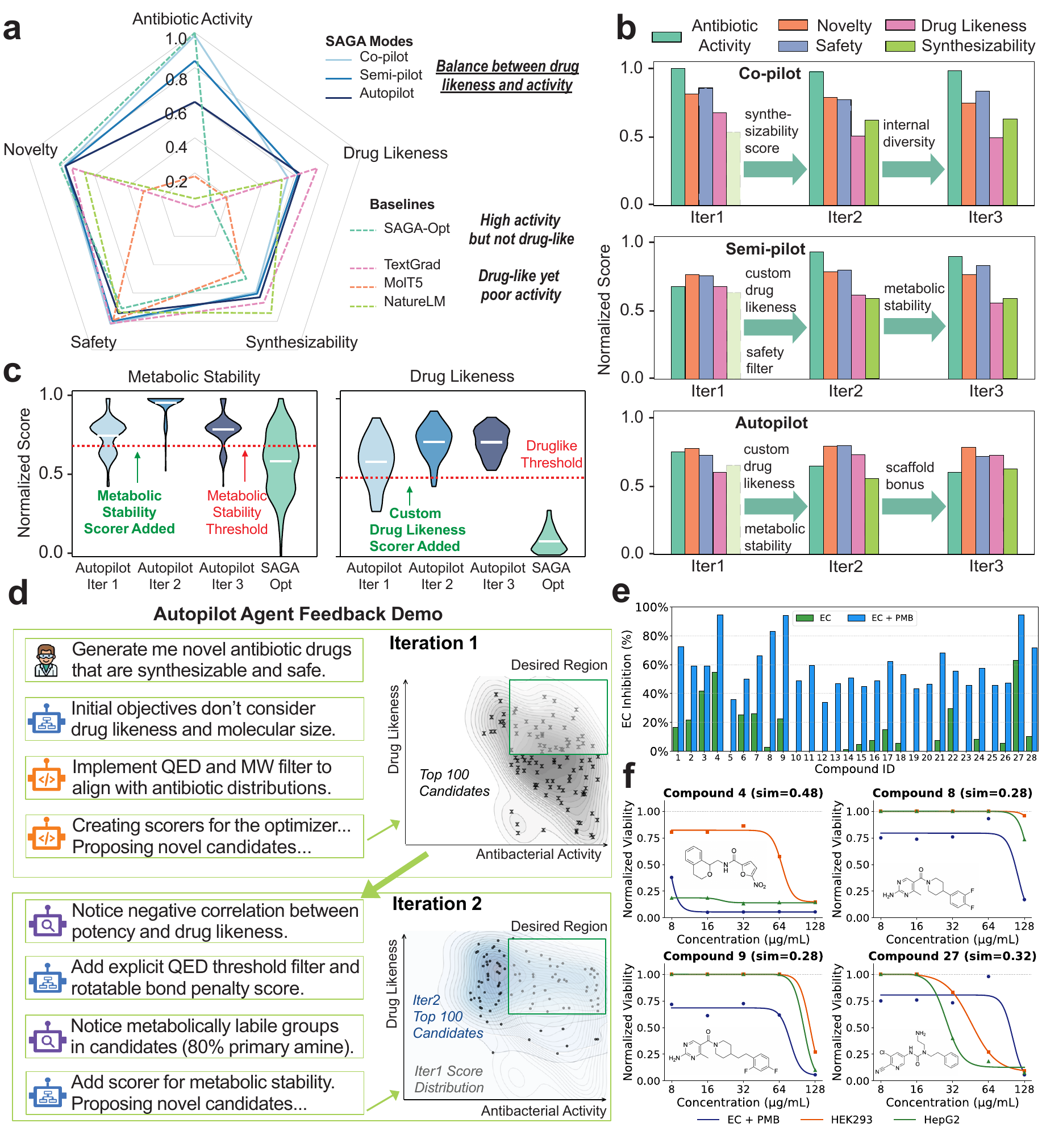}
    \vspace{-10pt}
    \caption{\textit{Results for antibiotic design.} 
    (a) Comparisons between \name and baselines. Candidates from all \name modes achieve the drug likeness and activity sweet spot, whereas baselines struggle to balance biological objectives.
    (b) Comparisons across \name iterations. Text annotations highlight the objective evolved in that iteration. The solid line means objectives address the evaluation metrics, and the dash line means the metric has not been addressed.
    (c) Distribution of metabolic stability score and drug likeness score for \name-Autopilot and \name-Opt. \name-Autopilot introduces the metabolic stability and custom drug likeness scorer that improve the overall chemical quality for the best molecules.
    (d) An example of the autopilot feedback loop. The analyzer identifies issues and the planner dynamically evolves objectives, shifting the distribution of the top 100 candidates toward the desired region of high activity and drug likeness.
    (e) Experimental validation of top 28 compounds designed by \name. Four compounds showed over 80\% growth inhibition for \textit{E. coli} at 128 $\mu$g/mL with polymyxin B (PMB). 
    (f) For these four compounds, we plot their antibiotic activity and cytotoxicity in human HEK293 and HepG2 cell lines at different concentrations. Lower viability means higher antibiotic activity and cytotoxicity.}
    \label{fig:drug_design_main}
\end{figure}

Antimicrobial resistance (AMR) is rapidly eroding our ability to treat bacterial infections such as Escherichia coli (\textit{E. coli}) and other critical priority pathogens identified by the World Health Organization (WHO) \citep{brown_antibacterial_2016, sati2025bppl2024}. However, designing novel antibiotics is notoriously difficult because optimization methods suffer from generating chemically unreasonable compounds that lack the necessary given objectives~\cite{van2025search}. %
To address this challenge, we demonstrate the ability of \name to design new antibiotics using \textit{E. coli} as a proof of concept. Rather than relying on a static scoring function that attempts to encode every rule upfront, \name begins with primary biological objectives to maximize potency and minimize toxicity, along with a constraint to avoid existing scaffolds. From this foundation, \name dynamically constructs a suite of auxiliary objectives that guide the generative process toward a realistic chemical space at all three levels of automation. This strategy enables \name to produce more valid candidates that pass the evaluation metrics provided by the scientists.

\textbf{\name discovers computationally selective and chemically reasonable candidates}. We run \name at all three levels of automation with the same prompt and primary biological objectives. \name then iterates at different levels of automation until the outer loop is complete. To evaluate the quality of proposed candidates, we select three biological evaluations: an antibacterial activity score, a novelty score, and a safety score defined as 1 - the toxicity score), as well as two chemical evaluations: drug likeness defined by the Quantitative Estimate of Drug Likeness (QED) score, and synthesizability defined by the Synthetical Accessibility (SA) score. These scores are further elaborated in \Cref{sup:expdetail_drug}.
As illustrated in \Cref{fig:drug_design_main}(a) and \Cref{supfig:pass_rate_drug}, \name achieves a more balanced overall score distribution and a significantly higher percentage of candidates passing all evaluations (detailed in \Cref{sup:expdetail_drug}) than molecular language models that take natural language instructions. Specifically, all other language model frameworks struggle to overcome the optimization difficulty of the antibacterial activity score alone, resulting in chemically valid but inactive molecules. In addition, the standalone Optimizer module (\name-Opt), which lacks the capacity to dynamically evolve objectives, over-optimizes the primary biological objectives. As a result of this exploitation of the activity score, its proposed candidates suffer from significantly lower average drug likeness scores (\Cref{fig:drug_design_main}(c)). In contrast, \name successfully balances the scores of both biological objectives and standard medicinal chemistry filters, discovering drug-like molecules with high predicted activity in all three modes of operation. In summary, \name’s objective evolution ensures that biological optimization does not come at the cost of chemical integrity, generating candidates that are both potent and practical.

\textbf{\name can effectively adjust its objectives to avoid optimization failures.}
The superior performance of \name comes from its ability to identify failure modes in the generated compounds and redirect its goal. As shown in \Cref{fig:drug_design_main}(b) and \Cref{subsec:antibiotic history}, \name co-pilot mode incorporates nuanced human feedback to address low synthesizability, but its semi-pilot mode strategically defers adding strict chemical constraints in early iterations to instead adjust weights to prioritize the antibacterial activity objective. In autopilot mode, the analyzer provides chemical insights that anticipate expert concerns. As shown in \Cref{fig:drug_design_main}(d) and \Cref{subsec:antibiotic history}, \name goes beyond individual molecular analysis and identifies population-level trends, such as ``negative correlation between antibacterial activity and drug likeness''. Furthermore, it performs granular structural analysis to pinpoint over-represented metabolically labile moieties in top-scoring molecules, like primary amines, phenols, and morpholines, which typically require manual systematic review to uncover. As shown in \Cref{sec:add_exp_drug}, the planner defines and implements new objective functions to assess metabolic stability, whose correctness was later confirmed by expert examination. \name-Autopilot then integrates the metabolic stability score and a custom drug likeness filter into the reward function and steers the generated population to the desired region of the physicochemical space, leading to a higher passing rate on our evaluation metrics (\Cref{fig:drug_design_main}(c) and (d)). Finally, although these objectives are evolved by \name in autopilot mode, we find that they can also improve the performance of traditional generative models.
As shown in \Cref{supfig:reinvent4_drug_results}, when paired with REINVENT4 \citep{loeffler2024reinvent}, the objectives proposed by \name enhance its optimization performance. Collectively, these examples demonstrate the practical utility of dynamic objective evolution in solving hard multi-objective optimization problems.

\textbf{Experimental validation confirms \name's ability to identify novel hits with promising potency and safety profiles.}
The first stage of real-world antibiotic discovery is hit discovery. Because it is unlikely that highly potent and nontoxic molecules can be found in one shot, the goal of this stage is to identify novel hits with promising physicochemical profiles suitable for further optimization. To demonstrate \name's hit discovery capability, we synthesized the top 28 molecules designed by \name and experimentally tested their antibacterial activity for \textit{E. coli}. 
We tested these compounds with or without polymyxin B nonapeptide (PMB), which disrupts the outer membrane and increases the permeability of bacterial cells. As shown in \Cref{fig:drug_design_main}(e), compound 4, 8, 9, and 27 show more than 80\% inhibition of bacterial growth at a concentration of 128$\mu$g/mL when combined with PMB.
Next, we tested their antibacterial activity and cytotoxicity in human HEK293 and HepG2 cell lines at multiple doses. \Cref{fig:drug_design_main}(f) shows the minimal inhibitory concentration (MIC) of compound 4 and the other three compounds is 16$\mu$g/mL and 128$\mu$g/mL, respectively. 
Among the four compounds, only compound 8 showed minimal cytotoxicity in both human cell lines at its MIC. Compound 8 is highly novel because its Tanimoto similarity to all known antibiotics is only 0.28, less than the 0.4 threshold commonly used in the antibiotic discovery community~\citep{wong_discovery_2024}.
We also did not find any publications reporting its antibacterial activity via SciFinder. Although the potency and permeability of compound 8 need to be further improved, these results demonstrate \name's ability to discover initial hits that satisfy multiple competing objectives.

\subsection{\name for Nanobody Design}
\begin{figure}
    \centering
    \includegraphics[width=0.95\linewidth]{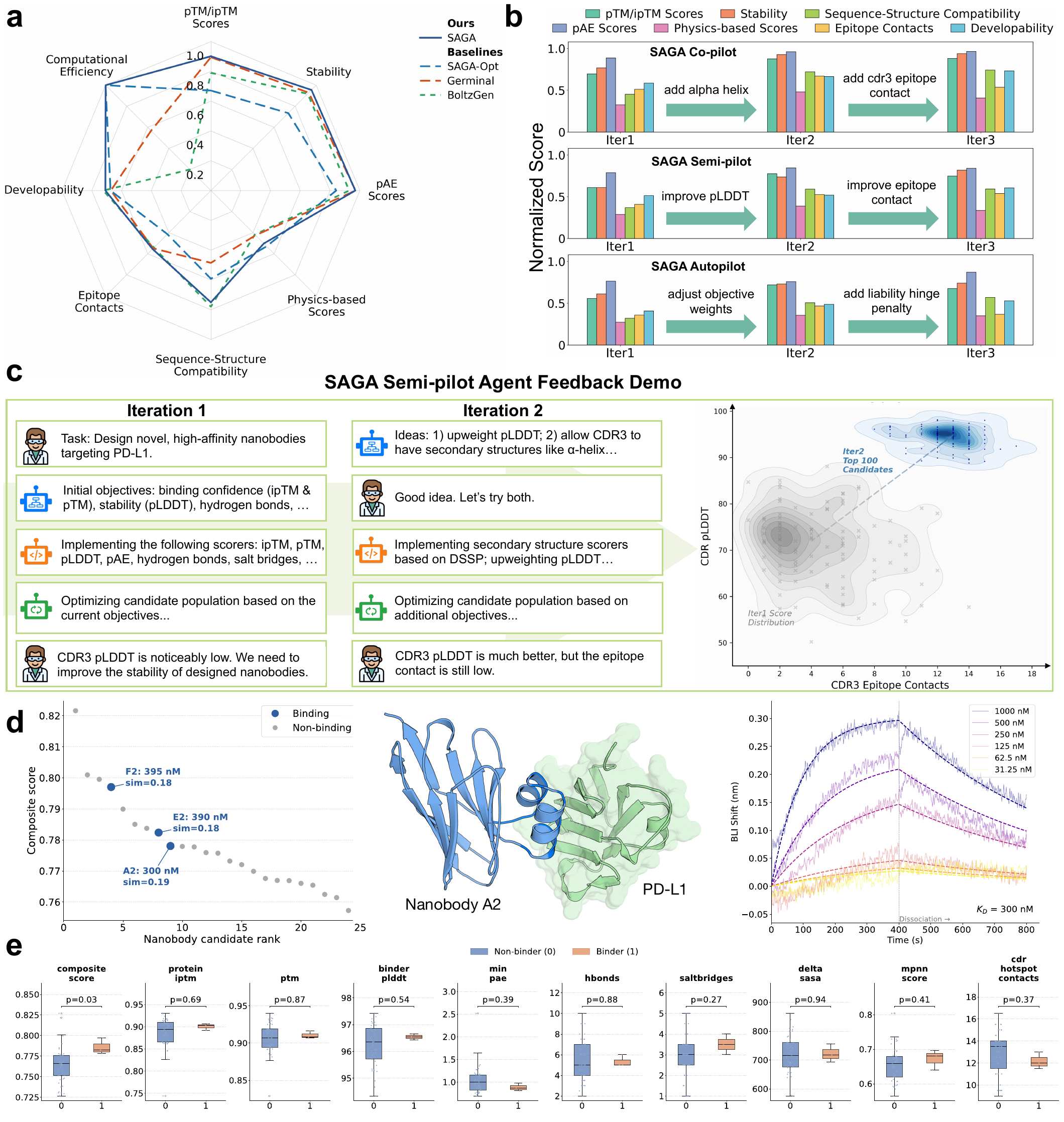}
    \caption{\textit{Results for nanobody binder design against PD-L1.}
    (a) Multi-objective optimization performance of \name, BoltzGen, and Germinal. \name exhibits a balanced profile across eight evaluation axes but with significantly higher computational efficiency.
    (b) Iterative performance improvement across three modes of human--agent collaboration. Bar plots show normalized metric scores over successive optimization iterations for the co-pilot, semi-pilot, and autopilot modes. Text annotations highlight representative objective updates introduced at each iteration.
    (c) Demonstration of the \name Semi-pilot feedback loop. Human experts provide high-level design goals and critiques, which \name translates into concrete objectives such as alpha-helix objectives, structural weight refinement, and epitope contact objectives.
    (d) Experimental validation shows three nanobodies designed by SAGA bind PD-L1 with $K_D$ ranging from 300 to 400nM. The structure of the best nanobody predicted by AlphaFold3 and its BLI traces are shown on the right.
    (e) Univariate analysis of all designed candidates reveals that the composite score developed by \name achieves much stronger binder ranking performance than existing metrics.}
    \label{fig:nanobody_design_main}
\end{figure}

Computational protein design has achieved remarkable success in recent years, driven by deep learning models such as RFDiffusion~\cite{watson2023novo}, ProteinMPNN~\cite{dauparas2022robust}, and BindCraft~\cite{pacesa2025one}. A key determinant of binder design success is the choice of computational objectives used to score or optimize designed candidates. The field has curated a growing list of \textit{in silico} metrics based on AlphaFold-based confidence scores~\cite{bennett2023improving} and physics-based energy functions, which together enable filtering of designs for fold stability, binding affinity, and interface quality. However, the ranking accuracy of these metrics varies substantially across targets and is particularly unreliable for antibodies and nanobodies~\cite{smorodina2026structural}, where flexible loops in their complementary-determining regions (CDR) confound structure prediction and energy estimation. With so many available metrics, the optimal combination and weighting are generally unknown \textit{a priori}. In practice, scientists must navigate a vast objective space largely by trial and error, refining their criteria in response to experimental failures. To address this challenge, we use \name to design \textit{de novo} nanobodies through dynamic objective evolution. Rather than relying on a static scoring function, \name begins with primary goals and iteratively proposes and weighs auxiliary objectives to avoid failure modes identified during optimization. \name is particularly suitable for nanobody design because the space of \textit{in silico} objectives is large, making it necessary to adapt optimization goals on the fly as failure cases emerge.

\textbf{\name achieves strong multi-objective optimization performance with substantially lower search cost.}
As proof of concept, we use \name to design nanobodies for Programmed Death Ligand~1 (PD-L1), a clinically relevant target in cancer immunotherapy.
We provide \name with a list of objectives commonly used by existing protein design methods, including binding confidence measured by ipTM, pTM, and min-pAE scores, stability measured by pLDDT scores, physics-based scores such as hydrogen bonds, salt bridges, and $\Delta$SASA, sequence-structure compatibility measured by ProteinMPNN score and recovery \citep{dauparas2022robust}, CDR-epitope contacts, and developability scores. We predict the structure of each sequence using AlphaFold3 \citep{abramson2024accurate} and Boltz2 \citep{passaro2025boltz} and average these metrics between the two predictors. More details of these scores are provided in \cref{sup:expdetail_nanobody}. We run \name for multiple iterations to evolve its design strategy and select candidates with the optimal combined scores in the last round of optimization. We compare these candidates with the PD-L1 nanobodies designed by BoltzGen~\citep{stark2025boltzgen} and Germinal~\citep{mille2025efficient}, two state-of-the-art nanobody design methods. For each method, we select the top 15 sequences ranked by the aforementioned metrics and report their performance. As shown in \Cref{fig:nanobody_design_main}(a), \name matches or exceeds both BoltzGen and Germinal in most metrics, resulting in a more balanced performance across binding confidence, physics-based scores, and developability. Detailed per-metric comparisons between the two structure predictors are available in \Cref{supfig:nanobody_metric_comparison}. Importantly, \name achieves this performance with 2-5 times fewer number of oracle calls than Germinal and BoltzGen. This suggests that \name can navigate the search space much more efficiently while maintaining strong multi-objective performance.

\textbf{\name can effectively diagnose and propose new objectives to address optimization bottlenecks.}
By comparing \name with the standalone optimizer module (\name-Opt), we find that dynamic objective evolution effectively resolves optimization bottlenecks (\Cref{fig:nanobody_design_main}(a)). In all settings, \name manage to identify the dominant bottlenecks from population-level trends and correct them by targeted objective changes (\Cref{fig:nanobody_design_main}(b)).In \name semi-pilot mode, for example, scientists notice that CDR3 pLDDT scores are particularly low after the first iteration and suggested that \name should improve the stability of the designed nanobodies. In response to this issue, \name proposed two solutions: increase the weight of pLDDT scores in the objective or allow the CDR3 region to have secondary structures like alpha-helices, which successfully increased AlphaFold and Boltz2 pLDDT scores (\Cref{fig:nanobody_design_main}(c)).
More details for the \name optimization and human-agent collaboration process are provided in \Cref{subsec:nanobody history}.

\textbf{Experimental validation confirms \name's ability to design effective nanobodies and propose better computational objectives.}
To demonstrate the utility of \name in real-world applications, we experimentally tested 24 nanobody candidates designed by \name using biolayer interferometry (BLI). We identified three true binders, with a lowest dissociation constant ($K_D$) of 300nM (\Cref{fig:nanobody_design_main}(d) and \Cref{supfig:nanobody_sp_structures}).
These binders are highly novel because their CDR3 sequences share less than 20\% similarity with all nanobodies designed by Germinal and BoltzGen, and all antibodies in the Structure Antibody Database (SAbDab)~\citep{dunbar2014sabdab} (\Cref{supfig:nanobody_sp_novelty}).
Notably, AlphaFold3 structural prediction shows nanobody A2 CDR3 adopts a novel alpha-helical structure rather than a canonical loop topology. The emergence of this non-canonical CDR3 topology suggests that \name is capable of exploring new regions of sequence-structure space beyond conventional design templates, potentially expanding the functional diversity of engineered nanobodies.
Lastly, we conduct univariate analyses for each \textit{in silico} metric to assess whether the composite objective evolved by \name better distinguishes binders from non-binders. As shown in \Cref{fig:nanobody_design_main}e, no single metric alone achieved statistical significance (all p > 0.05; Mann--Whitney U test), yet the composite scoring function autonomously constructed by \name yielded a p-value of 0.03, clearly separating binders from non-binders. This evaluation suggests that autonomous objective evolution is a promising paradigm not only for improving design efficiency but also for uncovering more predictive computational surrogates for nanobody design.

\subsection{\name for Functional DNA Sequence Design}
\begin{figure}[htpb]
    \centering
    \includegraphics[trim = 0 0 0 0, clip, width=0.93\linewidth]{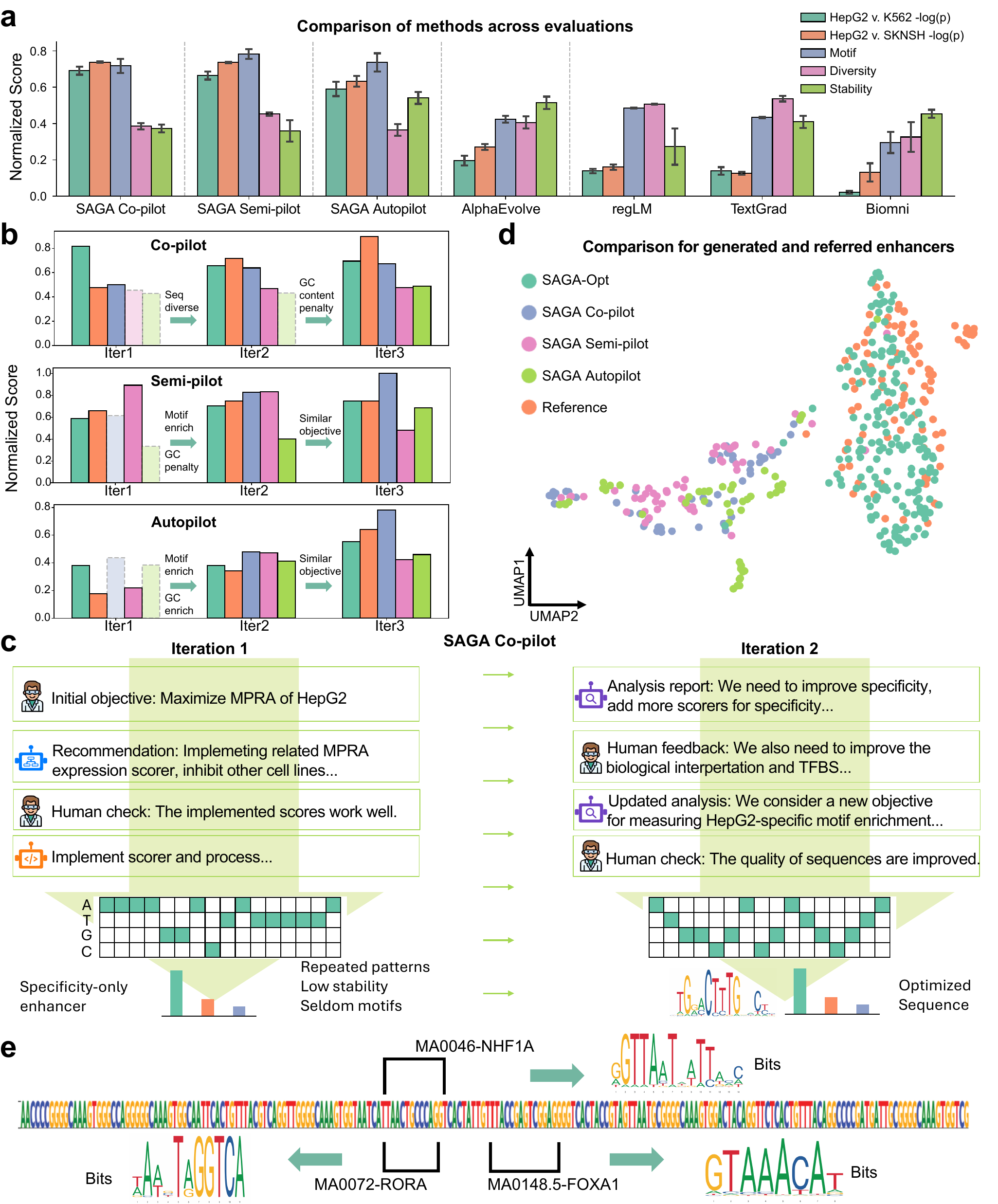}
    \caption{\textit{Results for functional sequence design.} (a) Comparisons between different levels of \name and selected baselines with evaluation metrics (both average score (higher is better) and scaled standard error are reported (lower is better)). Our selected task is to design HepG2 (a cell line of epithelial-like cells from liver) specific enhancers. (b) The comparisons of different iterations of two different levels with the same held-out metrics. Each iteration will create new objectives. The solid line means objectives address the evaluation metrics, and the dash line means the metric has not been addressed. (c) An example of \name to correct the issues in previous iterations. (d) UMAP visualization of enhancers designed by \name, SAGA-Opt, and from references. (e) HepG2-specific motif visualization.}
    \label{fig:dnaseq_design}
\end{figure}

Programmed, highly precise, and cell-type-specific enhancers and promoters are fundamental to the development of reporter constructs, genetic therapeutics, and gene replacement strategies \citep{gosai2024machine}. Such regulatory control is particularly important in HepG2, a human hepatocellular carcinoma cell line that retains key hepatic functions within a single cell type, including plasma protein synthesis and xenobiotic drug metabolism \citep{moyers2023characterization}. Although enhancers play a central role in establishing cell-type-specific gene expression programs \citep{andersson2020determinants}, their rational design remains challenging due to the vast combinatorial space of possible functional DNA sequences. This task can be naturally formulated as an optimization problem with predefined scoring functions, such as DNA expression level predictor \citep{liu2025unicorn}. However, optimizing solely against expression-based oracles often results in sequences that generalize poorly with respect to biologically relevant constraints, including transcription factor motif enrichment, sequence diversity, and DNA stability. To address these limitations, we apply \name to design novel cell-type-specific enhancers while iteratively refining the optimization objectives. Here, the \name framework is initialized using a predictor trained with cell-type-specific expression measurements obtained from Massively Parallel Reporter Assays (MPRA) \citep{agarwal2025massively} and subsequently performs optimization with respect to an initial set of objectives. Crucially, \name closes the design loop by systematically analyzing deficiencies in the designed sequences and adaptively modifying the objective functions to guide subsequent exploration. Through this iterative bi-level refinement process, \name converges towards a more comprehensive and biologically-grounded objective set, yielding optimized enhancer candidates that better satisfy multifaceted design requirements across multiple metric-level evaluations.

\textbf{\name effectively discovers biologically plausible functional DNA sequences}. We compare \name's discovery capabilities by benchmarking it against established domain-specific models and agents \citep{huang2025biomni,yuksekgonul2025optimizing,lal2024designing}. \Cref{fig:dnaseq_design}(a) reveals that \name in different modes surpass all baselines on metrics probing both statistical validity and biological function by 19.2-176.2\% across multiple independent runs. Under controlled conditions where all baselines target the same objectives, our system exhibits marked improvements in MPRA specificity (by at least 48.0\%), motif enrichment (by at least 47.9\% ), and sequence stability (by at least 1.7\%). To further demonstrate the superiority of dynamic objective evolution by \name, we utilize the Analyzer to examine the differences between enhancers produced by the Optimizer of \name with initial objectives only (\name-Opt) and \name (autopilot). \Cref{supfig:enhancer_analysis_agent} indicates that our designed enhancers exhibit obviously higher specificity and possess richer biological features compared to the former. These results suggest that \name effectively captures the complex interplay between statistical likelihood and biological constraints.

\textbf{\name proposes reasonable and helpful objectives to assist human scientists for enhancer design}. 
Now we examine \name's capabilities in collaborating with human scientist. \Cref{fig:dnaseq_design}(b) demonstrates the inclusion of human feedback via co-pilot and semi-pilot modes leads to marked improvements in biologically meaningful outcomes for DNA enhancer design.
For example, explicitly prioritizing transcription factor motif enrichment and sequence stability guided by expert input (\Cref{fig:dnaseq_design}(c) and \Cref{subsec:dna history}) can motivate \name to design objectives such as ``hepatocyte\_motif\_enrichment" and ``dna\_gc\_penalty'' to improve the enrichment of HepG2-specific motifs and ensure a reasonable GC content for sequence stability, and thus lead to the improvement of sequence quality. We also observe \name proposes objectives to penalize expression levels from the rest of two cell lines to design HepG2-specific enhancers. These objective-level contributions result in improved biological validity of the designed enhancer sequences.  %
Beyond human-\name collaboration, the autopilot mode can also fully automate the design of enhancers. \name autopilot achieves overall performance comparable to that obtained with human collaboration  with respect to HepG2 specificity and even demonstrates improvements in sequence diversity and stability. It also consistently outperforms other language model agent baselines across all evaluated metrics (Figures \ref{fig:dnaseq_design}(a) and (b)). As illustrated in \Cref{supfig:level3steps}, the objectives proposed by \name are highly consistent across independent replicates and optimization iterations, with the majority (88.8\%) corresponding to statistically driven objectives. Finally, we test if using the objectives proposed by \name can improve baseline's performance. According to \Cref{supfig:saga improvement}, we observe improvement of TextGrad after updating the objectives across several important evaluation metrics, ranging from expression level comparison and stability.

\textbf{\name uncovers both novel and highly-specific enhancer candidates validated with independent predictors}. %
Here we discuss the novelty and generalization of biological signals across modalities to showcase the quality of \name-designed enhancers. As shown in \Cref{fig:dnaseq_design}(d), we compare the distributions of enhancers optimized by \name and \name-Opt with those from a previously characterized experimental pool \citep{gosai2024machine}. Enhancers discovered by \name occupy a distinctly different distributional regime, reflecting systematic exploration beyond the initial training pool. Importantly, their strong performance on held-out evaluation metrics indicates that \name can be deployed in different operational modes to reliably generate additional high-quality enhancer candidates. Beyond quantitative performance, \name also recapitulates core biological principles of enhancer function. Specifically, it recovers multiple liver-specific transcription factor motifs \citep{sandelin2004jaspar} (\Cref{fig:dnaseq_design}(e)), supporting the biological plausibility and tissue specificity of the designed sequences. To further assess whether \name-designed enhancers generalize to biologically meaningful regulatory readouts not explicitly optimized during training, we evaluated them using multimodal regulatory predictions from independent predictors, including Cap Analysis Gene Expression sequencing (CAGE-seq) \citep{shiraki2003cap} and DNase I hypersensitive sites sequencing (DNase-seq) \citep{boyle2008high}. As shown in \Cref{supfig:enhancer_bio_analysis}(a), the designed enhancers exhibit strong HepG2-specific signals in both modalities. Moreover, they achieve higher predicted HepG2-specific expression levels than baseline methods (\Cref{supfig:enhancer_bio_analysis}(b)). These results demonstrate that \name effectively leverages information encoded in pre-trained sequence-to-function models, such as Enformer \citep{avsec2021effective}, to capture coherent multimodal regulatory programs rather than overfitting to a single assay. In cell types where enhancers are active, lineage-defining and signal-responsive transcription factors bind to the enhancer sequence and recruit chromatin remodeling complexes, leading to localized chromatin opening and elevated DNase I hypersensitivity. This accessible chromatin state further facilitates the recruitment of the transcriptional machinery, giving rise to enhancer-associated bidirectional transcription that is captured by CAGE-seq \citep{dasilva2024dna,guerrini2022cap,ampuja2017integrated}. Together, the coordinated elevation of DNase-seq and CAGE-seq signals provides complementary evidence of functional enhancer activity, reinforcing that \name successfully designs enhancers that recapitulate authentic, cell-type-specific regulatory programs rather than a single assay.

\subsection{\name for Inorganic Materials Design}
\begin{figure}[hptb]
    \centering
    \includegraphics[trim = 0 0 0 0, clip, width=1\linewidth]{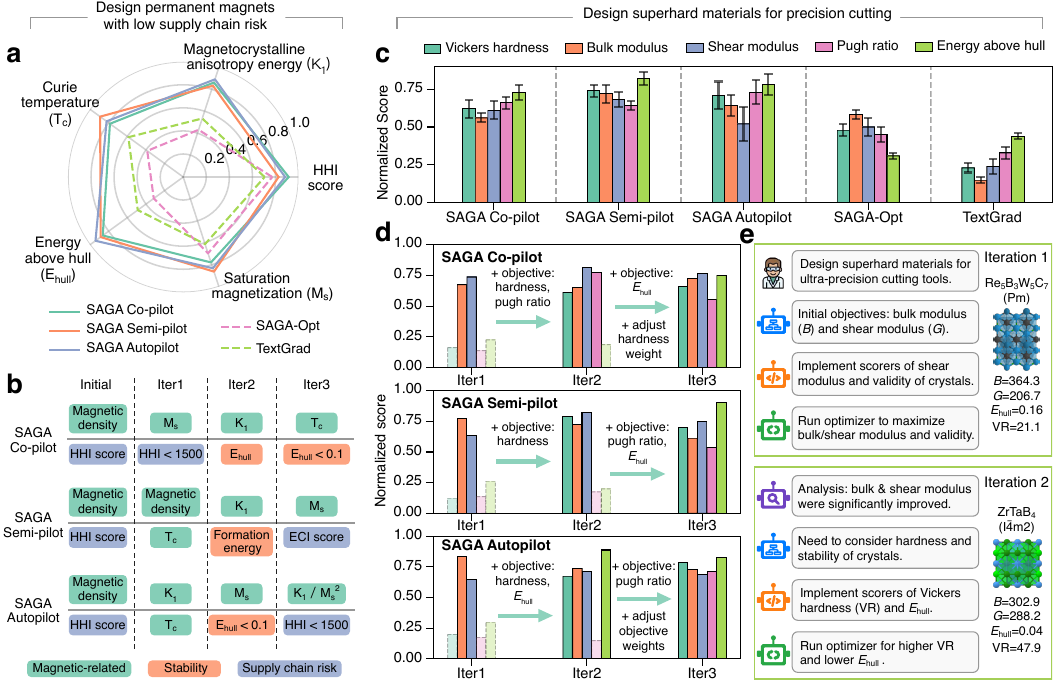}
    \caption{\textit{Results for inorganic materials design.}
    (a) Comparisons between \name and baselines. \name achieves a more balanced overall score distribution compared to the baselines.
    (b) Objectives proposed in early-stage iterations across three modes of \name.
    (c) Comparisons between different levels of \name and selected baselines on the design task of superhard materials for precision cutting. All evaluation metrics are normalized, with higher scores representing better performance.
    (d) Comparison of different \name modes over three iterations with the same held-out metrics. In each iteration, \name analyze the optimized crystal structures, propose new objectives, run property optimization, and select the best candidates across all current iterations. Text annotations highlight specific agent feedback on objective evolution that drives the improvement in metric scores across iterations. The solid line means objectives address the evaluation metrics, and the dash line means the metric has not been addressed.
    (e) An example of the autopilot feedback loop. \name identifies issues and dynamically evolves objectives, successfully proposed novel structures exhibiting high hardness, high elastic modulus, and thermodynamic stability.
    }
    \label{fig:crystal_design}
\end{figure}

The discovery of novel materials is critical for driving technological innovation across diverse fields, including catalysis, energy, electronics, and advanced manufacturing \citep{cascade,mattergen,deepmind_scaling,ml_alloy,ai4mat,matinvent,adsmt}. Most materials design tasks involve multiple objectives encompassing electronic, mechanical and physicochemical properties, as well as production costs \citep{mattergen, matinvent}. These design objectives are often intricately interrelated and may exhibit competitive or even conflicting trade-offs \citep{mat_review, matinvent, ml_tme}. Optimization with fixed objectives may overlook other important material properties or fail to refine optimization objectives based on deficiencies identified in proposed candidates. To address this challenge, we apply \name to design the novel materials desired for specific applications through iterative optimization with dynamic objectives. \name can guide LLMs to search for materials with desired properties, iteratively analyzing and adjusting optimization objectives, while automatically programming scoring functions to evaluate the new objectives and provide feedback. We study two design tasks to assess the \name's effectiveness.

\textbf{\name enables efficient design of magnet materials.}
First, we apply \name to design new permanent magnets with low supply chain risk, thereby avoiding the use of rare earth elements. Instead of pre-coding all design rules into static scoring functions, \name begins with two primary objectives for maximizing magnetic density and minimizing Herfindahl–Hirschman index (HHI) score, where a lower HHI score indicates lower supply chain risk and the absence of rare earth elements \citep{mattergen, hhi}. \name dynamically constructs auxiliary objectives that steer the generative space toward materials satisfying the full spectrum of desired properties. We run \name at all three automation levels, following the same prompts and initial objectives. As shown in \Cref{fig:crystal_design}(a), \name achieves a more balanced overall score distribution compared to the baselines. Specifically, the TextGrad method struggles to optimize magnetocrystalline anisotropy energy ($K_1$) without using rare earth elements, and the resulting crystals exhibit low thermodynamic stability. The standalone optimizer module (SAGA-Opt) lacks the ability to dynamically evolve the target, thus over-optimizing the saturation magnetization and HHI score without achieving reasonable thermodynamic stability and high Curie temperature. These results demonstrate that \name's dynamic objective evolution can achieve better overall performance in multi-objective tasks, even involving competing material properties. %

\textbf{\name enables the efficient design of superhard materials.}
Subsequently, we evaluate \name on the task of designing superhard materials for precision cutting and compare with an LLM-based optimization algorithm, TextGrad \citep{yuksekgonul2025optimizing}. This task involves more than three target material properties, whereas conventional methods that optimize with fixed targets may only achieve high scores on certain metrics but ignore other important properties of the designed materials. \name begins with two initial objectives: maximizing bulk modulus and shear modulus. As shown in \Cref{fig:crystal_design}(c), the crystal structures designed by three modes (co-pilot, semi-pilot, autopilot) achieve high scores on all metrics. Benefiting from iterative optimization and dynamic objective refinement, all \name modes successfully propose novel structures exhibiting high hardness, high elastic modulus, appropriate brittleness, and thermodynamic stability. In contrast, the TextGrad approach, which employs fixed optimization objectives, demonstrates moderate performance for energy above hull and Pugh ratio but achieves much lower scores for hardness and elastic modulus.  These results demonstrate that \name's iterative optimization and dynamic objective strategy are effective for complex multi-objective tasks. Furthermore, we analyze the crystal structures proposed by \name in the final iteration and confirm that the underlying patterns correlate with key factors for superhard material formation reported in experimental studies. More than 90 \% of the proposed crystals contain light elements such as boron, carbon, nitrogen, and oxygen, aligning with experimental findings that light elements are essential for superhard materials because their small atomic radii enable short, directional covalent bonds with high electron density \citep{exp_hardmat1, exp_hardmat2}. In addition, over 75\% of the proposed crystals are transition metal carbides, nitrides, and borides. Correspondingly, experimental studies have demonstrated that the combination of light elements (boron, carbon, nitrogen) with electron-rich transition metals can form dense covalent networks and enhance material hardness \citep{exp_hardmat1, exp_hardmat2}.

\textbf{\name proposes reasonable and important objectives aligned with materials scientists.}
As shown in \Cref{fig:crystal_design}(b), in the permanent magnet design task with low supply chain risk requirements, all three modes of \name are capable of proposing critical property objectives in the initial iterations, spanning magnetic-related properties, thermodynamic stability, and supply chain risk \citep{magnet1, magnet2}. Once the analysis agent determines that the objectives in the current iteration have been sufficiently optimized, the planning agent will design new objectives that are important to the design task in the next iteration. Moreover, during the iterative process, \name imposes reasonable thresholds on certain objectives, such as HHI score below 1500 and $E_{hull}$ below 0.1 eV/atom \citep{mattergen, hhi}, thereby preventing over-optimization of any single objective at the expense of overall performance. Furthermore, \name can propose new objectives that may supersede previous ones for specific material requirements. The semi-pilot mode of \name introduces the Elemental Criticality Index (ECI) in the third iteration, a novel objective that directly penalizes crystal structures containing rare-earth elements and may offer advantages over the originally specified HHI score objective. Additionally, \name is capable of proposing composite targets that combine two material properties, allowing the simultaneous optimization of two competing objectives. The Autopilot mode of \name introduces a new objective $K_1 / M_s^2$ in the third iteration, which accounts for the trade-off between the magnetocrystalline anisotropy energy ($K_1$) and saturation magnetization ($M_s$), as well as the magnetic hardness parameter \citep{magnet2, magnet3, magnet4}.

When designing superhard materials, all three modes of \name prioritized Vickers hardness, elastic modulus, and Pugh ratio \citep{superhard1, superhard2}, leading to substantial enhancement of mechanical properties in the proposed crystalline materials (\Cref{fig:crystal_design}(d)). In particular, the autopilot mode achieves comparable performance to co-pilot and semi-pilot across all five metrics, underscoring its remarkable planning and automation capabilities. \Cref{fig:crystal_design}(e) shows that the autopilot mode can correctly understand the design goal and analyze the properties of the proposed materials, subsequently proposing appropriate and highly relevant new objectives (e.g. Vickers hardness, Pugh ratio, and energy above hull) targeting mechanical performance and stability. For newly proposed objectives, \name implements scoring functions through web search and coding agent, leveraging publicly available pretrained models or empirical methods. Upon analyzing the designed structures and determining that a particular objective has been sufficiently optimized, \name automatically adjusts the optimization weight for that objective. Specifically, \name employs scaling or truncation of material property values to prevent over-optimization of individual objectives while neglecting others. In summary, these results demonstrate that \name enables materials design with different levels of human intervention through dynamic objective evolution.

\subsection{\name for Chemical Process Design}

Finally, we consider the use of \name for chemical process engineering applications, which is of high practical relevance within the chemical industry. While chemical process engineering has historically developed various heuristics and optimization-based approaches for the design of process flowsheets over the last decades,~\citep{biegler1997systematic, turton2008analysis}, more recently generative models combined with Reinforcement Learning (RL) have been investigated as a promising approach to automate chemical process design, with exemplary applications in reaction synthesis, separation, and extraction processes~\citep{gao2024deep, gottl2025deep, stops2023flowsheet}. However, the focus on single design objectives predefined by human experts~\citep{gao2024deep} can result in process flowsheets that lack characteristics of practical relevance and thus require iterative refinement in subsequent manual steps. It has also proven challenging to the LLM domain~\citep{rupprecht2025multi, du2025potential}, and only a few recent studies have utilized LLMs to optimize parameters for given chemical processes, e.g., in~\citep{zeng2025llm}. 
Here \name is used to advance automation of the chemical process design loop for separation of mixtures by proposing objectives that lead to more practical designs.

\begin{figure}[hptb]
    \centering
    \includegraphics[trim = 0 0 0 0, clip, width=1\linewidth]{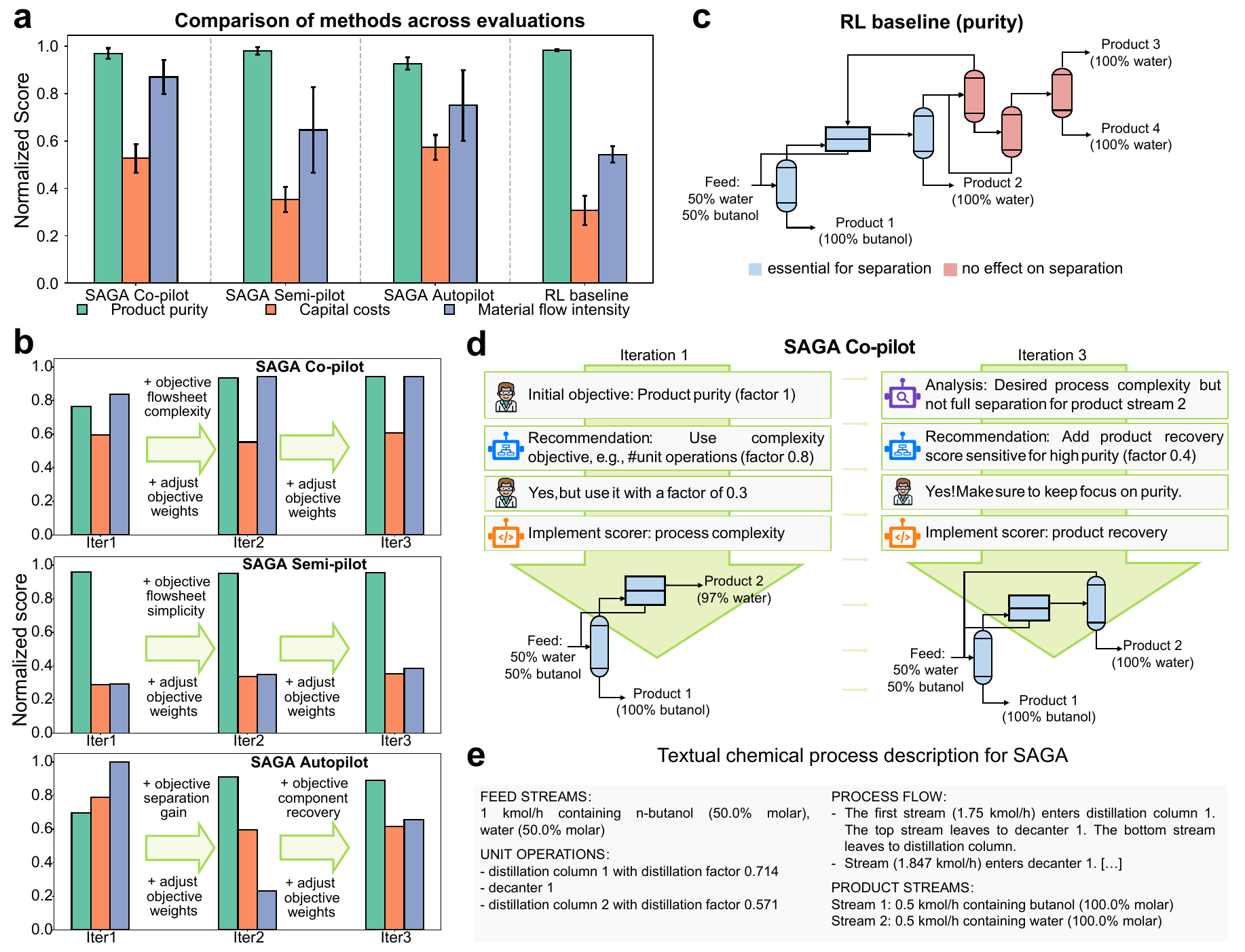}
    \caption{\textit{Results for chemical separation process design.} (a) Comparisons between different levels of \name and baseline RL (trained to maximize product purity) with evaluation metrics (higher is better). Our selected task is to design process flowsheets for separation of an azeotropic butanol/water mixture with varying feed compositions (between 2\%/98\%). (b) The comparisons of different iterations of the three \name levels ran for three iterations with the same evaluation metrics. In each iteration, \name will analyze the processes, create new objectives, run the RL optimization, and select the best candidates across all current iterations. (c) Exemplary designed process by baseline RL agent for separating a 50\%/50\% butanol-water mixture, demonstrating that maximizing the product purity only leads to full separation but also in applying unit operations that do not have an effect on the separation quality (marked in red), as the RL agent is not penalized for using unnecessary operations. (d) Workflow for using \name co-pilot with agent and user actions for two iterations, resulting in optimal process design for separating a 50\%/50\% butanol-water mixture. (e) Text description of an exemplary chemical process that is used by \name.}
    \label{fig:process_design}
\end{figure}

\textbf{\name finds practically relevant processes by refining objectives in RL-based flowsheet design.} 
As shown in \Cref{fig:process_design}(a) and (c), using only the key objective of product purity for designing a separation process, i.e., the baseline, incentivizes the baseline RL agent to propose a flowsheet that results in optimal purity. However, without considering further objectives, such as capital costs, the RL agent might place unit operations that do not have an effect on the separation quality, or use a more complex flowsheet structure than needed. When using \name, we observe increased objective scores for capital costs and material flow intensity compared to the RL baseline, while the purity is at a high level that is close to ideal separation (\Cref{fig:process_design}(a)).
\name effectively assists human experts (at the co- and semi-pilot levels) in the iterative refinement and addition of objectives which includes balancing multiple objective weighting factors, illustrated exemplary with the human-agent interaction in \Cref{fig:process_design}(d) and quantified along the iterations in \Cref{fig:process_design}(b).
Also at the autopilot level, we observe significantly increased process performance compared to the RL-based chemical process design, such that \name enables automation of practically improved chemical processes.

\textbf{\name identifies and implements objectives that align with early-stage chemical process design.} 
Starting with maximizing the product purity as an initial objective, \name proposes a diverse set of useful objectives, such as process complexity, component recovery, and material efficiency, to be considered in the design.
In fact, we find that \name identifies and implements suitable process objectives and scoring functions at all levels (co-, semi- and autopilot, \Cref{supfig:SI_process_objectives_overview}), leading to higher overall scores on the evaluation metrics, \Cref{fig:process_design}(a). Adding additional objectives to the optimization also requires setting appropriate objective weights, see, e.g., \Cref{fig:process_design} (b) and (d), since we combine multiple objectives into one reward for the RL design agent.
As the design is sensitive to these objective weights, we see larger variation in individual objectives with less human intervention, particularly, for material flow intensity at semi- and autopilot level, see \Cref{fig:process_design}(a).
Notably, the product purity across all levels also shows some slight variations, which can be explained by partly conflicting objectives, as \name achieves high gains in capital costs and material flow intensity compared to the baseline.
Therefore, \name is able to enrich the chemical process design by relevant early-stage objectives.

\textbf{\name effectively analyzes chemical processes based on text representations.}
To analyze the flowsheet designs and propose new objectives, \name requires a text representation of the chemical processes. 
As indicated in \Cref{fig:process_design}(e), we represent the flowsheets as natural text description with the four categories: feed streams, unit operations, process flow, and product streams. %
\name successfully utilizes this representation to analyze process design potentials, e.g., by highlighting suboptimal product purity, as shown in \Cref{fig:process_design}(d), and identifying unit operations and flowsheet patterns that result in desired separation, \Cref{sup:SI_process_add_epx}.
These examples highlight the capability of \name to capture complex process context and advance automated chemical process design.

\section{Discussion}

Scientific discovery is often limited not only by the vastness of the hypothesis space, but also by the ``creativity'' of defining objectives that ultimately leads to new discovery. Fixed surrogate objectives can be incomplete, problem-specific, or vulnerable to misalignment and reward hacking, and are rarely sufficient to navigate open-ended discovery problems. \name, as a generalist agentic framework, address this challenge by iteratively evolving objectives and their realizations based on observed failure modes. By introducing an outer loop that proposes new objectives, implements executable scoring functions, analyzes outcomes, and selects final candidates, \name makes objective formulation a dynamic and autonomous discovery process.

Across five tasks spanning antibiotic design, nanobody design, functional DNA sequence design, inorganic materials design, and chemical process optimization, we find that iterating objectives is the key driver of progress in achieving a novel discovery with practical viability. 
In antibiotic design, \name grounds the optimization of biological activity in chemical reality, ensuring that high predicted potency reflects genuine therapeutic potential rather than exploitation of the score of a predictive model. Through rigorous experimental validation, \name discovers a structurally novel hit with antibacterial activity against \textit{E. coli} and no cytotoxicity in human cell lines, representing a promising starting point for further optimization.
In nanobody design, experimental BLI testing confirms three true binders among 24 candidates ($K_D$ as low as 300nM), and the composite objective autonomously constructed by \name significantly distinguishes binders from non-binders (p = 0.03), validating autonomous objective evolution as a practical paradigm for \textit{de novo} protein design.
In the discovery of functional DNA sequences, \name develops biology-driven objectives for joint optimization with expression-associated objectives, and designs enhancers with both significant novelty and and strong cell-type-level specificity.
In inorganic materials design, \name proposes objectives targeting mechanical properties and thermodynamic stability for superhard materials, which lead to excellent quality in the designed materials via independent metrics.
For chemical process engineering, \name reveals excessively complex flowsheet structures in the designed processes and circumvents them by considering the process complexity and flow intensity objectives to realize more relevant flowsheets.

A clear advantage of \name comes from its alignment with scientific practice: discovery is typically an interactive loop in which scientists interpret intermediate results, revise what to optimize next, and decide which constraints matter the most at a given stage. \name balances the time and resources spent by human scientist effort and automated agents, and operationalizes the scientific workflow through an agentic system with multiple levels of automation. Furthermore, \name enables scientists to interact and intervene with the discovery process when warranted while retaining the ability to run autonomously when objectives and evaluation pipelines are sufficiently mature. This interpretability and controllability offer significant efficiency and flexibility across tasks. In antibiotic design, \name allows chemists to identify problematic motifs such as uncommon rings or extended conjugated systems at intermediate interaction points. As a result, chemists could inject synthesis-oriented constraints early on, effectively steering the model away from synthetic liabilities and toward a more realizable chemical space. Similarly, in functional DNA sequence design, \name analyzes the properties of designed enhancers or promoters from the previous iteration and figures out the problems such as high off-target rates and lack of cell-type-specific motif, which may inspire biologists to modify the coming objectives and improve the candidates to match the biological constraints.

Despite these strengths, \name currently relies on computationally verifiable objectives. For scientific problems where results cannot be validated computationally, \name would need to be extended to incorporate feedback from either human experts or autonomous lab-in-the-loop systems. Additionally, the current high-level goals considered by \name often predefine the design space, e.g.\ all possible small organic molecules. To handle more flexible tasks, \name would need to automatically formulate the design space from high-level goals alone, such as finding a drug for curing a certain disease, and determine whether the appropriate modality is a small molecule, an antibiotic, a nanobody, or a DNA/RNA sequence.

More broadly, \name instantiates a new path towards automated AI scientist, where most current AI scientists rely on scaling model capability and tool space. The autopilot mode effectively discovers important objectives aligning with the goal of the task and implements the scoring functions to guide the optimization process. By assessing the failure modes in the optimized candidates across iterations, \name effectively leverages the optimization algorithm in the discovery process. One key advantage of this structure is mimicking the two thought modes, ``thinking, fast and slow'' \citep{kahneman2011thinking}. The inner loop optimization is thinking fast, exploring all reachable solutions given specific objectives and preferences, and the outer loop is thinking slow, evolving objectives and preferences given the full optimization results. We envision that addressing the limitations above, closing the loop with physical experiments and expanding the scope of autonomous problem formulation, represents a natural next step toward fully autonomous scientific discovery systems.

\section{Methods}

\subsection{\name Framework}

\subsubsection{Overview}

\name transforms open-ended scientific discovery into structured, iterative optimization by dynamically decomposing the high-level discovery goal into computable objectives. The framework comprises two nested loops: an \textit{outer loop} that explores and evolve objectives for the optimization; and an \textit{inner loop} that systematically optimize candidates against the scoring functions of the specified objectives.

The workflow proceeds as follows (\Cref{fig:framework_combined}(c)): users provide a \textit{high-level goal} in natural language, such as ``design novel antibiotic that are highly potent, safe, and synthesizable '' and can optionally provide more context information such as task background or specific requirements, as well as initial objectives and initial candidates as the starting points. The system then iterates through four core agentic modules: (1) \textit{Planner} formulates measurable objectives aligned with the overarching goal and informed by previous analysis; (2) \textit{Implementer} realizes executable scoring functions for proposed objectives; (3) \textit{Optimizer} optimizes candidates by iteratively generating and assessing candidates that maximize the objective scores, as the inner loop; and (4) \textit{Analyzer} assesses progress by analyzing objective score changes and examining specific candidates.

\subsubsection{Core Modules}

\noindent \textbf{Planner.} This agent decomposes the scientific goal into concrete optimization objectives at each iteration. Given the goal and current status analysis, it identifies gaps between the present state and desired outcome, proposing computable objectives with associated names, descriptions, optimization directions (e.g., maximize or minimize), and (optional) objective weights.

\noindent \textbf{Implementer.} This agent instantiates callable scoring functions for proposed objectives. When the implementation of an objective does not exist, it develops custom implementations by conducting web research on relevant computational methods and software packages and then implementing and validating the scoring function within a standardized Docker environment to ensure executability and correctness. If the implementer determines that an objective is not computable in a reliable and feasible manner, for instance, due to the objective being inherently intractable or the absence of available tools to support its computation, it notifies the planner to revise the plan accordingly.

\noindent \textbf{Optimizer.} This module constitutes the inner optimization loop. Given objectives and their scoring functions, it employs established optimization algorithms to generate improved candidates. The process alternates between batch evaluation using objective scoring functions and generation of new candidates designed to outperform previous iterations. The architecture accommodates diverse optimization strategies, such as prompted language models, trained reinforcement learning agents, or any optimization algorithms, enabling flexible tuning. The default optimizer for \name is a simple LLM-based evolutionary algorithm with two essential steps: (1) candidate generation: LLM proposes new candidates based on the current candidate pool, and (2) candidate scoring: scores all proposed candidates and selects top performers to update the pool.

\noindent \textbf{Analyzer.} This agent evaluates optimization status from two complementary perspectives. First, it tracks objective scores across iterations by computing statistical metrics (e.g., mean, variance, and improvement rate) and characterizing score trends to assess overall optimization progress. Second, it conducts in-depth analysis of specific candidates by writing code and employing other computational tools to examine their structural and property-level characteristics. Based on these analyses, the analyzer synthesizes an analysis report that summarizes the current optimization status and provides actionable suggestions for future optimization directions, serving as a key reference for the planner when formulating objectives in the next iteration. The analyzer also determines whether candidates satisfy the goal and can trigger early termination when success criteria are met.

\subsubsection{Autonomy Levels}
\label{sec:auto_level}

\name aligns with human scientific discovery workflows and seamlessly supports human intervention at varying levels. We define three operational modes based on the degree of autonomy (\Cref{fig:framework_combined}(d)):

\begin{itemize}
    \item \textbf{Co-pilot}: Human scientists collaborate closely with both the planner and analyzer. At each iteration, these agents generate proposals (i.e., new objectives from the planner, and analysis from the Analyzer), which scientists can either accept directly or revise based on domain expertise. The implementer and optimizer operate autonomously within the outer loop, executing the human-approved objectives. This mode maximizes human control while automating implementation details.
    
    \item \textbf{Semi-pilot}: Human intervention is limited to the analyzer stage. Scientists review progress reports and optimization outcomes, providing feedback that guides the planner's subsequent objective proposal. The planner, implementer, and optimizer function autonomously, but strategic decisions about continuation, termination, or pivoting remain human-guided. This mode balances automation with critical oversight at decision points.
    
    \item \textbf{Autopilot}: All four modules operate fully autonomously without human intervention. The system independently plans objectives, implements scoring functions, optimizes candidates, and analyzes results. This mode enables large-scale automated exploration when domain constraints are well-specified and trust in the system is established.
\end{itemize}

This tiered design ensures scientists can interact with \name in ways that maximize productivity for their specific research context, from hands-on collaboration to fully autonomous discovery.

\subsection{Task Configurations}

\textbf{Antibiotic discovery}. We formulate this task to discover novel antibiotics. In practice, we set the high-level discovery objective as designing candidates with strong predicted antibacterial efficacy while maintaining structural novelty, favorable predicted mammalian-cell safety, avoidance of dominant known-antibiotic motifs, and practical feasibility aligned with purchasable-like chemical space for wet-lab validation (details in \cref{sup:expdetail_drug}). Both the goal and the accompanying contextual information explicitly encode our design target and related constraints (detailed in \Cref{sup:expdetail_drug}). For each \name instance, our initial objectives are always to maximize antibiotic activity, molecule novelty, and synthesizability, while minimizing toxicity to human and similarity to known antibiotic motifs in the designed molecules. During the loop of optimization, we use the default LLM-based evolutionary algorithm. The initial populations are selected from the Enamine REAL Database \citep{shivanyuk2007enamine}, which also serves as the first group of molecules in the parent node. We provide molecules from the parent node, individual score from each objective, and an aggregated score (by product individual scores) to the LLM, and generate new molecules after crossover operation. We then select the top molecules based on the list containing both generated molecules and molecules from the parent node. To encourage diversity, we consider a cluster-based selection strategy (Butina cluster-based selection \citep{butina1999unsupervised} detailed in \cref{sup:expdetail_drug}). Finally, we combine all scoring functions into a single scalar value by product of expert to discourage ignoring any objective and select top molecules across all iterations. We use the standard implementation of the planner, implementer, optimizer, and analyzer modules.

\textbf{Nanobody design}.
Nanobodies, also known as single-domain antibodies derived from camelids, represent a promising therapeutic modality due to their small size, high stability, and ease of engineering \citep{muyldermans2021applications}. We formulate this task as the \textit{de novo} design of high-affinity nanobodies targeting PD-L1, a key immune checkpoint exploited by tumors for immune evasion. Our high-level discovery objective specifies designing candidates with strong predicted binding affinity, favorable interface quality, and practical sequence developability (details in \cref{sup:expdetail_nanobody}). Both the high-level goal and the accompanying contextual information explicitly encode the binding target and key residue contacts on the PD-L1 epitope (detailed in \Cref{sup:context_nanobody}). We adopt the nanobody scaffold provided by BoltzGen, based on caplacizumab (PDB: 7EOW), which defines the framework regions and three designable Complementarity-Determining Region (CDR)  loops with variable-length insertions. The initial population is sampled from LLM-generated random nanobody sequences. The optimization process is initialized with a set of scoring terms commonly used in protein binder design, including AlphaFold-derived confidence metrics (iPTM, pTM, and PAE), physicochemical interface features (e.g., hydrogen bonds, salt bridges, and buried surface area), as well as sequence-level penalties such as hydrophobicity and liabilities. These terms are combined through a weighted aggregation scheme that is updated throughout the search procedure. During optimization, we employ a genetic algorithm with hybrid crossover operators consisting of 40\% CDR swap, 40\% single-point crossover, and 20\% uniform crossover, together with random CDR mutation. To encourage diversity while maintaining quality, we use tournament selection for parent pairing and elitism-aware survival selection. Candidate evaluation uses structure prediction with Boltz2. We apply diversity filtering based on CDR-only sequence similarity, rejecting any candidate with more than 50\% CDR identity to a selected sequence. Finally, we combine objectives using normalized weighted aggregation and select top candidates based on rank-based scoring across all iterations. We use the standard implementation of the planner, implementer, optimizer, and analyzer modules.

\textbf{Functional DNA sequence design}. Functional DNA sequences, also referred to as cis-regulatory elements (CREs), primarily include enhancers and promoters and play a central role in regulating gene expression levels \citep{wittkopp2012cis, de2024hold}. We focus on the \textit{de novo} design of cell-type-specific enhancers and promoters across multiple cellular contexts, including HepG2 (enhancer and promoter), K562 (enhancer and promoter), SKNSH (enhancer and promoter), A549 (promoter only), and GM12878 (promoter only). The selection of these cell lines is constrained by the availability of high-quality, publicly accessible datasets. We formulate the discovery task using a high-level natural-language prompt that specifies the objective of generating functional DNA sequences with strong cell-type specificity. Both the high-level goal and the accompanying contextual information explicitly encode target and off-target cell-type constraints (see \Cref{sup:highlevel dna,sup:context dna}). During optimization, the primary objectives are to maximize predicted expression in the target cell line while suppressing activity in non-target cell lines. For optimization, we employ a default LLM-based evolutionary algorithm. The initial population is selected by sampling from a pool of random DNA sequences. 
During candidate selection, we keep all candidates that satisfy the expression selectivity. Moreover, we also keep top 50\% diverse candidates measured by average pairwise Hamming distance. Finally, we use the standard implementation of the outer loop and the analyzer, planner, and implementer agents.

\textbf{Inorganic materials discovery}.
We consider two materials inverse design tasks. The first task aims to design permanent magnets with low supply chain risk, specified by two objectives: magnetic density higher than 0.2 $\text{\AA}^{-3}$ and HHI score below 1500. The initial objective is set to maximize magnetic density. The \name Co-pilot mode is deployed with iteratively refined objectives: maximizing magnetic density in the first iteration, followed by the addition of HHI score minimization in the second. This task provides a direct comparison with MatterGen \citep{mattergen}. The second task is to design superhard materials for precision cutting, requiring high hardness, high elastic modulus, appropriate brittleness, and thermodynamic stability. The high-level goal and contextual information explicitly encode design requirements and constraints (Supplementary Section \Cref{sup:mat_goal} and \Cref{sup:mat_context}). For each \name experiments of superhard materials design, initial objectives are set to maximize bulk modulus and shear modulus, which are important indicators for screening superhard materials \citep{superhard1}. The optimization loop employs a default LLM-based evolutionary algorithm. Initial populations are randomly sampled from the Materials Project database \citep{mp}, which also serves as the first group of crystals in the parent node. Based on LLM-proposed chemical formulas, pretrained diffusion models provide 3D crystal structures, with geometric optimization performed using universal ML force fields \citep{mattersim}. Evaluators assign objective scores based on the 3D structure of each crystal. Chemical formulas from the parent node and individual score of each objective were provided to the LLM, which generated new formulas through crossover operations. Optimal structures are then selected via Pareto front analysis from a combined pool of generated and parent crystals. The standard implementation of the planner, implementer, optimizer, and analyzer modules are used for all materials design tasks.

\textbf{Chemical process design}.
We use \name for the design of chemical process, more specifically separation process flowsheets, which is a central task in chemical engineering~\citep{biegler1997systematic, turton2008analysis, gao2024deep}.
The high-level goal is formulated as a natural language prompt targeting the design process flowsheets for the steady-state separation of an azeotropic binary mixture of water and ethanol at different feed compositions into high purity streams, cf. \Cref{sup:SI_process_exp}.
For the optimizer designing process flowsheets in the inner loop, we use an RL agent based on the separation process design framework by~\citet{gottl2025deep}, see details in \Cref{sup:SI_process_exp}.
The the action space of the RL agent comprises the (1) selection of suitable unit operations, such as decanters, distillation columns, and mixers with their specifications, and (2) determination of the material flow structures (including recycles) that connect the unit operations.
We translate the RL-internal matrix representation of process flowsheets to a text description, see \Cref{supfig:SI_process_representation} for details.
We use the standard implementation of the analyzer, planner and implementer, and the text description is provided to the agents in the outer loop.
The proposed new objectives -- with corresponding weighting factors to aggregate the objective values into one reward value -- are automatically added to the RL framework and used for the next iteration of process design, which always starts from scratch without an initial population, whereby the initial objective for the first iteration is the product purity.
We thus focus on the iterative addition and refinement of suitable chemical process design objectives and their weighting factors.

\subsection{Task Evaluations}

We validate the performance of \name on each individual task by setting up a set of evaluation metrics. The evaluation metrics are unseen during the online running procedure of \name. Below, we briefly discuss the evaluation procedures for each task.

\textbf{Antibiotic discovery}. To mimic real-world lab experiment, we consider evaluating the candidates from the perspectives of biological objectives, synthesizability, and drug likeness. These three areas can be covered with 11 different computational metrics. To evaluate generated molecules with biological objectives, we consider antibiotic activity score, novelty score, toxicity score, and known motif filter score as metrics. For synthesizability, we consider a synthetic accessibility score as the metric. Last but not least, to evaluate drug likeness, we consider QED score, DeepDL prediction score, molecular weight score, PAINS filter score, BRENK filter score, and RING score as metrics. Detailed implementation and evaluation protocols are provided in \Cref{sup:expdetail_drug}. When evaluating candidates proposed by baselines and \name, we compute both the absolute score and pass rate of the top 100 molecules selected using each model's optimization objectives for a fair comparison. 

\textbf{Nanobody design}. To emulate real-world therapeutic antibody development, we adopt a comprehensive computational assessment framework spanning structural quality, binding interface characteristics, epitope engagement, and sequence developability. Structural quality is evaluated using confidence metrics from structure prediction, including interface predicted TM-score (iPTM), overall predicted TM-score (pTM), and predicted local distance difference test (pLDDT) for the full binder, the CDR regions, and the CDR3 loop, together with predicted aligned error (PAE) at the binding interface. Binding interface characteristics are assessed using physically interpretable interaction metrics computed on predicted complex structures, including the number of hydrogen bonds, salt bridges, and the change in solvent-accessible surface area upon binding ($\Delta$SASA). We further quantify epitope engagement using CDR-hotspot and CDR3-hotspot contact counts, measuring how many CDR residues fall within contact distance of predefined PD-L1 epitope residues. Sequence–structure compatibility is assessed with ProteinMPNN \citep{dauparas2022robust} by computing the negative log-likelihood score and expected sequence recovery on the predicted structure. We validate CDR3 secondary structure using DSSP assignment \citep{kabsch1983dictionary} on predicted structures, verifying proper alpha-helical content within the specified positional constraints. Sequence developability is evaluated with a liability score that penalizes known sequence liabilities such as deamidation sites, oxidation-prone residues, and aggregation motifs. To improve robustness to predictor-specific biases, we perform structure prediction with both AlphaFold3 and Boltz2. When evaluating candidates proposed by baselines and \name, we report metrics computed under both structure prediction backends and select top candidates using rank-based aggregation across objectives. Detailed implementation and evaluation protocols are provided in \Cref{sup:expdetail_nanobody}.

\textbf{Functional DNA sequence design}. To emulate real-world experimental evaluation, we adopt a blind computational assessment framework based on five established computational oracles drawn from prior studies \citep{gosai2024machine,dasilva2024dna,chen2025ctrl,lal2024designing}. As a representative task, we focus on the design of HepG2-specific enhancer sequences. The evaluation metrics include statistical comparisons of MPRA-measured expression between the target cell line and non-target cell lines (e.g., HepG2 vs. K562 and HepG2 vs. SKNSH), together with knowledge-driven criteria such as transcription factor motif enrichment, sequence diversity, and sequence stability. Detailed implementation and evaluation protocols are provided in \Cref{sup:expdetail dna}.

\textbf{Inorganic materials design}.
To evaluate material properties, density functional theory (DFT) calculations were employed to determine the electronic, magnetic, and mechanical properties of generated materials, as well as energy above hull \citep{mattergen, matinvent}. HHI scores were computed using the pymatgen package \citep{pymatgen}. In the task of designing permanent magnets with low supply chain risk, two objectives were specified: magnetic density higher than 0.2 $\text{\AA}^{-3}$ and HHI score less than 1500. In the task of designing superhard materials for precision cutting, the evaluation metrics include Vickers hardness, bulk modulus, shear modulus, Pugh ratio, and energy above hull. More details of the evaluation protocols are described in \Cref{sup:mat_eval}.

\textbf{Chemical process design}. To cover early-stage process design goals, we utilize the short-cut simulations models developed in~\citep{gottl2025deep} and calculate three process performance indicators.
These are used as the evaluation metrics and include the product purity, capital costs, and material flow intensity.
The product purity corresponds to the average purity of the product streams received from the simulation.
The capital costs represent the sum of individual unit operation costs estimated on a simple heuristic, similar to~\citep{gottl2025deep}.
For the material flow intensity, we calculate the recycle ratios and introduce penalty terms for excessive ratios and very small streams ($<1\%$ of the feed stream).
We refer to the \Cref{sup:SI_process_exp} for further implementation details.

\subsection{Experimental Validation}

\textbf{Antibiotic discovery}.
Compounds with high purity (>90\%) were synthesized by Enamine and Onepot Inc. To test the antibacterial activity of each compound, we grow \textit{E. coli} cells overnight in 3 mL LB medium and diluted 1/10,000 into fresh LB. In 96-well flat-bottom plates (Corning), cells are then introduced to compound at an initial concentration of 128$\mu$g/mL, mixed with 32$\mu$g/mL polymyxin B nonapeptide. The plates are then incubated at 37°C without shaking until untreated control cultures reach stationary phase, at which time plates were read at 600 nm using a SpectraMax M3 plate reader. Cell viability values are normalized by the mean of two DMSO controls. For compounds 4, 8, 9, and 27, we repeat the same procedure with lower concentrations to plot their dose response curves and determine their MIC.

Cytotoxicity in human cells was assayed using a resazurin (alamarBlue) assay. HepG2 and HEK293 cells are obtained from ATCC, passaged <10 times, and grown to log phase in high-glucose Dulbecco's Modified Eagle Medium (DMEM; Corning 10-013-CV) supplemented with 10\% fetal bovine serum (FBS; ThermoFisher 16140071) and 1\% penicillin-streptomycin (ThermoFisher 15070063). Cells are plated into 96-well clear flat-bottom black tissue-culture-treated plates (Corning 3603) at a density of 104 cells/well using 100$\mu$l working volumes, then incubated at 37°C with 5\% CO2. 24h after plating, test compounds are added and automatically mixed to facilitate homogeneous distribution of compounds. Cells were re-incubated for two days, with the incubation period chosen to reflect the relative timescales of cell doubling for each cell type. After an additional 4 to 24h of incubation, fluorescence excitation/emission at 550/590nm was read using a SpectraMax M3 plate reader or an EnVision plate reader and EnVision Workstation software (version 1.14.3049.1193, PerkinElmer). Cell viability values are normalized by the mean of two DMSO controls.

\textbf{Nanobody design}.
All BLI experiments were performed on a Gator Plus (Gator Bio) at 25C with a shaking speed of 400 rpm. The nanobodies were produced using the PURExpress In Vitro Protein Synthesis Kit (New England BioLabs) with a miniaturized reaction volume of 5uL and a 4-hour incubation at 37C. The reaction was diluted 1:100 in sample buffer (PBS, pH 7.4, 0.05\% (v/v) Tween-20 and 0.1\% (v/v) recombinant albumin) and loaded onto Streptactin-XT probes (Gator Bio) using a 2nm wavelength shift threshold. Recombinant Human PD-L1/B7-H1 His-tag Protein (R\&D Systems) was serially diluted in sample buffer to between 1000 nM and 31.25 nM. The sensorgrams were obtained with a 120s baseline, loading to threshold, 180s post-loading baseline, 400s association, and 400s dissociation. Raw data were corrected with double referencing, Savitzky-Golay filtering, and alignment to the association phase. Sensorgrams were analyzed in the Gator Plus Results \& Analysis software. Kinetic fitting was performed using a global 1:1 binding model with unlinked Rmax, from which the $K_{on}$ and $K_{off}$ were calculated. Steady-state analysis was performed using the measured response values across the analyte concentration series to estimate equilibrium binding affinity ($K_D$).

\section*{Code Availability}
Our code is available at \url{https://github.com/btyu/SAGA}. The license is MIT license.

\section*{Acknowledgments}
YD acknowledges the support of Cornell University. YD thanks Bowen Deng, Peter Clark and Reece Adamson for helpful feedback. TL acknowledges the support of Wu Tsai Institute at Yale especially Ping Luo, and Zhaorong Li for suggestions in manuscript. KS thanks Jie Li and Michael K. Gilson for helpful feedback. CPG acknowledges the support of an AI2050 Senior Fellowship, a Schmidt Sciences program, the National Science Foundation (NSF), the National Institute of Food and Agriculture (USDA/NIFA), the Air Force Office of Scientific Research (AFOSR), and Cornell University. ZS, HJ, and CD thank their entire team from Deep Principle for support. CD thanks Yi Qu for discussions. JC and PS acknowledge support from the NCCR Catalysis (grant number 225147), a National Centre of Competence in Research funded by the Swiss National Science Foundation.
DBR acknowledges support from the 2024 Larry Leeds, Jenny \& Larry Goichman, and Ben Shenkman – PCF Young Investigator Award.
NH acknowledges support from the Torrey Coast Foundation and NIH/NCI IMAT R61 CA281807.
JGR acknowledges funding by the Swiss Confederation under State Secretariat for Education, Research and Innovation SERI, participating in the European Union Horizon Europe project ILIMITED (101192964). CM acknowledges Valence Labs for financial support. HS acknowledges the support of NSF CAREER \#1942980. KS, YW, and THG thank the National Institute of Allergy and Infectious Disease grant U19-AI171954 for support. WJ thanks Divya Nori and Weian Mao for discussions and acknowledges funding from Google Research Scholar Award and support from NVIDIA.

\section*{Author Contributions} 
Coordination and planning: Yuanqi Du (lead), Botao Yu; 
Framework design and development: Botao Yu (lead); 
Task implementation: Tony Shen, Tianyu Liu, Junwu Chen, Jan G. Rittig, Yikun Zhang, Kunyang Sun, Cassandra Masschelein;
Antibiotic design: Tony Shen (co-lead), Kunyang Sun (co-lead), Tianyu Liu (co-lead), Yikun Zhang, Yingze Wang, Bo Zhou and Cassandra Masschelein;
Experimental validation (antibiotics): Aarti Krishnan (lead), Yu Zhang;
Nanobody design: Yikun Zhang (lead);
Experimental validation (nanobodies): Daniel Rosen (lead), Rosali Pirone;
Functional DNA sequence design: Tianyu Liu (lead); 
Inorganic materials design: Junwu Chen (lead); 
Chemical process design: Jan G. Rittig (lead); 
Writing of the original draft: Yuanqi Du, Botao Yu, Yikun Zhang, Tianyu Liu, Tony Shen, Jan G. Rittig, Kunyang Sun, Junwu Chen, Wengong Jin; 
Editing of the original draft: everyone; 
Supervision, conceptualization and methodology: Yuanqi Du, Teresa Head-Gordon, Carla P. Gomes, Huan Sun, Chenru Duan, Philippe Schwaller and Wengong Jin.

\section*{Competing Interests}
The authors declare that they have no conflict of interests at this time.

\defaultbibliography{ref}       

\clearpage
\putbib
\end{bibunit}

\clearpage

\begin{bibunit}
\appendix

\crefalias{section}{suppsection}
\crefalias{figure}{suppfig}
\crefalias{table}{supptab}

\begin{center}
\vspace*{1em}
{\fontsize{30}{19.2}\selectfont\textit{Supplementary Information} for \name\par}
\vspace{1.5em}

\authorblock{
  \authorrow{
  Yuanqi Du\aumark{1,\emark,${\dagger}$},
  Botao Yu\aumark{2,\emark},
  Tianyu Liu\aumark{3,\emark},
  Tony Shen\aumark{4,\emark},
  Junwu Chen\aumark{5,\emark},
  Jan G. Rittig\aumark{5,\emark}, Kunyang Sun\aumark{6,\emark},
  Yikun Zhang\aumark{7,\emark},
  }
  \authorrow{
  Aarti Krishnan\aumark{8,9,10,11},
  Yu Zhang\aumark{8,9,10},
  Daniel Rosen\aumark{8,12},
  Rosali Pirone\aumark{8},
  Zhangde Song\aumark{13},
  Bo Zhou\aumark{14},
  Yingze Wang\aumark{6},
  }
  \authorrow{
  Cassandra Masschelein\aumark{5},
  Haorui Wang\aumark{15},
  Haojun Jia\aumark{13},
  Chao Zhang\aumark{15},
  Hongyu Zhao\aumark{3},
  Martin Ester\aumark{4},
  Nir Hacohen\aumark{8,16},
  }
  \authorrow{
  Teresa Head-Gordon\aumark{6,${\dagger}$},
  Carla P.~Gomes\aumark{1,${\dagger}$},
  Huan Sun\aumark{2,${\dagger}$},
  Chenru Duan\aumark{13,$^{\dagger}$},
  Philippe Schwaller\aumark{5,${\dagger}$},
  Wengong Jin\aumark{7,8,${\dagger}$}
  }
}
\end{center}

\begingroup
\renewcommand{\thefootnote}{}
\footnotetext{\scriptsize
  $^{1}$Cornell University, Ithaca, NY, USA;
  $^{2}$The Ohio State University, Columbus, OH, USA;
  $^{3}$Yale University, New Haven, CT, USA;
  $^{4}$Simon Fraser University, Burnaby, BC, Canada;
  $^{5}$École Polytechnique Fédérale de Lausanne, Lausanne, Switzerland;
  $^{6}$University of California Berkeley, Berkeley, CA, USA;
  $^{7}$Northeastern University, Boston, MA, USA;
  $^{8}$Broad Institute of MIT and Harvard, Cambridge, MA, USA;
  $^{9}$Massachusetts Institute of Technology, Cambridge, MA, USA;
  $^{10}$Wyss Institute for Biologically Inspired Engineering, Harvard University, Boston, MA, USA;
  $^{11}$Whitehead Institute for Biomedical Research, Cambridge, MA, USA;
  $^{12}$Brigham and Women's Hospital and Dana-Farber Cancer Institute, Boston, MA, USA;
  $^{13}$Deep Principle, Boston, MA, USA;
  $^{14}$University of Illinois Chicago, Chicago, IL, USA;
  $^{15}$Georgia Institute of Technology, Atlanta, GA, USA;
  $^{16}$Massachusetts General Hospital, Krantz Family Center for Cancer Research, Boston, MA, USA;
  $^{\star}$These authors contribute equally
  $^{\dagger}$Correspondence to: yd392@cornell.edu, thg@berkeley.edu, gomes@cs.cornell.edu, sun.397@osu.edu, duanchenru@gmail.com, philippe.schwaller@epfl.ch, w.jin@northeastern.edu
}
\endgroup

\renewcommand{\thesection}{S\arabic{section}}%
\counterwithin{figure}{section}
\counterwithin{table}{section}
\counterwithin{equation}{section}
\renewcommand{\thefigure}{\thesection.\arabic{figure}}%
\renewcommand{\thetable}{\thesection.\arabic{table}}%
\renewcommand{\theequation}{\thesection.\arabic{equation}}%

\clearpage
\startcontents[sections]
\printcontents[sections]{l}{1}{\setcounter{tocdepth}{3}}
\clearpage

\section{Implementation Details}

\subsection{Framework Design}

The \name framework translates high-level scientific goals into iterative optimization procedures targeting dynamically evolving sets of specific objectives. The system accepts the following inputs:

\begin{itemize}
\item \textbf{Goal}: A high-level design goal specified in natural language that defines the scientific task.
\item \textbf{Context information} (optional): Supplementary descriptions of task background or specific requirements regarding objectives, enabling the framework to better understand the task domain and propose objectives aligned with domain-specific scientific needs.
\item \textbf{Initial objectives} (optional): Predefined objectives accompanied by their corresponding scoring functions, which are incorporated into the first iteration as a foundation for optimization.
\item \textbf{Initial candidates} (optional): Randomly initialized candidate hypotheses defining the initial search space.
\end{itemize}

Upon completion, \name outputs design hypotheses that satisfy the specified goal.

\subsubsection{Key Concepts}

The framework defines four key concepts that structure information flow throughout the system:

\textbf{Candidate} represents an individual solution in the optimization space. Each candidate encapsulates a domain-specific representation (e.g., SMILES strings for molecular structures, multi-domain dictionary object for materials) and objective scores from multiple objectives. Candidates are uniquely identified and tracked across iterations, enabling provenance tracking and historical analysis.

\textbf{Population} manages collections of candidates as a cohesive unit. Beyond storing candidate lists, each population provides methods for batch objective scoring and statistics over stored candidates. Populations serve as the primary data structure passed between modules during optimization.

\textbf{Objective} specifies what should be optimized using a natural language description. Each objective can be one of the three types: candidate-wise objectives that evaluate individuals independently, population-wise objectives that assess collective properties, and filter objectives that enforce binary constraints. Each objective includes an optimization direction (maximize or minimize, not applicable for filter objectives) and an optional weight for multi-objective aggregation.

\textbf{Scoring Function} implements the computational logic for evaluating candidates against objectives. Scoring functions accept candidate representations and return numerical scores (for candidate-wise or population-wise objectives) or boolean values (for filter objectives). Each scoring function is implemented as an Model Context Protocol module, enabling isolated execution in Docker containers with standardized input/output interfaces. This design provides dependency isolation, reproducibility across environments, and safe execution of potentially unsafe code.

\subsubsection{Modules}

\begin{figure}
    \centering
    \includegraphics[width=1\linewidth]{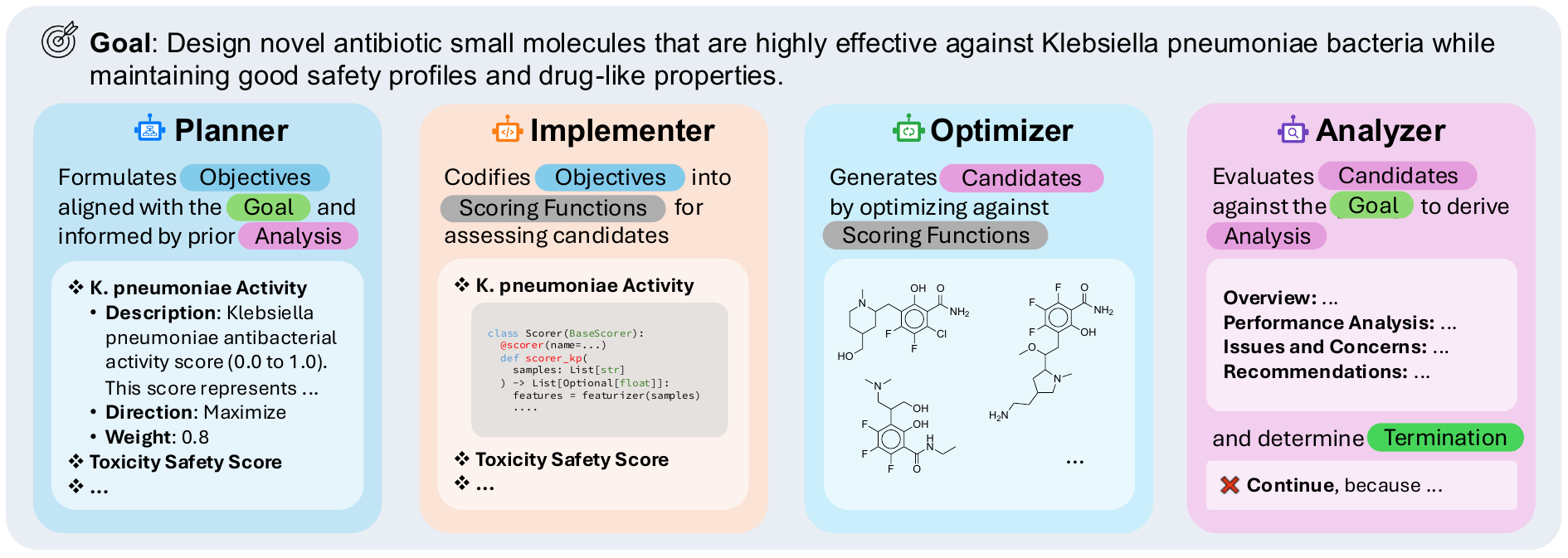}
    \caption{The functional illustration of the four core modules.}
    \label{fig:core_modules_illustration}
\end{figure}

The framework comprises four \textbf{core modules} (\Cref{fig:core_modules_illustration}) that implement the main optimization workflow:

\textbf{Planner} systematically decomposes the high-level scientific goal into concrete, computationally tractable objectives at each iteration. The module receives as input the optimization goal, optional context information that introduces the task background or specifies specific requirements, and analysis report from the last iteration. Through LLM-powered reasoning, it identifies gaps between the current state and desired outcomes, formulating objectives that specify what should be measured (description), how it should be optimized (maximize or minimize), and its relative importance (weight). Each objective must be specific enough to guide implementation and capture meaningful scientific properties. The module outputs a list of objectives with complete specifications that guide subsequent implementation and optimization phases.

\textbf{Implementer} codifies abstract objectives into executable scoring functions. Given a list of proposed objectives, the module first attempts to match each objective with existing scoring functions that may be provided with the initial objectives or proposed in previous iterations through semantic similarity analysis, comparing objective descriptions using LLM-based pairwise evaluation. When an exact match cannot be found, the Implementer autonomously implements new scoring functions through a multi-step process: (1) conducting web-based research to identify relevant computational methods, software packages, and scientific models; (2) synthesizing this information into executable Python code following standardized interfaces; (3) testing the implementation with specified dependencies and sample candidates. All scoring functions, whether retrieved or newly created, are deployed in isolated Docker containers with specified dependencies, preventing conflicts and ensuring safety execution. The module outputs objectives with attached, validated scoring functions ready for optimization.

\textbf{Optimizer} executes the inner optimization loop, systematically generating and evaluating candidate solutions to optimize objective scores. The optimizer receives the current population and objectives with scoring functions as input, and outputs an improved population. In practice, each task typically implements its own task-specific optimizer tailored to the structure and constraints of the problem, enabling stronger performance through domain-appropriate optimization strategies such as evolutionary algorithms and reinforcement learning agents. Regardless of the specific algorithm, the optimizer generally operates in two steps iteratively: (1) \textit{candidate generation}, where new candidates are proposed based on the current population, for example, by using an LLM to analyze high-performing candidates and generate novel ones designed to improve upon observed patterns, or by applying task-specific mutation and crossover operators; and (2) \textit{candidate scoring}, where all candidates (existing and newly generated) are evaluated against each objective using their attached scoring functions, with parallel execution for efficiency, followed by ranking candidates according to weighted objective scores to assemble the next generation's population while maintaining a proper population size. For tasks without a dedicated optimizer, \name provides a default LLM-based evolutionary algorithm that implements the above two-phase procedure using an LLM as the candidate generator. 

\textbf{Analyzer} performs comprehensive evaluation of optimization progress and synthesizes actionable insights to guide subsequent planning. The module receives the current optimized population, active objectives, and historical summaries from previous iterations. It conducts analysis from two complementary perspectives. First, it tracks objective scores across iterations by computing statistical metrics (mean, standard deviation, extrema) for each objective and characterizing score trends and improvements relative to prior iterations, identifying signs of convergence, stagnation, or conflicting objectives. Second, it conducts in-depth analysis of specific candidates by writing and executing Python code to compute domain-relevant metrics and extract structural or property-level features from the candidate population; it also leverages a provided set of domain-specific tools (e.g., scientific simulation or evaluation tools) from ToolUniverse~\citep{gao2025tooluniverse} to perform deeper, task-specific investigation, for example, examining top-performing and underperforming candidates to understand what properties drive high scores, assessing population diversity, and uncovering quality aspects or failure modes not captured by current objectives. The Analyzer integrates these two perspectives to synthesize a structured analysis report containing four sections: (1) \textit{overview} summarizing the current optimization state and key population characteristics; (2) \textit{performance analysis} highlighting score improvements, regressions, and trends across iterations; (3) \textit{issues and concerns} identifying problems such as stagnation, poor diversity, or conflicting objectives; and (4) \textit{strategic recommendations} proposing actionable adjustments for the next iteration, serving as a key reference for the Planner's subsequent objective formulation. Beyond reporting, the Analyzer makes a termination decision by evaluating whether optimization goals have been satisfied, whether scores have converged, and whether continued iteration would yield meaningful improvements. The module outputs an analysis report that informs the Planner's next objective formulation and a termination decision (continue or stop).

In addition to the above core modules, upon the end of the iterations, a \textbf{selector} agent is used to systematically evaluates all candidates generated throughout the optimization process, selecting solutions that best satisfy the discovery goal. It gets all candidates along with all objective scores as input, and writes code and optionally call scientific tools from ToolUniverse to comprehensively assess and rank all candidates and returns a specified number of top candidates.

\subsection{Execution Workflow}

The \name framework executes through a structured outer loop consisting of the following phases:

\textbf{Phase 0: Initialization.} Before the first iteration begins, the system instantiates all modules and processes user-provided inputs. If users provide an initial population, it is stored and marked as iteration 0. If initial objectives with scoring functions are provided, the Planner will be suggested to use them, and the Implementer would omit implementing them on its own.

\textbf{Phase 1: Planning.} At the start of each iteration, the Planner receives the optimization goal, optional context information that introduces task background or specifies specific requirements, and the analysis report from the previous iteration (iterations 2 onward). The Planner formulates objectives for the current iteration, considering optimization progress, remaining gaps, and strategic priorities.

\textbf{Phase 2: Implementation.} The  processes each objective in parallel, attempting to match it with existing scoring functions through LLM-based semantic comparison or creating new implementations when matches are not found. If some objectives cannot be matched or implemented, the system records the unmatched objectives and invokes the Planner with information about which objectives failed and why. The Planner then revises its objective proposals to better align with available computational capabilities. This planning-implementation retry loop continues for up to a configurable maximum number of attempts until all objectives have attached scoring functions. Successfully matched objectives proceed to optimization.

\textbf{Phase 3: Optimization.} The optimizer receives the current population (from the previous iteration) and objectives with scoring functions. If configured, a specified ratio of the population is randomly replaced with newly generated random candidates to maintain diversity. The optimizer then executes its algorithm-specific optimization process, which for the default LLM-based evolutionary optimizer involves multiple generations of candidate proposal, batch evaluation, and selection until convergence criteria are met or generation limits are reached.

\textbf{Phase 4: Analysis.} The Analyzer receives the optimized population, current objectives, and historical summaries. It evaluates all candidates against objectives, computes score statistics and trends, and conducts detailed candidate investigations using coding and domain-specific tools. The analysis results are synthesized into a structured report. The Analyzer also evaluates termination criteria, including goal satisfaction, score convergence, and resource constraints, and recommends whether to continue to the next iteration or terminate optimization. If termination is not recommended and the maximum iteration count has not been reached, the workflow returns to phase 1 for the next iteration.

\textbf{Phase 5: Finalization.} Upon termination (either by Analyzer decision or maximum iterations reached), the system collects the results and process logs and saves them for reproducibility and provenance. Human scientists can optionally use the selector to perform retrospective candidate evaluation. It retrieves all candidates generated across all iterations, evaluates them comprehensively against the original discovery goal using custom analysis code and computational tools, and selects the top candidates holistically with all objectives considered rather than solely by final-iteration scores. This retrospective selection ensures that high-quality solutions from early iterations exploring different objective combinations are retained. 

\subsection{Experimental Setups}

We used the following LLM settings for each module:
\begin{itemize}
    \item \textbf{Planner}: \texttt{gpt-5-2025-08-07}
    \item \textbf{Implementer}: \texttt{gpt-5-2025-08-07} for matching existing scoring functions to objectives, and the Claude Code agent with \texttt{claude-sonnet-4-5-2025-0929} for implementing new scoring functions.
    \item \textbf{Optimizer}: Task-specific. Different tasks may use different LLMs as backbones, or no LLM if optimization strategies are not LLM-based.
    \item \textbf{Analyzer}: the Claude Code agent with \texttt{claude-sonnet-4-5-2025-0929} for investigating specific candidates, \texttt{gpt-5-2025-08-07} for writing comprehensive analysis and making termination decisions.
    \item \textbf{Selector}: the Claude Code agent with \texttt{claude-sonnet-4-5-2025-0929}.
\end{itemize}

For each of the tasks, the input includes a high-level goal that briefly describes the design goal, context information that provides task background or specifies specific requirements, a set of initial objectives with scoring functions and a population of randomly initialized candidates as the starting point. See task-specific sections for more details.

All code and configurations will be released online.

\subsection{Extensibility and Customization}

The \name framework is designed for extensibility at multiple levels:

\textbf{Custom Module Implementations.} Each module defines an abstract base class with required methods. Users can implement custom modules by subclassing these bases and registering implementations in the module registry. For example, users can easily replace the default LLM-based optimizer with a genetic algorithm or Bayesian optimization, or implement new workflow and add advanced features for the Planner. This modular and extensible codebase enables \name to be continuously updated and customized to more tasks.

\textbf{Custom Scoring Functions.} Users can add manually implemented task-specific scoring functions by following a specific code template. This allows users to provide their existing objectives and corresponding computational methods to the system, for more accurate and faster optimization. The Implementer agent will also refer to user-provided scoring functions when implementing new ones.

\textbf{Configuration-Based Customization.} The framework uses structured configuration files (JSON or YAML) that specify module selections and versions, LLM settings per module (model, temperature, max tokens), loop parameters (max iterations, convergence thresholds, random injection ratio), and logging verbosity and output directories.
This configuration-driven design enables experimentation with different setups without code modification.

\subsection{\name Reduces to the Optimizer agent (\name-Opt)}
The default optimizer agent is a simple evolutionary optimization algorithm, thus \name reduces to \name-Opt when the outer loop is disabled. \name-Opt only optimizes with fixed objectives as input, similar to other frameworks such as AlphaEvolve \citep{novikov2025alphaevolve}. For \name-Opt experiments, we always use this default version without any specialized design.

\subsection{Comparing \name with Other AI Scientists}

We discuss the comparison between \name and other AI scientists from multiple dimensions in \Cref{tab:ai_scientist_comparison}.

\begin{table*}[t]
\centering
\small
\resizebox{0.98\textwidth}{!}{
\begin{tabular}{lccccccc}
\toprule
\textbf{System} & \textbf{Type} & \textbf{Tool Use} & \textbf{Human} & \textbf{Auto Researcher} & \textbf{Self-evolving} &  \textbf{Search Target} & \textbf{Domain} \\

\midrule

ChemCrow \citep{m2024augmenting}& Tool asst. & \cmark & \cmark & \cmark & \xmark & Tool & Chemistry \\
Biomni \citep{huang2025biomni}& Tool asst. & \cmark & \xmark & \xmark& \xmark & Tool & Biomedicine \\
Coscientist \citep{boiko2023autonomous} & Experiment & \cmark & \xmark & \cmark & \xmark & Experiment plan & Chemistry \\
TextGrad \citep{yuksekgonul2025optimizing} & Optimization & \xmark & \xmark & \xmark & \xmark&  Hypothesis & Multi-domain \\
AlphaEvolve \citep{novikov2025alphaevolve} & Optimization & \xmark & \xmark & \xmark& \xmark&  Hypothesis & Multi-domain \\
Virtual Lab \citep{swanson2025virtual} & Research & \cmark & \cmark & \cmark& \cmark & Hypothesis & Biomedicine \\
AutoDiscovery \citep{agarwalautodiscovery2025} & Research & \xmark & \xmark & \cmark& \cmark & Hypothesis & Multi-domain \\
AI Co-scientist \citep{gottweis2025towards} & Research & \cmark & \cmark & \xmark & \cmark&  Hypothesis & Biomedicine \\
Kosmos \citep{mitchener2025kosmos} & Research & \cmark & \xmark & \cmark & \cmark &  Hypothesis & Multi-domain \\
AI Scientist \citep{lu2024ai} & Research & \cmark & \xmark & \cmark& \cmark& Research & Computer science \\
\midrule
SAGA (Ours) & Research & \cmark & \cmark & \cmark & \cmark&  Objective \& Hypothesis & Multi-domain \\
\bottomrule
\end{tabular}
}
\caption{
Comparison of representative AI scientific discovery systems along dimensions of agent types, tool use (agents determine tools), human (human-in-the-loop, human is involved in decision processes), auto researcher (agents autonomously answer research questions without human intervention), self-evolving (agents evolve its own knowledge), target to discover, and domain generality.
}
\label{tab:ai_scientist_comparison}
\end{table*}

\subsection{Adapting \name to a New Problem}
\label{sup:adapting}

\name is designed to be domain-agnostic: adapting it to a new scientific design problem requires four steps. The following describes these steps in order, using examples from the tasks reported in this paper.

\noindent \textbf{Step 1: Specify the Discovery Goal.}
The primary input to \name is a natural-language goal that succinctly describes what is to be discovered. A well-formed goal should identify (i) the class of entities to be designed, (ii) the desired functional properties, and (iii) any hard constraints that every valid candidate must satisfy. Optionally, a separate context information block can accompany the goal to supply domain background, known scientific principles, and preferences about which types of objectives the Planner should consider.

As an example, for the antibiotic discovery task the goal reads ``\textit{Find new antibiotics that are safe and synthesizable},'' and the context information explicitly encodes structural novelty requirements, mammalian-cell safety constraints, and the chemical space from which purchasable candidates should be drawn.

Both texts are passed directly to the Planner and Analyzer at every iteration, so they serve as the persistent scientific charter against which all objectives and analysis reports are evaluated.

\noindent \textbf{Step 2: Instantiate an Optimizer.}
The optimizer constitutes the inner optimization loop. Any optimization strategies (e.g., rule- or LLM-based evolutionary algorithms, trained deep learning optimization models, simulation-coupled search) can be implemented in the optimizer, as long as it (i) accepts a population of candidates and a list of objectives with callable scoring functions, and (ii) returns an improved population with regard to the objectives.

Inside the optimizer, the scoring functions attached to each objective can be invoked directly, enabling the optimizer to evaluate candidates against them and propose better ones. Additionally, thanks to the separation of objectives and optimization strategies, the same optimizer can be reused across iterations even as the  proposes entirely different objective sets.

\noindent \textbf{Step 3: Provide Initial Objectives and Scoring Functions (Optional).}
While \name can autonomously propose and implement all objectives from scratch, human scientists often have validated computational protocols they wish to incorporate from the outset. Providing initial objectives can provide \name with value knowledge of related objectives. Human scientists can also provide their corresponding \textbf{scoring functions}, which offers two benefits: (i) the Implementer skips re-implementing these metrics and uses the human-verified code directly, and (ii) the provided implementations serve as stylistic and technical references for the Implementer when it must create new scoring functions for novel objectives proposed by the Planner. Each scoring function, either provided by human or automatically implemented by the Implementer, is packaged as a self-contained Python module. When called, it runs inside a Docker container and communicates over the standardized interface of model context protocol (MCP).

\noindent \textbf{Step 4: Configure and Run \name.}
Once the goal, optimizer, and (optionally) initial objectives are in place, the experiment is configured through a structured script that assembles all module settings and runs the SAGA outer loop. Key parameters include the number of outer-loop iterations, the LLM backbone for each module, and the autonomy level (\S\ref{sec:auto_level}).
During the run, in Co-pilot and Semi-pilot modes, the outer loop pauses during the Planner and/or the Analyzer phase and surfaces a lightweight interface through which scientists can review, edit, or approve the Planner's proposed objectives or the Analyzer's analysis before execution continues.
Upon completion, \name writes a structured result record containing every candidate evaluated across all iterations together with their objective scores, the full sequence of objective sets proposed by the Planner, analysis reports from the Analyzer, and all generated scoring-function implementations. This record enables further selection of the candidates and retrospective analysis of how the objective space evolved throughout the search.

\section{Antibiotic Discovery}
\subsection{Supplementary Figures}

\begin{figure}[H]
    \centering
    \includegraphics[trim = 0 0 0 0, clip, width=1.0\linewidth]{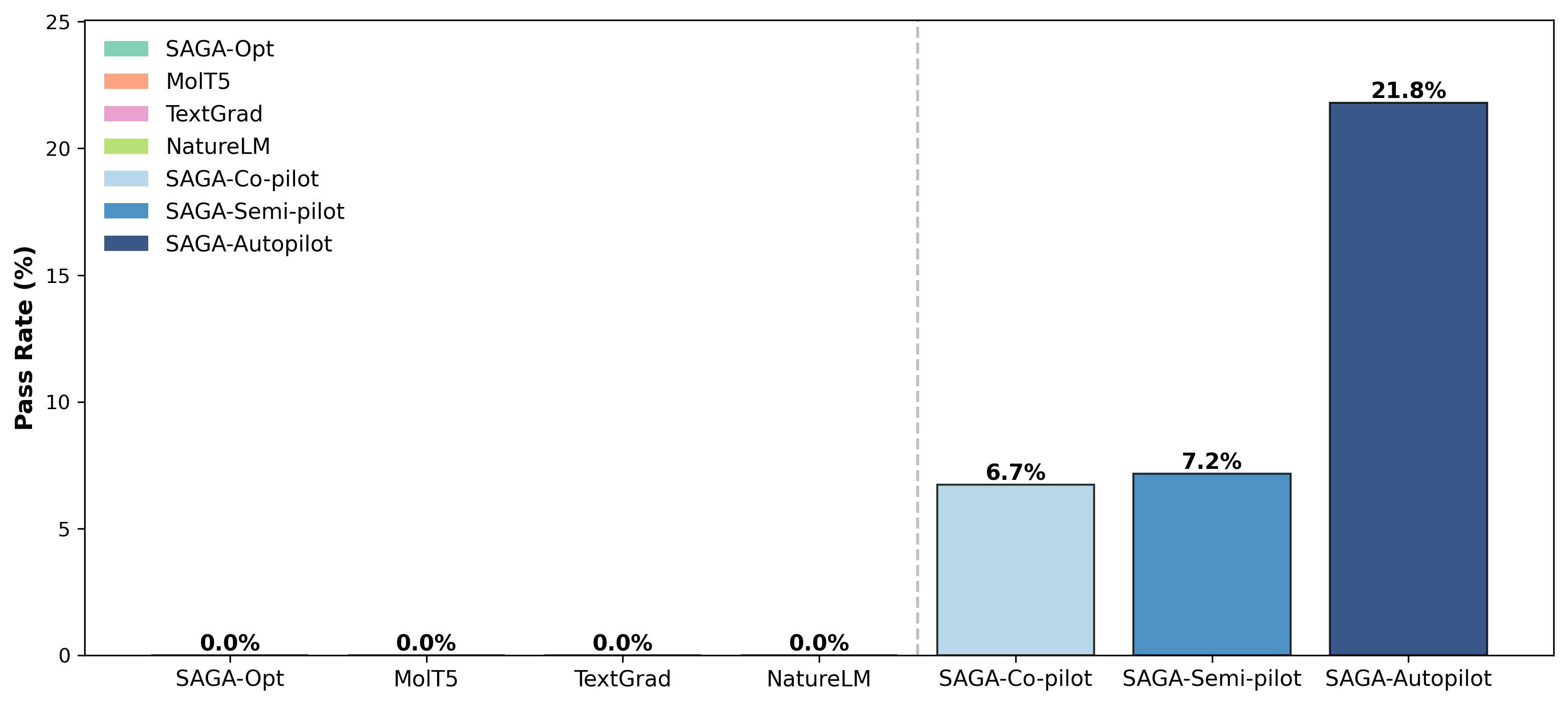}
    \caption{ Pass rate for all final molecules proposed by \name and baseline methods based on external evaluations. Definition of the property scores and pass rate can be found in \Cref{sup:expdetail_drug}.}
    \label{supfig:pass_rate_drug}
\end{figure}

\clearpage

\begin{figure}[H]
    \centering
    \includegraphics[trim = 0 0 0 0, clip, width=1.0\linewidth]{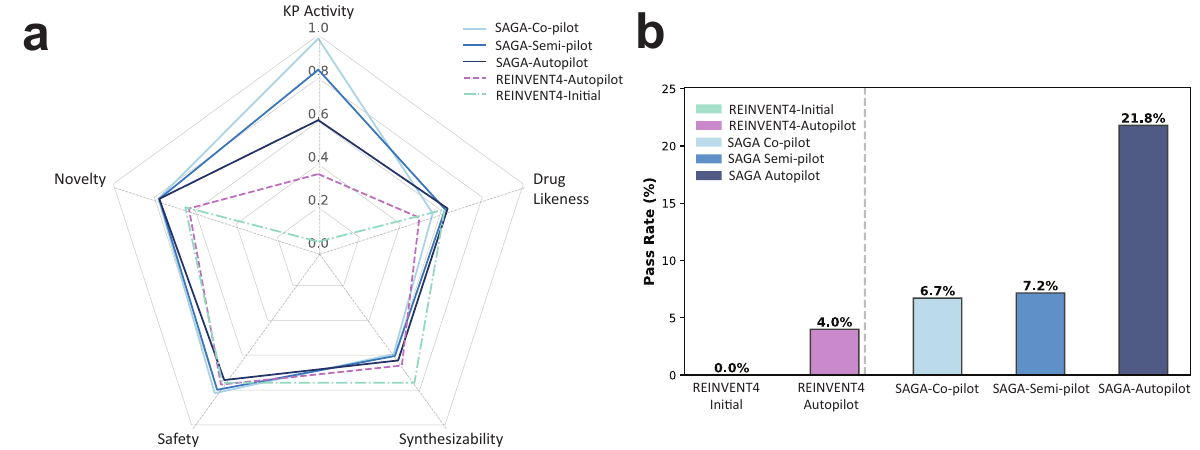}
    \caption{Additional benchmarks of antibiotic drug design. (a) Comparison of the property distributions of final candidates proposed by three \name levels and REINVENT4 \citep{loeffler2024reinvent} with initial objectives and SAGA-Autopilot final objective functions. (b) Pass rate for all final molecules proposed by \name and REINVENT4 based on external evaluations.}
    \label{supfig:reinvent4_drug_results}
\end{figure}

\begin{figure}[H]
    \centering
    \includegraphics[trim = 0 0 0 0, clip, width=1.0\linewidth]{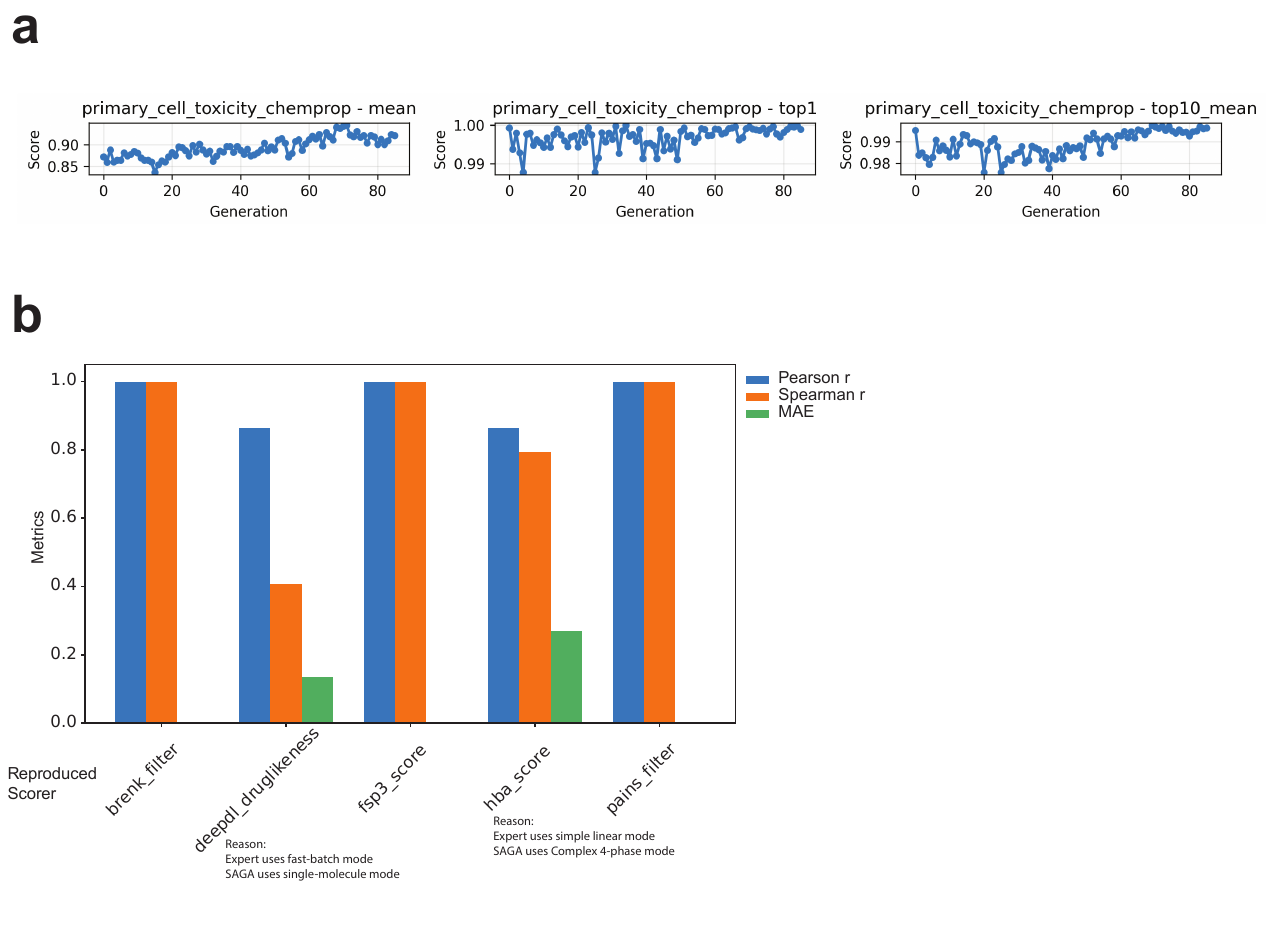}
    \caption{Ablation studies of antibiotic design. (a) The change of scores from different objectives across all epochs in the optimization process, based on the optimizer only. (b) Examining the designed scorers based on random molecules ($n=1000$). The output score correlation between human-implemented and Implementer-implemented scoring functions. We report Pearson correlation, Spearman correlation and mean squared errors.}
    \label{supfig:scorecheck_drug}
\end{figure}

\begin{figure}[H]
    \centering
    \includegraphics[trim = 0 0 0 0, clip, width=1.0\linewidth]{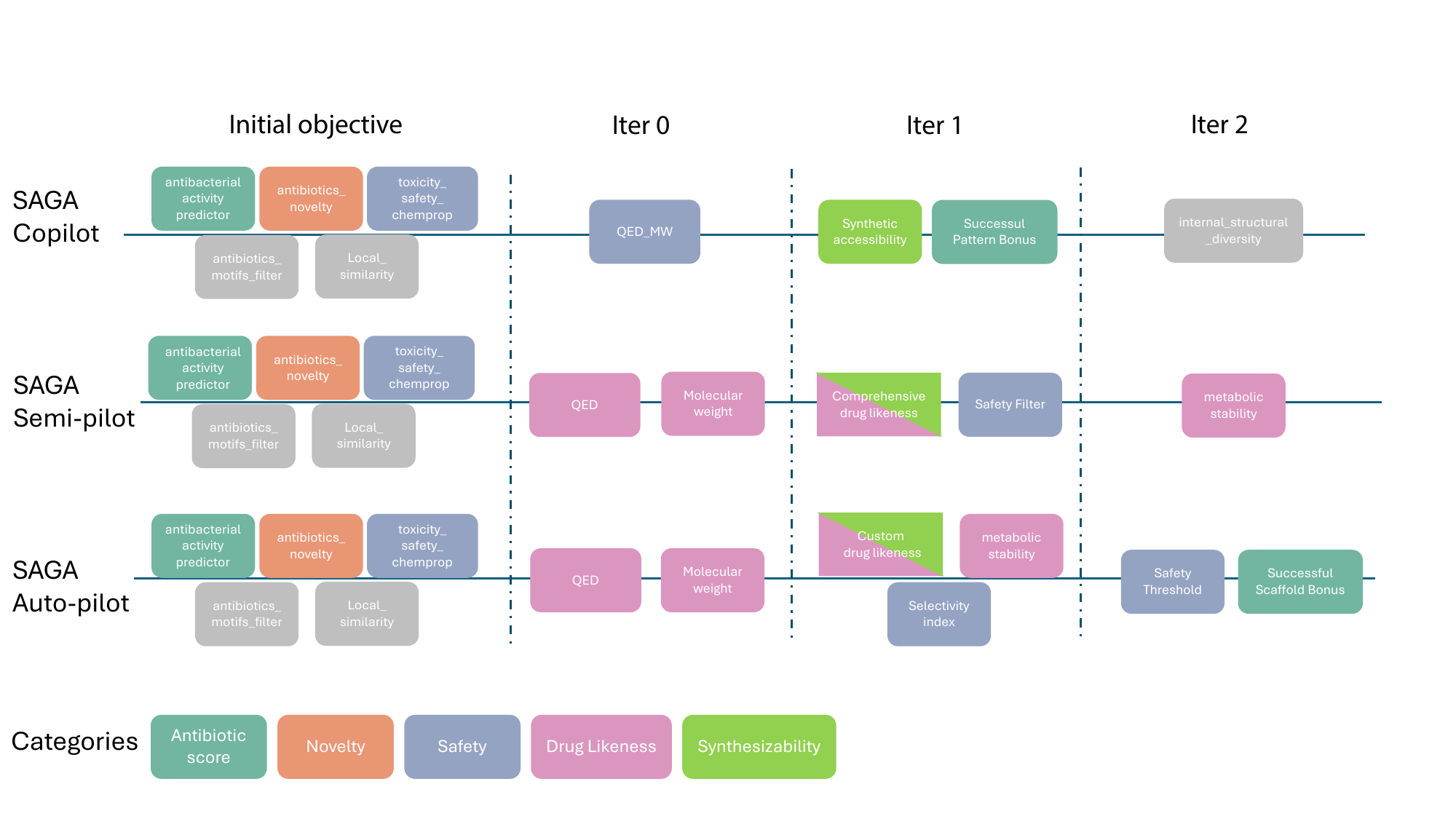}
    \caption{Objectives proposed in early-stage iterations across three modes of \name.}
    \label{supfig:scorecheck_drug_iter}
\end{figure}

\begin{figure}[H]
    \centering
    \includegraphics[trim = 0 0 0 0, clip, width=0.9\linewidth]{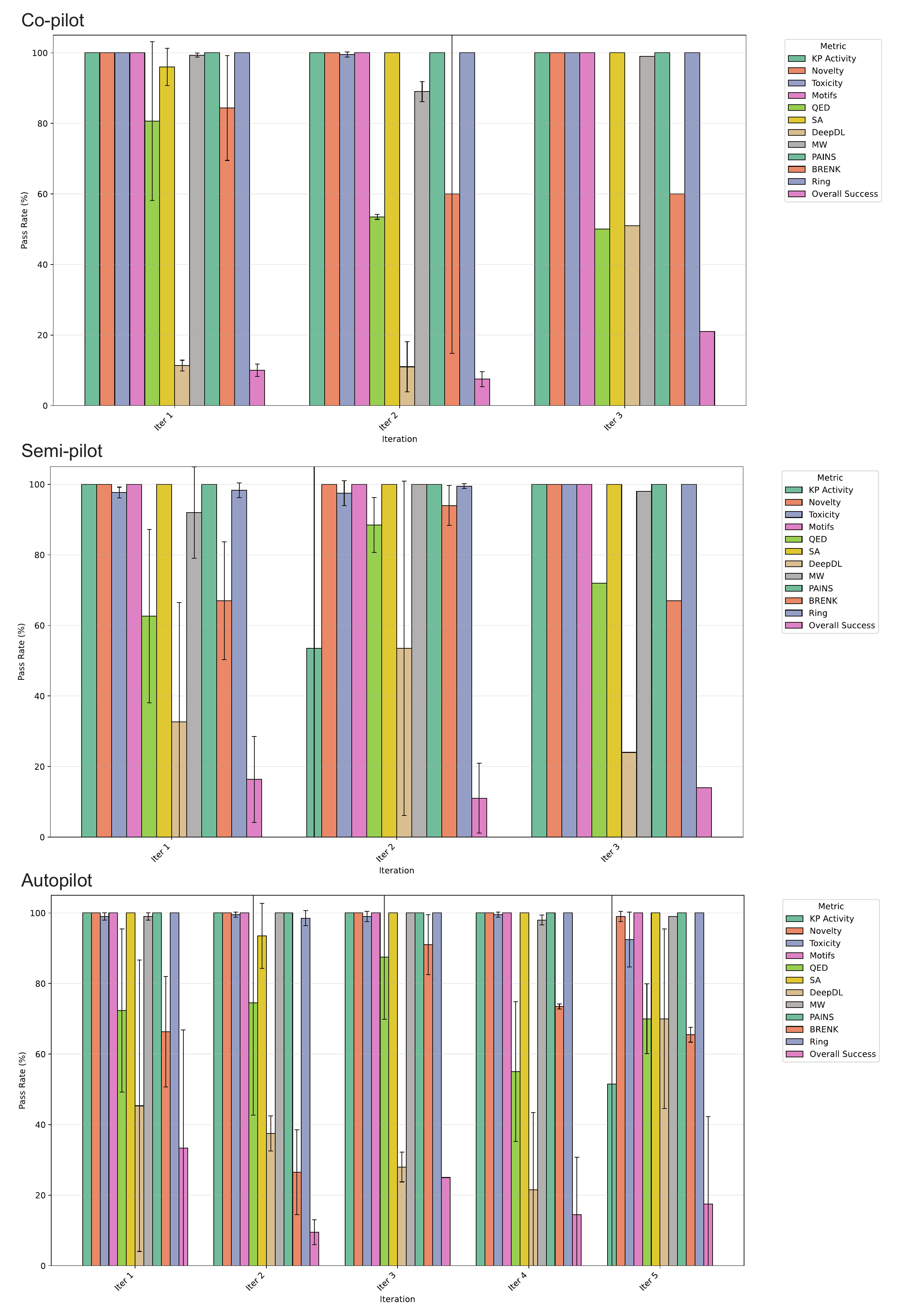}
    \caption{Pass rate (proportion of molecules passing the metric-specific threshold) changes across iterations from different modes in \name.}
    \label{supfig:scorecheck_drug_passratechange}
\end{figure}

\subsection{Experimental Setups}
\label{sup:expdetail_drug}

\subsubsection{Objectives, metrics and baselines}

Here we describe the experimental setup for antibiotic discovery targeting on Klebsiella pneumonia.

\textbf{Initial objectives}. Our initial objectives are:
\begin{itemize}
\item Maximize: \textbf{Antibiotic activity}, predicted by an ensemble of 10 Minimol \citep{minimol} models trained on an internal high-throughput screening dataset for bacteria growth inhibition. We utilize 5-fold cross validation to select the best model.
\item Maximize: \textbf{Novelty}, defined as $1 - s_{\mathrm{tan}}$, where $s_{\mathrm{tan}}$ is the Tanimoto similarity between the selected molecule and the closest known compound from a pre-defined pool with antibiotic indication. 
\item Minimize: \textbf{Toxicity}, predicted by an ensemble of Chemprop \citep{chemprop} models trained on mammalian cell toxicity data \citep{zhang2015toxicity}.
\item Minimize: \textbf{Known antibiotic motifs}, comprising six major motif classes (sulfonamides, aminoglycosides, tetracyclic\_skeletons, beta\_lactams, pyrimidine\_derivatives, quinolone), implemented via 19 SMARTS patterns covering common variations.
\item Maximize: \textbf{Synthesizability}, measured as the Tanimoto similarity to the closest compound from a subset of Enamine REAL database space \citep{shivanyuk2007enamine} to ensure the existence of a purchasable analog.
\end{itemize}

\textbf{Evaluation metrics}. We evaluate generated molecules using the following predictive scores and rule-based filters, together with fixed acceptance thresholds. Unless otherwise noted, all metrics are normalized to $[0,1]$, with higher values indicating more desirable properties.

\begin{itemize}
\item \textbf{Antibiotic activity score} is predicted by an ensemble of 10 Minimol \citep{minimol} models trained on an internal high-throughput screening dataset for bacterial inhibition. The final prediction is rescaled to be from 0 to 1. We classify a candidate as computationally active if it has a score of $0.2$ or higher. This threshold is selected based on previous experimental validation \citep{wong_discovery_2024}.
\item \textbf{Novelty score} is defined as $1 - s_{\mathrm{tan}}$, where $s_{\mathrm{tan}}$ is the Tanimoto similarity to the closest known or investigated compound with antibiotic indication. We require novelty to be greater than or equal to $0.6$.
\item \textbf{Toxicity score} is predicted by an ensemble of Chemprop \citep{chemprop} models trained on mammalian cell toxicity data. Molecules are considered acceptable if toxicity to be greater than or equal to $0.5$, indicating lower predicted cytotoxic risk.
\item \textbf{Known antibiotic motifs filter} comprises six major motif classes (sulfonamides, aminoglycosides, tetracyclic\_skeletons, beta\_lactams, pyrimidine\_derivatives, quinolone), implemented via 19 SMARTS patterns covering common variations \citep{schmidt2019comparing}. A score of $1.0$ indicates no known motifs are present.
\item \textbf{Quantitative estimate of drug likeness (QED) score} is the quantitative estimation of drug likeness \citep{bickerton2012quantifying}, combining physicochemical properties into a single score in $[0,1]$. We require QED to be greater than or equal to $0.5$.
\item \textbf{Synthetic accessibility (SA) score} estimates how easily a molecule may be synthesized \citep{chemprop}, normalized to $[0,1]$, where higher values indicate easier synthesis. We require SA to be greater than or equal to $0.5$.
\item \textbf{DeepDL drug likeness} is an unsupervised deep-learning score trained on approved drugs and normalized from its original scale to $[0,1]$. We require DeepDL to be greater than or equal to $0.3$ \citep{lee2022drug}.
\item \textbf{Molecular weight score} is a pass indicator that equals $1.0$ if molecular weight lies between $150$ and $500\,\mathrm{Da}$ and $0.0$ otherwise; we require MW to be equal to $1.0$.
\item \textbf{PAINS filter} is a binary score returning $1.0$ if no PAINS (A/B/C) alerts are present and $0.0$ otherwise; we require PAINS to be equal to $1.0$.
\item \textbf{BRENK filter} assigns $1.0$ for no structural alerts, $0.5$ for exactly one alert, and $0.0$ for two or more alerts; we require BRENK to be greater than or equal to $1.0$.
\item \textbf{Ring score} measures how common the molecule’s ring systems are relative to ChEMBL statistics \citep{gaulton2012chembl}, where $1.0$ indicates common (or no) rings and $0.0$ indicates rare or unseen ring chemotypes. We require ring\_score to be equal to $1.0$.
\end{itemize}

\noindent Summarizing the above threshold, we define \textbf{pass rate} as the proportion of molecules whose selected properties are higher than a given threshold, over the whole population (Numerical setting determined by chemists: Activity$\geq 0.2$, Novelty$\geq 0.6$, Toxicity$\geq 0.5$, Motifs$\geq 1.0$, QED$\geq 0.5$, SA$\geq 0.5$, DeepDL$\geq 0.3$, MW$\geq 0.5$, PAINS$\geq 1.0$, BRENK$\geq 1.0$, RING$\geq 1.0$. These sets of \textbf{evaluation thresholds} are also used to select final candidates for future experiments.

\textbf{Baselines}. We benchmark against four representative previous approaches spanning (1) general-purpose LLM-driven optimization, (2) generalist science language models, and (3) RL-based molecular generative models. 

\begin{itemize}
  \item \textbf{TextGrad} \citep{yuksekgonul2025optimizing} is an LLM-based optimization framework that iteratively edits molecular SMILES strings using critique-style feedback from the scoring functions and LLMs. 
  \item \textbf{NatureLM} \citep{xia2025naturelanguagemodeldeciphering} is a multi-domain, sequence-based science foundation model enabling instruction-driven generation/optimization across molecules, proteins, nucleic acids and materials.
  \item \textbf{REINVENT4} \citep{loeffler2024reinvent} is based on reinforcement-learning fine-tuning of a SMILES generator to maximize (possibly multi-objective) scoring functions, with diversity-aware goal-directed design. We utilize the latest version of this package.
  \item \textbf{MolT5} \citep{edwards2022translationmoleculesnaturallanguage} is a T5-based molecule--language translation model supporting text-to-molecule and molecule-to-text generation.
\end{itemize}

\textbf{Hyperparameters}. Antibacterial small-molecule design: For each iteration, the LLM produces 70 offspring per generation via crossover and mutates 7 of the best molecules from the current population. The LLM molecule crossover and mutation are all based on the SMILES strings of the parent molecules in a way similar to MolLEO\citep{wangefficient} via the prompt. Then 120 molecules are selected as the next population. The total oracle budget is capped at 10{,}000 evaluations. Parents are chosen via \textbf{size-3 tournament selection} (two parents per mating event), and survival selection uses a diversity-aware \texttt{diverse\_top} strategy: we perform \textbf{top-\(k\)} selection by the aggregated score and then preserve chemical diversity by enforcing a \textbf{Tanimoto similarity} constraint, retaining a candidate only if its fingerprint similarity to all already-selected survivors is below 0.4 (i.e., \(s_{\mathrm{tan}}<0.4\)). Multi-objective optimization is performed by a simple product aggregator (\texttt{simple\_product}) over all objective scores. We preserve \textbf{elitism} by retaining the top 5\% candidates each generation, where elites are selected by the antibacterial activity field. %

\textbf{Workflows}. Our three different levels (co-pilot, semi-pilot, and autopilot) follow the default setting, discussed in \Cref{sec:auto_level}. For all experiments, we have three replicates.

Moreover, to increase the diversity, we consider applying Butina cluster-based selection \citep{butina1999unsupervised} at the end of each iteration. The clustering steps include 1. Compute pairwise similarities (Calculate all pairwise Tanimoto similarities between fingerprints); 2. Choose a similarity cutoff (our current setting is 0.4); and 3. Greedy clustering (Find the molecule with the largest number of neighbors above the cutoff, perform clustering, remove all clustered molecules from the pool, and repeat until no molecules remain). The selection process means we select the top samples by aggregated scores in each cluster as final molecules.

We consider two sets of molecules for wet-lab validation. For the molecules tested with Enamine set, the experimental molecules are generated from 10 seeds of \name's Co-pilot mode. We find Enamine close neighbors (sim > 0.6) for all generated molecules passing held-out metrics. And then revaluate the held-out objectives on these Enamine close neighbor. The Enamine close neighbour molecules that pass the held-out requirements are manually selected by human expert for wet-lab validation. For the molecules tested with One Pot set, the experimental molecules are generated from three levels of \name. We find One Pot close neighbors (sim $>$ 0.5, as the molecules from One Pot are generally smaller) for all generated molecules passing held-out metrics. And then revaluate the held-out objectives on these One Pot close neighbor. The One Pot close neighbour molecules that pass the held-out requirements are manually selected by human expert for wet-lab validation.

\subsubsection{High-level Goal}
Design novel antibiotic small molecules that are highly effective antibiotics while maintaining good safety profiles and drug likeness-related properties. 

\subsubsection{Context Information}
For this task, we want to design novel antibiotics. The molecules should: 1. Show high predicted antibacterial activity. 2. Maintain low toxicity to human cells 3. Avoid problematic substructures for medicinal chemistry 4. Show structural novelty compared to existing antibiotics 5. Have good drug likeness-related properties and molecular weight for small molecule drug design. The optimizer will automatically enforce SMILES validity and length constraints, so do not propose objectives related to these. IMPORTANT SCORER REQUIREMENTS:
\begin{itemize}
\item For candidate-wise objectives: Scores must be normalized to [0, 1] range, where higher values are better (maximization direction).
\item For filter objectives: Scores must return 1.0 for pass and 0.0 for fail. Filters do not need normalization or inversion when multiplied into aggregated scores.
\end{itemize}

\subsection{Additional Experimental Results}
\label{sec:add_exp_drug}

\subsubsection{Analyses of optimization convergence in the experiment}

\textbf{\name different modes have similar performances}. We observed an interesting phenomenon in this task: the performance of the three different modes was relatively similar. To explain this reason, we further investigated the types of objectives suggested by the Analyzer for \name. According to \Cref{supfig:scorecheck_drug_iter}, the objectives proposed by different modes in each iteration are convergent to a specific list of types (e.g., QED scores occur in the iteration 1 from all modes), which can explain the similarity of performances across different modes.

\textbf{Important objectives have been proposed during the early iterations}.
We analyze all the proposed objectives from \name and find that the objectives coming out after iteration 1 cannot improve the quality of molecules obviously, shown in the pass rate comparison across different iterations (\Cref{supfig:scorecheck_drug_passratechange}). This can be explained by the relatively comprehensive information provided by the Analyzer in the analysis report. Overall, this phenomenon demonstrates that \name can efficiently identify and enhance objectives that bind to outputs while reducing the cost of running agents.

\subsubsection{Ablation studies}

\textbf{Evaluation of optimizer}. To validate whether our optimizer can work as we expect, we include an ablation study to check the performance of this agent in improving the proposed objectives via iteration. As shown in Supplementary Figure \Cref{supfig:scorecheck_drug} (a), the optimizer can improve the corresponding objective as the iteration of optimization increases. Therefore, our LLM-based optimizer can successfully improve the quality of generated molecules. 

\textbf{Evaluation of the Implementer}. We test the ability of the Implementer by validating whether it can propose similar objectives that have human implementation. Here, we focus on several important drug- and biology-related metrics, and compare the similarity between the results produced by methods from different sources with the same group of molecules as inputs (10,000 random sampled molecules from the Enamine REAL Database). According to \Cref{supfig:scorecheck_drug} (b), our implementer successfully implements five different objectives from different categories, and the comparison result shows low MSE and high correlation. Therefore, our Implementer can successfully create scorers corresponding to the assigned objectives.

\subsubsection{\name-Autopilot new proposed objectives after iteration 1.}

\textbf{Custom Drug Likeness Score.} Constrained Quantitative Estimate of Drug-likeness (QED) score with complexity penalties (value range: 0.0 to 1.0). This score starts with the standard RDKit QED calculation (composite metric considering molecular weight, LogP, HBD/HBA, PSA, rotatable bonds, aromatic rings, and structural alerts), then applies penalties for excessive molecular complexity that degrades drug-likeness: (1) Rotatable bonds penalty: if n\_rotatable\_bonds > 6, apply penalty of 0.9\verb|^|(n\_rotatable\_bonds - 6); (2) Fraction Csp3 penalty: if frac\_Csp3 < 0.45, apply penalty of 0.95\verb|^|((0.45 - frac\_Csp3) × 20); (3) Molecular weight soft penalty: if MW > 400, apply penalty of 0.98\verb|^|((MW - 400) / 10). Final score = base\_QED × rotatable\_penalty × csp3\_penalty × mw\_penalty, normalized to [0, 1]. High scores (>0.7) indicate excellent drug-like properties with appropriate complexity, while low scores (<0.5) suggest poor drug-likeness or excessive complexity. This addresses the -0.054 QED decline and negative correlation with activity (r=-0.370) observed in iteration 1 by explicitly penalizing the complexity increases (mean 6.5 rotatable bonds, only 44.2\% meeting Csp3 threshold) that drove QED degradation.

\textbf{Metabolic Stability Score.} Metabolic stability score based on structural alerts (value range: 0.0 to 1.0). This score identifies and penalizes structural features associated with rapid metabolism or metabolic liabilities: (1) Primary aliphatic amines (-NH2 attached to aliphatic carbon): penalty 0.15 per occurrence (susceptible to oxidative deamination and conjugation); (2) Morpholine rings: penalty 0.12 per occurrence (metabolically labile via N-oxidation); (3) Unprotected phenols: penalty 0.18 per occurrence (rapid glucuronidation); (4) Aliphatic aldehydes/ketones: penalty 0.10 per occurrence (carbonyl reduction). Score = max(0.0, 1.0 - sum\_of\_penalties), normalized to [0, 1]. High scores (>0.8) indicate good predicted metabolic stability with few labile groups, while low scores (<0.5) suggest multiple metabolic soft spots that could lead to rapid clearance. This addresses the observation that 80\% of high-activity molecules in iteration 1 contain primary amines and 40\% contain morpholine rings, both metabolically labile groups. Implementation uses SMARTS patterns: primary amine '[NH2][CX4]', morpholine 'C1COCCN1', phenol '[OH]c', aliphatic carbonyl '[CX3](=O)[CX4]'.

\section{Nanobody Design}
\subsection{Supplementary Figures}

\begin{figure}[H]
    \centering
    \includegraphics[trim = 0 0 0 0, clip, width=1.0\linewidth]{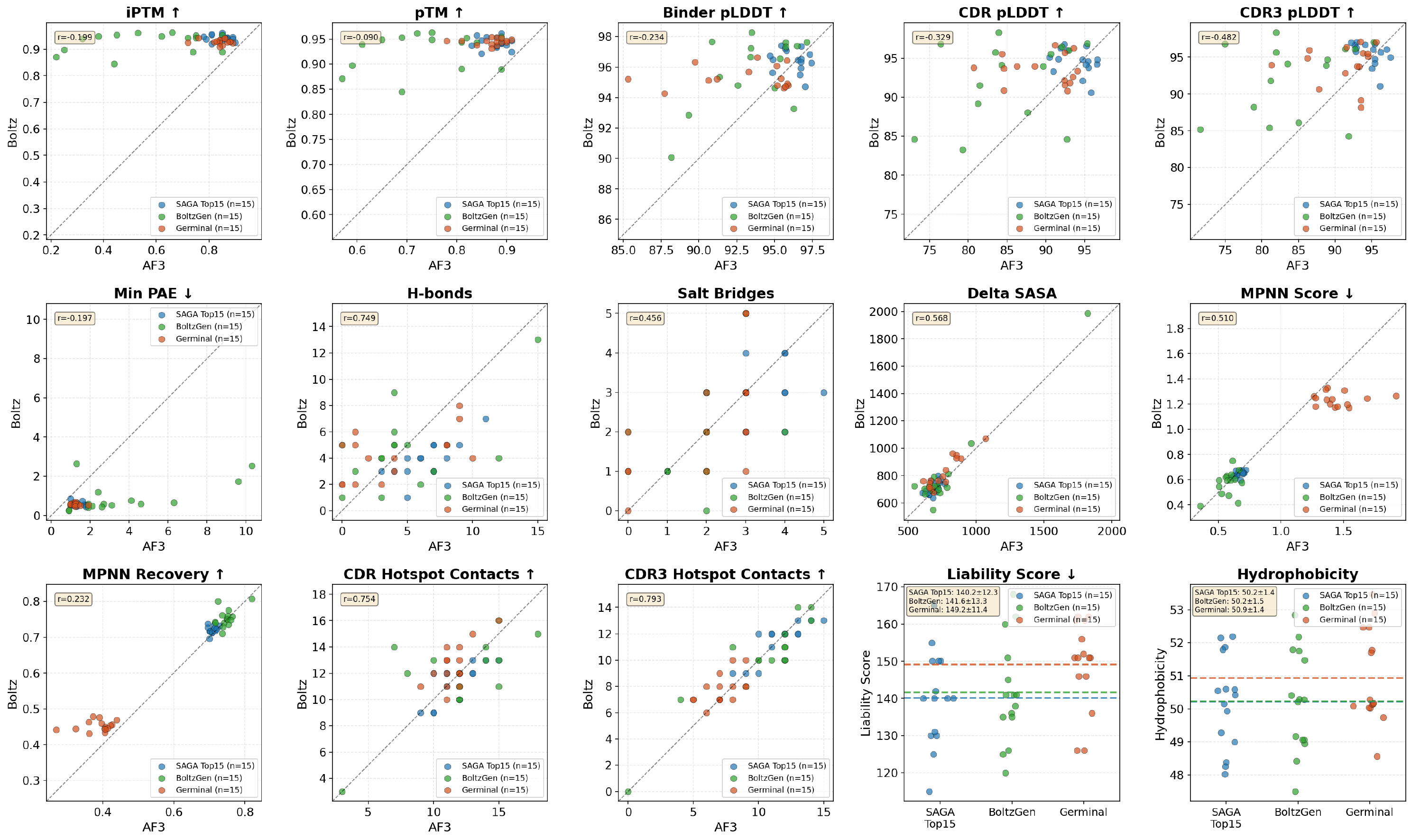}
    \caption{Comprehensive metric comparison between \name BoltzGen and Germinal nanobody candidates across structure predictors. Scatter plots compare AF3 and Boltz2 predictions for the top 15 nanobodies designed by \name the 15 PD-L1 nanobodies reported by BoltzGen and the 15 top designs selected from Germinal. Each point represents a single candidate evaluated under both structure predictors. Metrics include predicted binding confidence (ipTM, pTM), stability (binder pLDDT, CDR pLDDT, CDR3 pLDDT), interface confidence (minimum PAE), physics-based scores (hydrogen bonds, salt bridges, $\Delta$SASA), sequence structure compatibility (ProteinMPNN score and recovery), epitope contacts (CDR hotspot contacts and CDR3 hotspot contacts), and developability-related properties (liability score and hydrophobicity). The dashed diagonal indicates parity between AF3 and Boltz2 predictions. Pearson correlation coefficients are reported in each panel. Box plots summarize liability and hydrophobicity distributions across methods. }
    \label{supfig:nanobody_metric_comparison}
\end{figure}

\clearpage

\begin{figure}[H]
    \centering
    \includegraphics[width=0.9\linewidth]{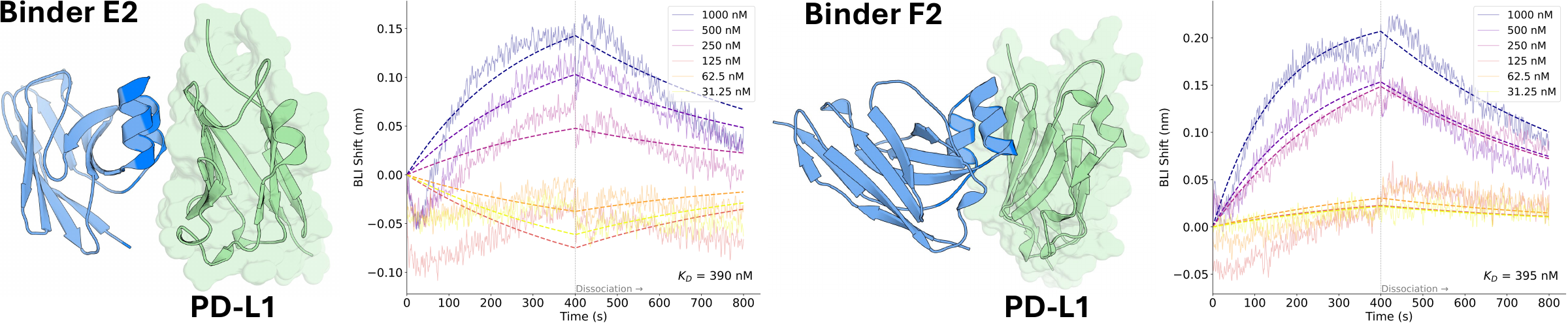}
    \caption{Additional validated SAGA-designed PD-L1 binders. Experimental validation shows that three nanobodies designed by SAGA bind PD-L1 with dissociation constants $K_D$ ranging from 300 to 400 nM. Beyond the top-performing candidate shown in \Cref{fig:nanobody_design_main}, predicted structures by AlphaFold3 and corresponding biolayer interferometry (BLI) binding traces for the remaining two binders are shown here, with measured $K_D$ values of 390 nM and 395 nM, respectively.}
    \label{supfig:nanobody_sp_structures}
\end{figure}

\begin{figure}[H]
    \centering
    \includegraphics[width=1.0\linewidth]{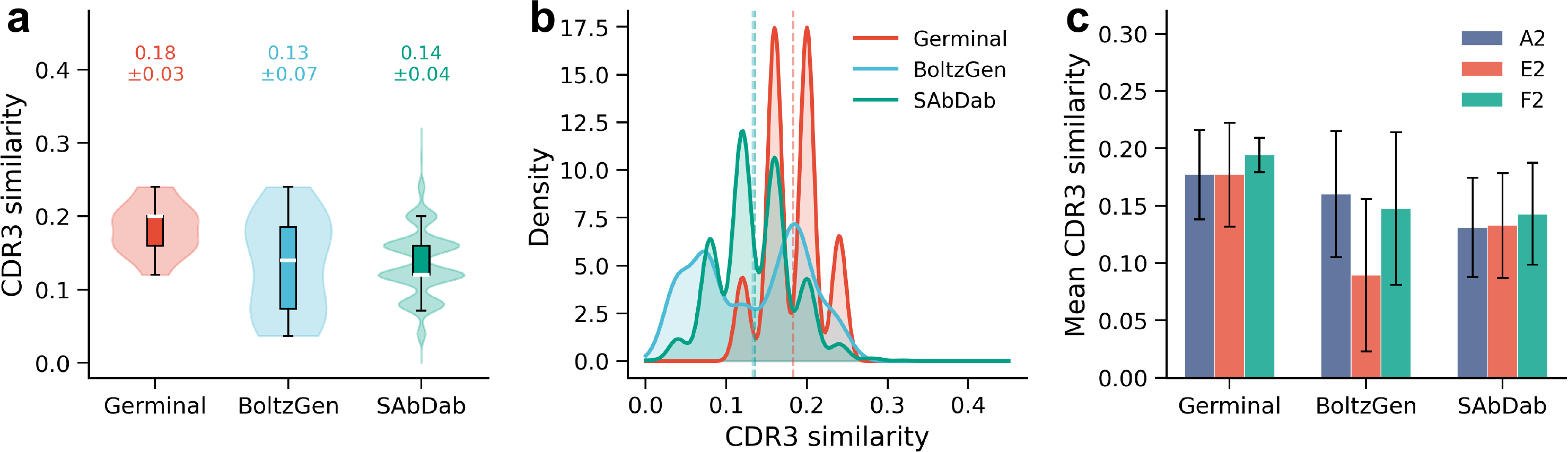}
    \caption{CDR3 sequence novelty of \name-designed nanobody binders. (a) Violin and box plots of CDR3 similarity (normalized Levenshtein) between three experimentally validated SAGA binders and reference sequences from Germinal (n=7), BoltzGen (n=6), and SAbDab (n=1,145). (b) Kernel density estimates of CDR3 similarity distributions for each reference group. (c) Per-binder mean CDR3 similarity (±s.d.) across reference groups (A2, E2 and F2 denote individual SAGA binders).}
    \label{supfig:nanobody_sp_novelty}
\end{figure}

\begin{figure}[H]
    \centering
    \includegraphics[width=1.0\linewidth]{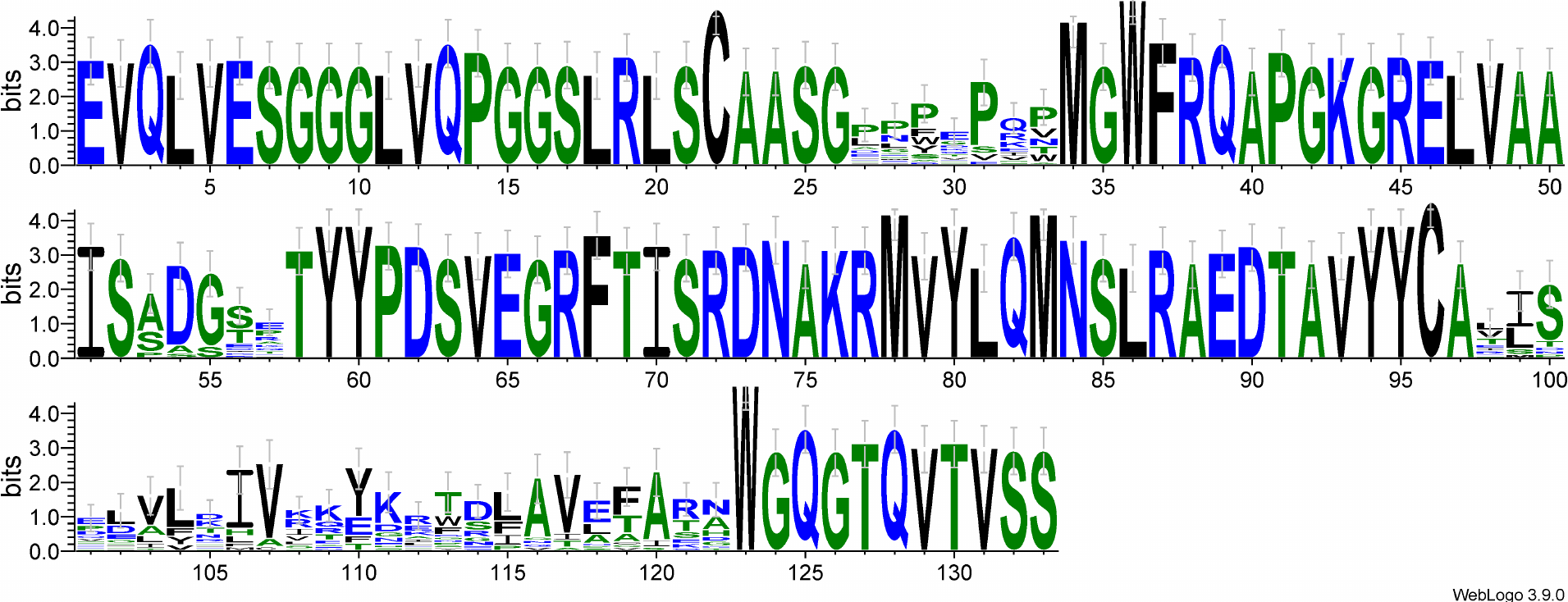}
    \caption{Sequence logo of top \name designed nanobodies, highlighting conserved framework positions and diversified CDR regions.}
    \label{supfig:nanobody_sp_logo}
\end{figure}

\subsection{Experimental Setups}
\label{sup:expdetail_nanobody}

\subsubsection{Objectives, metrics, and baselines}

Here we describe the experimental setup for \textit{de novo} nanobody design targeting PD-L1 (Programmed Death-Ligand 1).

\textbf{Initial objectives}. Our initial objectives are:
\begin{itemize}
\item Maximize \textbf{protein iPTM}, the interface predicted TM-score from structure prediction, indicating binding interface quality. Range $[0,1]$, with values $>0.6$ considered excellent.
\item Maximize \textbf{pTM}, the overall predicted TM-score indicating global structure quality. Range $[0,1]$, with values $>0.8$ considered excellent.
\item Minimize \textbf{min design-to-target PAE}, the minimum predicted aligned error between the nanobody and target at the interface, indicating prediction confidence. Lower values, typically $<10$,\AA, suggest higher confidence.
\item Maximize \textbf{binder pLDDT}, the per-residue confidence score averaged over the nanobody, measuring structural stability and fold reliability.
\item Maximize \textbf{interface hydrogen bonds (PLIP)}, the number of hydrogen bonds at the binding interface computed on predicted structures using PLIP \citep{salentin2015plip}.
\item Maximize \textbf{interface salt bridges (PLIP)}, the number of salt bridges at the binding interface computed on predicted structures.
\item Maximize \textbf{delta SASA}, the change in solvent-accessible surface area upon complex formation, indicating buried interface area.
\item Maximize \textbf{CDR–epitope contacts}, defined as the number of CDR residues with any heavy atom within $6\text{\AA}$ of predefined epitope hotspot residues on the target, capturing the extent of engagement with functionally important binding sites.
\item Minimize \textbf{ProteinMPNN score}, which evaluates sequence likelihood conditioned on the backbone structure, serving as a measure of sequence–structure compatibility.
\item Maximize \textbf{ProteinMPNN recovery}, defined as the fraction of residues recovered by ProteinMPNN redesign, providing an additional signal of structural plausibility.
\item Minimize \textbf{liability score}, a composite score penalizing known sequence liabilities including deamidation sites (NG, NS, NT motifs), oxidation-prone residues (exposed M, W), isomerization sites (DG, DS), and aggregation motifs.
\end{itemize}

\textbf{Evaluation metrics}. We evaluate generated nanobody sequences using structure prediction-based metrics and sequence-based filters. Structure predictions are performed using both AlphaFold3 \citep{abramson2024accurate} and Boltz2 \citep{passaro2025boltz}. Unless otherwise noted, metrics are reported on their natural scales, and we report values computed under both predictors.
\begin{itemize}
\item \textbf{Protein iPTM} measures predicted binding interface quality. Range $[0,1]$. 
\item \textbf{pTM} measures global structure quality. Range $[0,1]$. 
\item \textbf{Binder pLDDT} is the average pLDDT for the nanobody chain. Range $[0,100]$. 
\item \textbf{CDR pLDDT} is the average pLDDT over all CDR residues. Range $[0,100]$. 
\item \textbf{CDR3 pLDDT} is the average pLDDT for the CDR3 loop. Range $[0,100]$.
\item \textbf{Min PAE} is the minimum predicted aligned error between nanobody CDR residues and target epitope residues. 
\item \textbf{Hydrogen bonds} counts interface hydrogen bonds computed with PLIP \citep{salentin2015plip}. 
\item \textbf{Salt bridges} counts interface salt bridges. 
\item \textbf{Delta SASA} is the change in solvent-accessible surface area upon binding in $\text{\AA}^2$, computed as $\text{SASA}*{\text{complex}} - \text{SASA}*{\text{nanobody}} - \text{SASA}_{\text{target}}$.
\item \textbf{MPNN score} is the ProteinMPNN \citep{dauparas2022robust} negative log-likelihood computed on the predicted structure. 
\item \textbf{MPNN expected recovery} is the expected recovery rate from ProteinMPNN. Range $[0,1]$. 
\item \textbf{CDR-epitope contacts} counts CDR residues within 6,\AA\ of predefined target epitope hotspot residues. 
\item \textbf{CDR3-epitope contacts} counts CDR3 residues contacting target hotspots.
\item \textbf{Liability score} aggregates sequence liability penalties. Range $[0,300+]$. 
\end{itemize}

\textbf{Radar chart dimensions.} To provide an aggregated comparison across methods, we define eight evaluation dimensions. For each dimension, constituent metrics are first normalized to $[0,1]$ (inverting metrics where lower is better), then averaged across metrics within each dimension, and finally averaged across sequences within each method. The dimensions and their constituent metrics are:

\begin{itemize}
\item \textbf{pTM/ipTM}: Protein iPTM and pTM (both already in $[0,1]$; higher is better).
\item \textbf{Stability}: Binder pLDDT and CDR pLDDT (divided by 100 to normalize to $[0,1]$; higher is better).
\item \textbf{pAE scores}: Min PAE (divided by 32 and inverted, since lower PAE indicates better interface confidence).
\item \textbf{Physics-based Scores}: Hydrogen bonds (normalized to $[0,12]$), salt bridges (normalized to $[0,8]$), and Delta SASA (normalized to $[0,1200]$~\AA$^2$); all higher is better.
\item \textbf{Sequence-Structure Compatibility}: MPNN score (divided by $\ln 20$ and inverted, since lower negative log-likelihood indicates better compatibility) and MPNN expected recovery (already in $[0,1]$; higher is better).
\item \textbf{Epitope Contacts}: CDR-hotspot contacts (normalized to $[0,22]$; higher is better).
\item \textbf{Developability}: Liability score (raw range $[0, 300+]$; we linearly map the effective range $[100, 250]$ to $[0,1]$ with clipping, then invert, since lower liability indicates better developability. Scores $\leq 100$ receive the maximum dimension score of 1.0, and scores $\geq 250$ receive $\leq 0.0$).
\item \textbf{Computational Efficiency}: Defined as the total number of structure predictions (forward passes through structure prediction models) required by each method. We compute the budget score as $\text{score} = B_{\min} / B$, where $B$ is the method's budget and $B_{\min} = 12{,}000$ is the minimum budget across all compared methods. This yields a score in $(0, 1]$ where 1 indicates the most computationally efficient method. The budgets are: SAGA ($B = 12{,}000$, score $= 1.00$), Germinal ($B = 21{,}120$, score $\approx 0.57$), and BoltzGen ($B = 60{,}000$, score $= 0.20$).
\end{itemize}

\textbf{Baselines}. We benchmark against BoltzGen and Germinal to evaluate \name across distinct design paradigms. BoltzGen serves as an all-atom generative pipeline that integrates structural generation with sequence redesign and structural refinement. We also include Germinal, which functions as a nanobody-specific version of the BindCraft\citep{pacesa2025one}, utilizing modular pipelines for template selection and CDR optimization. These baselines represent the state-of-the-art in current community binder design challenges and embody diverse and highly representative design methodologies which provide a rigorous standard for evaluating our approach.

\textbf{Hyperparameters}. For each iteration, the genetic algorithm maintains a population of $100$ nanobody sequences. In each generation, $70$ offspring are produced via crossover and $30$ via mutation. Crossover uses a hybrid strategy: $40\%$ CDR-swap crossover, $40\%$ single-point crossover, and $20\%$ uniform crossover within CDRs. Mutation applies a random CDR mutation. Parents are chosen via size-15 tournament selection. Survival selection preserves the top $15\%$ as elites and enforces a CDR sequence similarity constraint, retaining a candidate only if its CDR identity to all already-selected survivors is below $0.5$, computed on the concatenated CDR1, CDR2, and CDR3 sequence. The optimization runs for up to $10$ generations per iteration, with early stopping if improvement plateaus for $5$ generations.

\textbf{Workflows}. We evaluate \name under three levels of human-in-the-loop interaction that reflect increasing degrees of agent autonomy while keeping the same initial objective specification.

\textit{Level 1 (\name Co-pilot)}. Starting from an LLM-generated initial population optimized under the baseline objectives in the first iteration, the human introduces a CDR3 alpha-helix structural constraint in the second iteration, requiring all designed nanobodies to exhibit a proper alpha helix within the CDR3 loop as determined by DSSP secondary-structure assignment on predicted structures. In the third iteration, observing that helix formation alone does not guarantee epitope engagement, the human further introduces a CDR3-hotspot contact objective, encouraging direct contacts between the CDR3 helix region and predefined target epitope residues.

\textit{Level 2 (\name Semi-pilot)}. Starting from an LLM-generated initial population optimized under baseline objectives during the first iteration, the human provides high-level feedback that structural confidence is low while requesting improvement. In the second iteration, the agent operationalizes this feedback by adding alpha-helix objectives and structural-weight refinement for the full binder and CDR regions. Following the second iteration the human observes that many candidates still lack sufficient binder-target contacts and requests stronger epitope engagement. In the third iteration, the agent responds by upweighting the ipTM objective and adding CDR3-hotspot contact objectives to ensure strong epitope engagement.

\textit{Level 3 (\name Autopilot)}. Starting from an LLM-generated initial population optimized under the baseline objectives in the first iteration, the agent autonomously proposes and implements additional objectives in the second and third iterations without any human feedback, guided only by the observed optimization trajectory and population-level trade-offs.

\subsubsection{High-level Goal}
\label{sup:highlevel_nanobody}

Design high-affinity nanobodies that bind to PD-L1 for therapeutic applications in cancer immunotherapy.

\subsubsection{Context Information}
\label{sup:context_nanobody}

\textbf{Background}. PD-L1 is a transmembrane protein that plays a major role in suppressing the adaptive immune system. PD-L1 binds to its receptor PD-1 found on activated T cells, B cells, and myeloid cells.

\textbf{Target details}. PD-L1 is overexpressed in many tumor types and contributes to immune evasion. The PD-1 and PD-L1 pathway is a critical immune checkpoint exploited by tumors.

\textbf{Optimization Focus}:
\begin{itemize}
\item Maximize binding confidence with high \textbf{iPTM} and \textbf{pTM}, and low \textbf{min PAE}
\item Ensure good interface quality with high \textbf{hydrogen bond} and \textbf{salt bridge} counts
\item Maximize buried interface area with high \textbf{$\Delta$SASA}
\item Ensure structural stability with high \textbf{binder pLDDT}
\item Promote epitope engagement with high \textbf{CDR-epitope contacts} and \textbf{CDR3-epitope contacts}
\item Maintain sequence–structure compatibility with low \textbf{ProteinMPNN score} and high \textbf{ProteinMPNN expected recovery}
\item Maintain developability with low \textbf{liability score}
\end{itemize}                                                                                                                                                                      
\textbf{IMPORTANT:} The objectives protein\_iptm, ptm, min\_pae, plip\_hbonds, plip\_saltbridges, delta\_sasa, hydrophobicity, and liability\_score MUST always be included. This experiment uses NON-LLM crossover and mutation operators. The Planner can adjust objective weights and propose new objectives, but the genetic operators remain genetic algorithmic.

\section{Functional DNA Sequence Design}
\subsection{Supplementary Figures}

\begin{figure}[H]
    \centering
    \includegraphics[trim = 0 0 0 0, clip, width=1.0\linewidth]{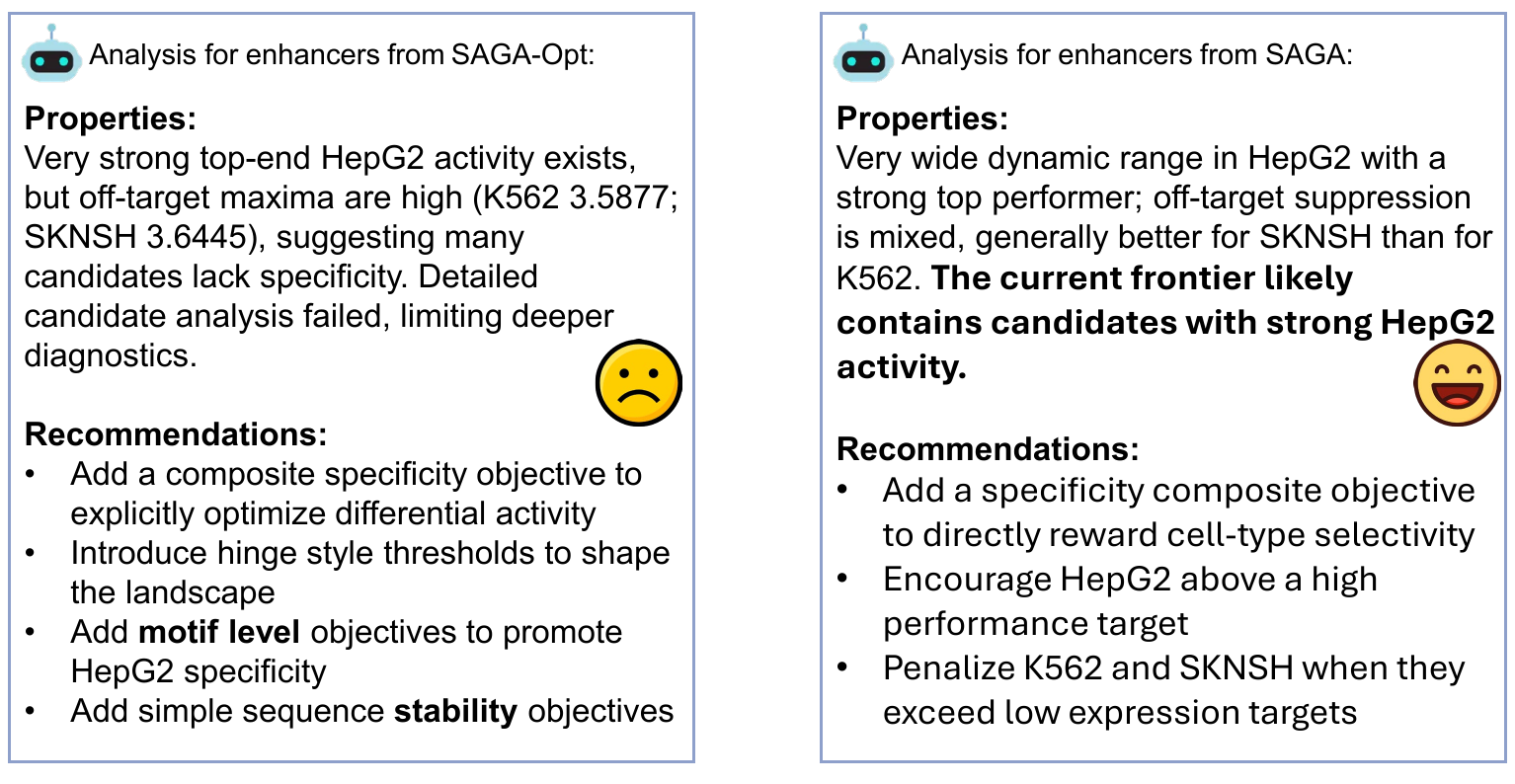}
    \caption{Illustration of analysis report generated by the analysis agent for two iterations.}
    \label{supfig:enhancer_analysis_agent}
\end{figure}

\clearpage

\begin{figure}[H]
    \centering
    \includegraphics[trim = 0 0 0 0, clip, width=1.0\linewidth]{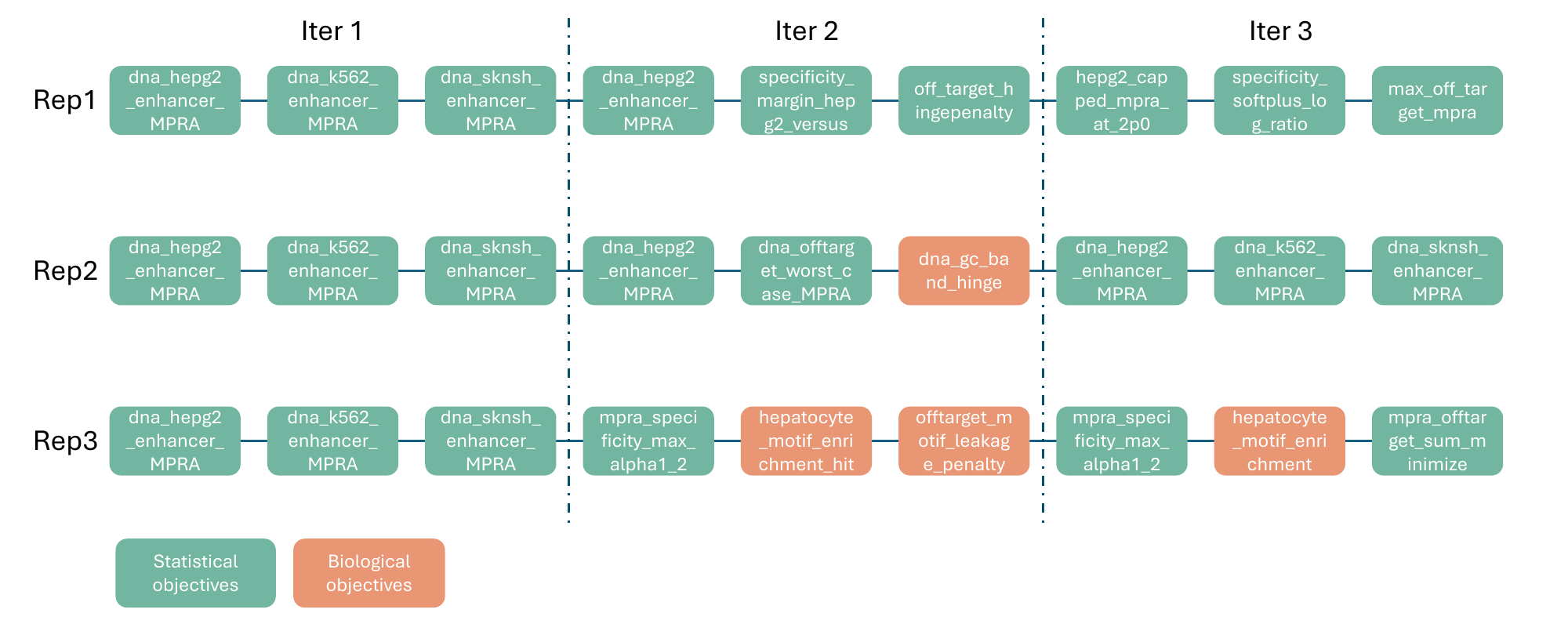}
    \caption{Objectives proposed in each iteration across three runs of the autopilot mode of \name.}
    \label{supfig:level3steps}
\end{figure}

\clearpage

\begin{figure}[H]
    \centering
    \includegraphics[trim = 0 0 0 0, clip, width=1.0\linewidth]{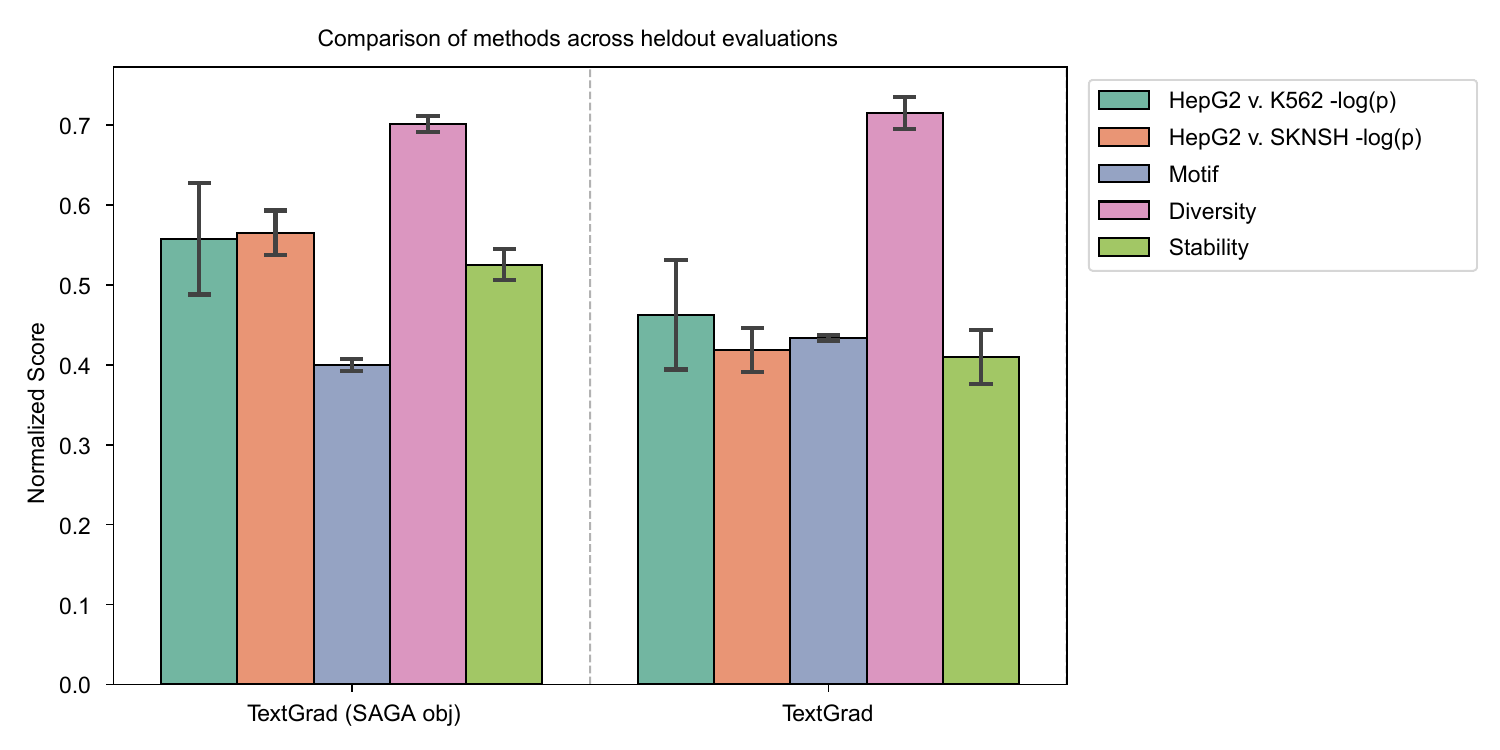}
    \caption{Comparison between TextGrad and the version trained with objectives proposed by \name.}
    \label{supfig:saga improvement}
\end{figure}

\clearpage
\begin{figure}[H]
    \centering
    \includegraphics[trim = 0 0 0 0, clip, width=1.0\linewidth]{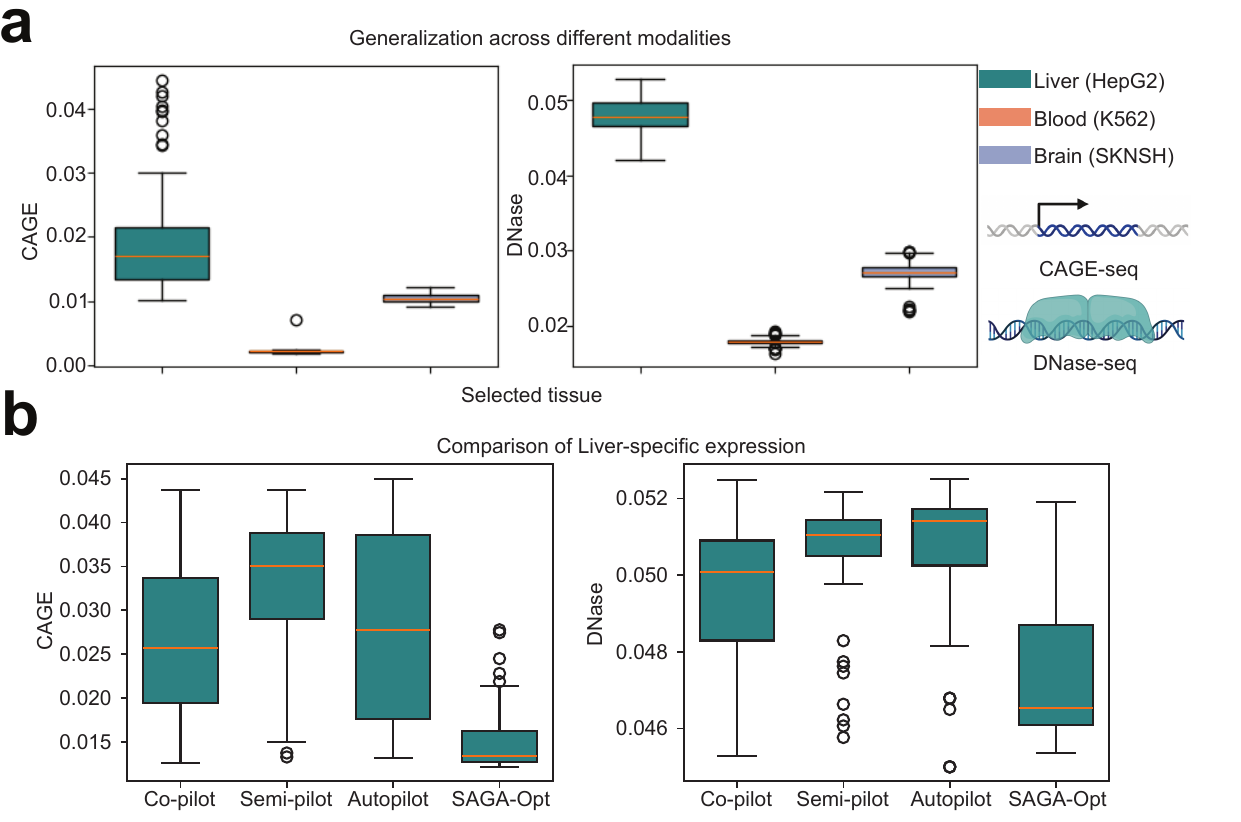}
    \caption{Cross-modality prediction analysis. (a) Consistency between the MPRA and other sequence expression measurement, including CAGE-seq and DNase-seq. The prediction is performed with Enformer. (b) Comparison of predicted CAGE-seq and DNase-seq expression levels from the enhancers generated by two different methods.}
    \label{supfig:enhancer_bio_analysis}
\end{figure}

\begin{figure}[H]
    \centering
    \includegraphics[trim = 0 0 0 0, clip, width=1.0\linewidth]{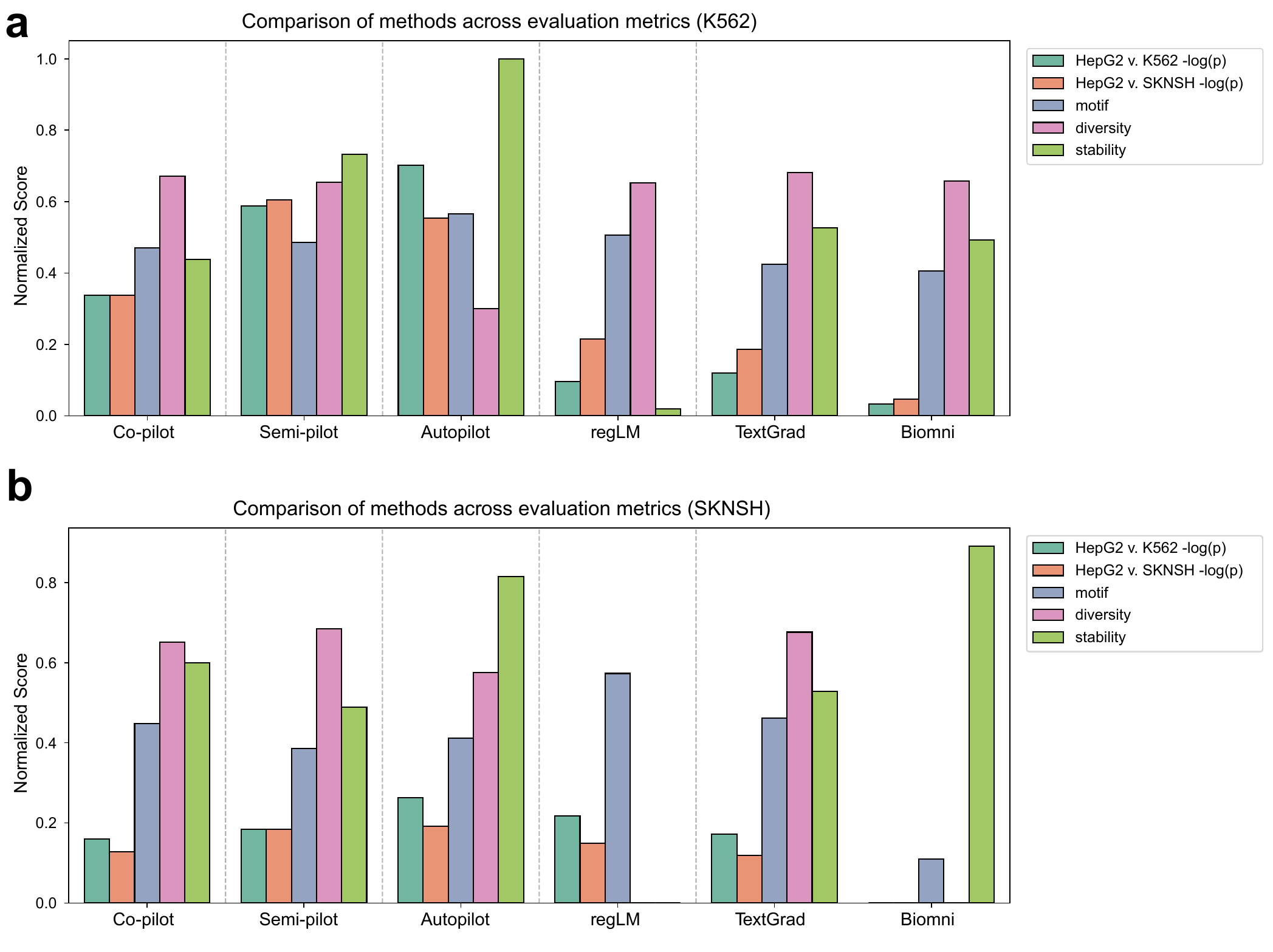}
    \caption{Comparison between \name and other baselines for (a) K562-specific enhancer design and (b) SKNSH-specific enhancer design.}
    \label{supfig:enhancer_resttwo}
\end{figure}

\begin{figure}[H]
    \centering
    \includegraphics[trim = 0 0 0 0, clip, width=1.0\linewidth]{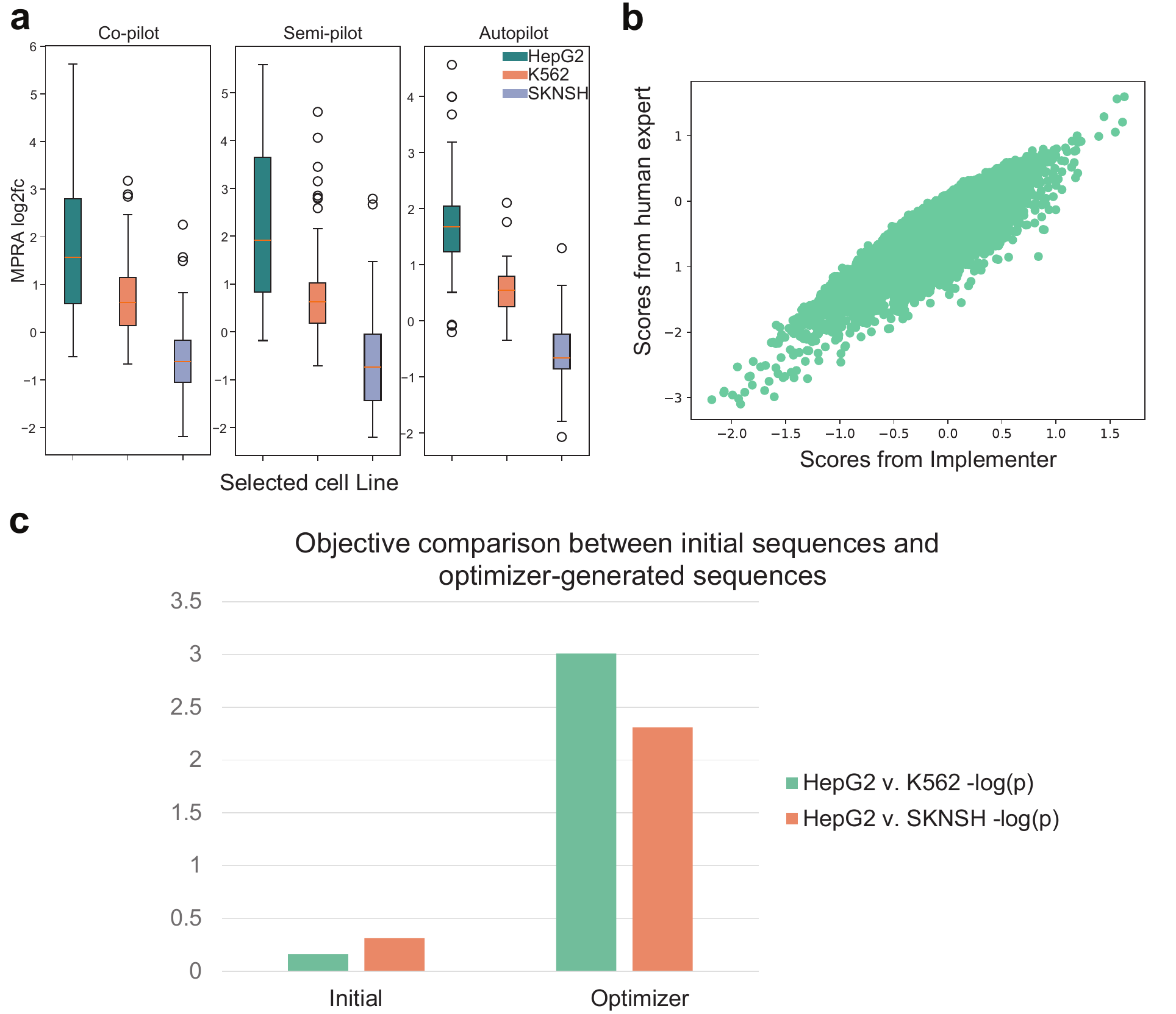}
    \caption{Ablation studies for functional DNA sequence design. (a) MPRA expression of generated DNA sequences across different levels in \name (b) Correlation between human-proposed HepG2 scoring function and Implementer-proposed HepG2 scoring function. The difference comes from the scaling constant. (c) Validation of using LLMs and evolutionary algorithm as the optimizer for optimizing the cell-type-specific expression levels.}
    \label{supfig:dna_merged_abla}
\end{figure}

\begin{figure}[H]
    \centering
    \includegraphics[trim = 0 0 0 0, clip, width=1.0\linewidth]{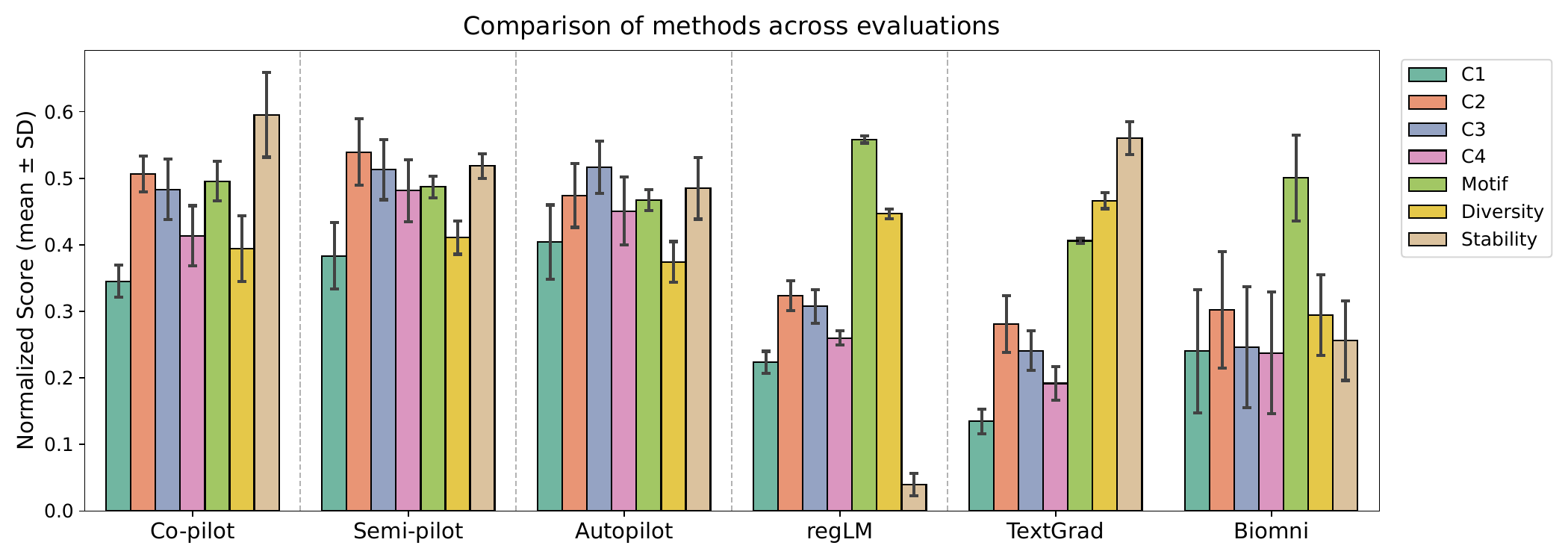}
    \caption{Comparison between \name and other baselines for promoter design across five different cell lines, and the variation is computed across five cell lines. The held-out metrics are consistent, while C1-C4 represents the -log(p-value) test result between the MPRA expression levels of the targeted cell line and the rest of cell lines.}
    \label{supfig:promoter design}
\end{figure}

\subsection{Experimental Setups}
\label{sup:expdetail dna}

\subsubsection{Objectives, metrics, and baselines}

Here we describe the experimental setup for HepG2-specific enhancer design.

\textbf{Initial objectives}. Our initial objectives are:
\begin{itemize}
\item Maximize: HepG2-specific MPRA prediction score.
\item Minimize: K562-specific MPRA prediction score.
\item Minimize: SKNSH-specific MPRA prediction score.
\end{itemize}
To predict MPRA activity from generated DNA sequences, we fine-tune Enformer-based predictors using published MPRA datasets \citep{avsec2021effective}, with training, validation, and test splits performed at the chromosome level to prevent data leakage. The datasets used for enhancer design include MPRA measurements from three cell lines (HepG2, K562, and SKNSH) \citep{gosai2024machine}, while those for promoter design comprise five cell lines (K562, HepG2, SKNSH, GM12878, and A549) \citep{reddy2024strategies}. In both cases, the raw MPRA measurements are transformed into log-fold-change values prior to model training and evaluation.

\textbf{Evaluation metrics}. We adopt evaluation metrics that are standard in prior domain-specific studies on functional DNA sequence design and enhancer modeling \citep{gosai2024machine,dasilva2024dna,chen2025ctrl,lal2024designing}. Importantly, all metrics are computed on held-out predictions and are not directly optimized during generation, ensuring a fair comparison across methods. Together, these metrics capture complementary aspects of statistical expression specificity, biological plausibility, and generative quality.

\begin{itemize}

\item \textbf{Specificity1 (HepG2 vs. K562 MPRA, $-\log p$)}.
We quantify HepG2 specificity by applying a one-sided Wilcoxon rank-sum test between predicted HepG2 MPRA activities and K562 MPRA activities across the generated sequences. This choice mirrors experimental practice in MPRA studies, where statistical significance is used to assess whether a candidate enhancer shows cell-type-specific activity rather than global activation. Compared to simple score differences, a non-parametric test is robust to distributional shifts and outliers in model predictions, making it well suited for large, heterogeneous sequence sets. The metric ranges from $0$ to $\infty$, with larger values indicating stronger and more statistically robust HepG2 specificity.

\item \textbf{Specificity2 (HepG2 vs. SKNSH MPRA, $-\log p$)}.
Analogous to Specificity1, we evaluate specificity against SKNSH, a neuronal cell line that is biologically distant from hepatocytes. Including a second, orthogonal off-target cell type ensures that designed enhancers are not merely suppressing hematopoietic programs, but instead exhibit broader hepatocyte-specific regulation. This dual-contrast design reduces the risk of overfitting to a single negative control and provides a more stringent assessment of cell-type specificity. As above, higher values indicate stronger specificity.

\item \textbf{Motif enrichment score}.
We assess motif enrichment by scanning designed sequences for known transcription factor binding sites (TFBSs) relevant to hepatocyte biology and computing the proportion of motif occurrences across the generated sequence set. The dataset of TFBSs is JASPAR18 \citep{sandelin2004jaspar}. This metric provides an explicit, interpretable link between sequence design and known regulatory mechanisms, complementing purely predictive MPRA-based scores. By grounding evaluation in curated TF motifs, motif enrichment serves as a biological plausibility check, ensuring that high-scoring sequences are consistent with established transcriptional programs rather than exploiting model artifacts. The metric ranges from $0$ to $\#\text{sequences} \times \#\text{motifs}$, with higher values indicating stronger enrichment.

\item \textbf{Diversity score}.
We compute diversity as the average pairwise Hamming distance across all generated sequences. This metric evaluates whether a method produces a diverse set of solutions rather than collapsing to a small number of high-scoring templates. Diversity is particularly important for regulatory sequence design, as multiple distinct sequence architectures can realize similar functional outputs in vivo. By explicitly measuring sequence-level variation, this metric discourages mode collapse and complements functional scores that alone could be optimized by near-duplicate sequences. The score ranges from $0$ to $\infty$, with higher values indicating greater diversity.

\item \textbf{Stability score (GC content).}
We quantify sequence stability using the proportion of G/C nucleotides across the generated sequences. GC content is a well-established proxy for DNA thermodynamic stability due to increased hydrogen bonding and stacking interactions, and it has been widely used in prior sequence design work as a simple, interpretable constraint. While not a direct measure of enhancer activity, this metric helps ensure that generated sequences remain within a biologically reasonable compositional regime and avoids extreme or degenerate nucleotide distributions. The score ranges from $0$ to $1$, with higher values indicating higher GC content and increased stability.

\end{itemize}

Collectively, these metrics provide a balanced evaluation of functional specificity (Specificity1/2), mechanistic plausibility (motif enrichment), and generative quality (diversity and stability), reflecting both experimental and biological considerations in enhancer design.

\textbf{Baselines}. We benchmark against both state-of-the-art domain-specific methods and general-purpose AI agents. The baselines include regLM \citep{lal2024designing}, TextGrad \citep{yuksekgonul2025optimizing}, and Biomni \citep{huang2025biomni}. Details of each baseline are described below:

\begin{itemize}
\item \textbf{regLM} is a fine-tuned genomic language model (base model: HyenaDNA \citep{nguyen2023hyenadna}). We finetune this model with same datasets used to train the predictor for MPRA prediction. We then generate 5,000 candidate HepG2-specific enhancers using regLM and randomly select 20 sequences for evaluation.
\item \textbf{TextGrad} is an LLM-based optimization framework. Initialized with the same objective functions, TextGrad is used to design 20 HepG2-specific enhancer sequences.
\item \textbf{Biomni} is an autonomous AI scientist designed for general biomedical tasks. Given the same objective specifications, Biomni is used to generate 20 HepG2-specific enhancer sequences.
\end{itemize}

\textbf{Hyperparameters}. Functional DNA sequence generation follows a general outer-loop framework coupled with a domain-specific optimizer, in which large language models act as optimizers to iteratively mutate sequences and improve the associated objective scores. We initialize the process with 5,000 random DNA sequences and use a batch size of 20. For agents from different modes, the optimization is run for up to three iterations, with early stopping governed by a selection agent; all experiments support this early-stopping mechanism. The initial objectives are defined by predicted MPRA expression levels from a trained predictor. When reporting both optimization scores and held-out evaluation metrics, we apply min–max normalization to place all scores on a comparable and interpretable scale. Each baseline method also has three replicates under different random seeds.

\textbf{Workflows}. Our three different levels (co-pilot, semi-pilot, and autopilot) follow the default setting, discussed in \Cref{sec:auto_level}. For all experiments, we have three replicates.

\subsubsection{High-level Goal}
\label{sup:highlevel dna}
Here we use HepG2 as an example: Generate a set of cell-type-specific enhancers for the HepG2 cell line, each with a length of 200 base pairs.

\subsubsection{Context Information}
\label{sup:context dna}
Here we use HepG2 as an example: For this task, the enhancers should be specific to the HepG2 cell line, meaning they should drive high expression in HepG2 cells while minimizing expression in other cell lines (e.g., K562 and SKNSH). The sequences should also be diverse to cover a broad range of potential enhancer activities. You can consider including objectives related to known enhancer motifs and stability of DNA sequences. The optimizer will automatically enforce the length constraint, so do not propose any objectives related to enhancer length.

\subsection{Additional Experimental Results}

\subsubsection{Experiments for all cell lines}

\textbf{Examination of cell-type specificity in MPRA expression.}
According to \Cref{supfig:dna_merged_abla} (a), the designed HepG2-specific enhancers from \name all have obvious specificity, which is equivalent to the cell-type-specific expressions in predicted MPRA score. 

\textbf{Examination of evaluation metrics in enhancer design for different cell types.}
We further evaluate the performance of \name in designing K562-specific and SKNSH-specific enhancers, with results summarized in \Cref{supfig:enhancer_resttwo}(a) and (b). These results demonstrate that our system generalizes effectively across cell lines, consistently preserving both MPRA-based specificity and biology-driven metrics. In terms of average performance, \name outperforms all baselines by at least 16.3\% in the K562 cell line and by at least 1.5\% in the SKNSH cell line.

\textbf{Examination of evaluation metrics in promoter design for different cell types.}
We also examine the performances of \name in designing cell-type-specific promoters across five different cell lines, including K562, HepG2, GM12878, SKNSH, and A549. The results (Co-pilot, Semi-pilot, and Autopliot versus other baselines) are summarized in \Cref{supfig:promoter design}. \name can also generalize promoters with good quality and specificity, which further supports the capacity of our method in handling tasks from different modalities. 

\subsubsection{Ablation Studies}

\textbf{Evaluation of the Implementer}. We test the ability of the Implementer by validating whether it can propose similar objectives that have human implementation. Here, we focus on one example of predicting the difference between HepG2's MPRA and other cell lines' MPRA. \Cref{supfig:dna_merged_abla} (b) also shows high and significant correlations, confirming that the Implementer can internalize the structure of the design landscape and construct objectives that reflect biological ground truth.

\textbf{Evaluation of optimizer}. To validate whether our optimizer can work as we expect, we include an ablation study to check the performance of this agent in improving the initial objectives versus the initial populations, which are random DNA sequences. As shown in \Cref{supfig:dna_merged_abla} (c), the optimizer can already identify HepG2-specific enhancer patterns when considering MPRA differences alone. However, because the optimizer cannot balance competing biological constraints, we need to integrate different modules, including Planner, Analyzer, and Implementer to propose more complicated objectives and produce final candidates. This motivates the development of more advanced agents.

\section{Inorganic Materials Design}

\subsection{Supplementary Figures}

\begin{figure}[H]
    \centering
    \includegraphics[trim = 0 0 0 0, clip, width=1.0\linewidth]{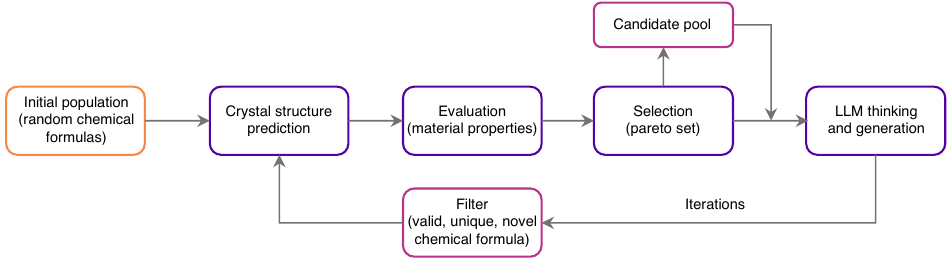}
    \caption{Schematic diagram of the optimizer using LLM-based evolutionary algorithm.}
    \label{supfig:crystal_opt}
\end{figure}

\begin{figure}[H]
    \centering
    \includegraphics[trim = 0 0 0 0, clip, width=0.4\linewidth]{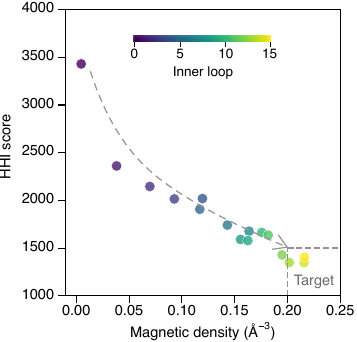}
    \caption{In Co-pilot mode, the LLM-based optimizer simultaneously optimizes two material properties, targeting magnetic density higher than 0.2 $\text{\AA}^{-3}$ and HHI score less than 1500.}
    \label{supfig:crystal_opt_loop}
\end{figure}

\begin{figure}[H]
    \centering
    \includegraphics[trim = 0 0 0 0, clip, width=0.5\linewidth]{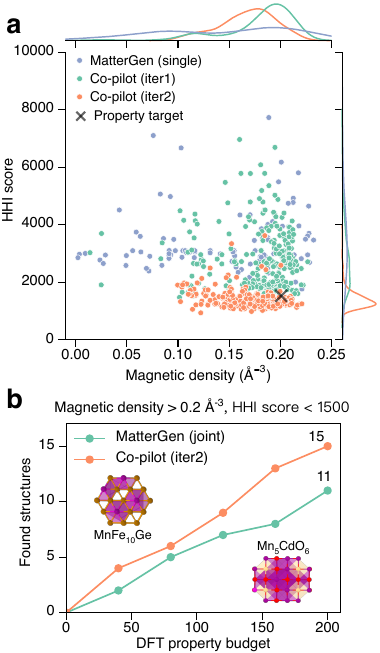}
    \caption{(a) Property distributions of generated structures from co-pilot across different iterations and from MatterGen (single) targeting only high magnetic density.
    (b) Number of stable and novel structures satisfying property requirements found by co-pilot and MatterGen (joint) within 200 DFT property calculations, for targets with magnetic density above 0.2 $\text{\AA}^{-3}$ and HHI score below 1500. It also displays 3D visualizations of two crystal structures proposed by the co-pilot mode that satisfy the design goal.}
    \label{supfig:mattergen}
\end{figure}

\begin{figure}[H]
    \centering
    \includegraphics[trim = 0 0 0 0, clip, width=1.0\linewidth]{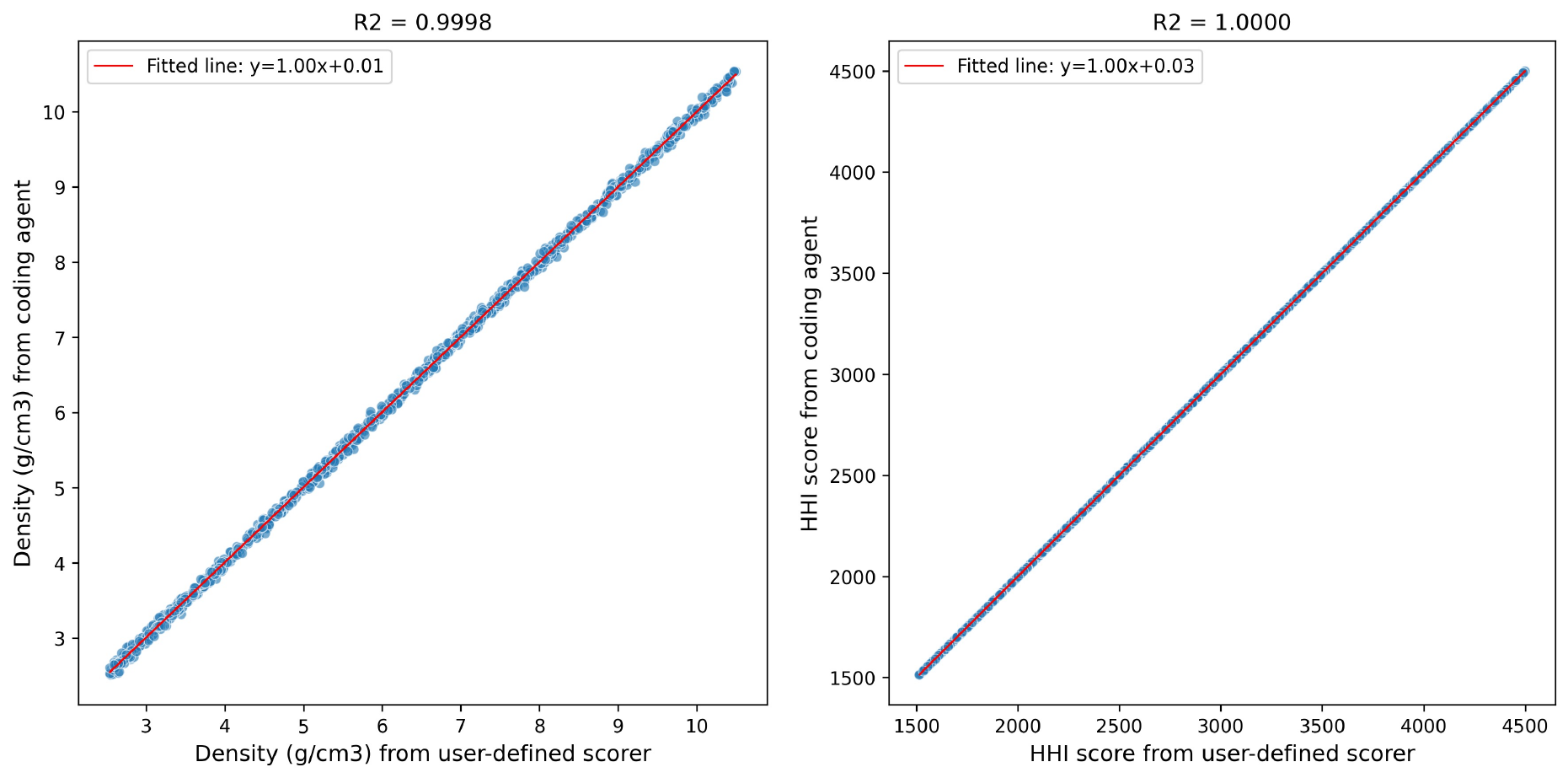}
    \caption{Correlation between human-proposed scoring function and Implementer-proposed scoring function.}
    \label{supfig:crystal_code_agent}
\end{figure}

\subsection{Experimental Setups}
\subsubsection{LLM-based evolutionary algorithm for materials design}
As shown in \Cref{supfig:crystal_opt}, the material property optimization loop employed a LLM-based evolutionary algorithm. Initial populations (chemical formulas of crystals) were randomly sampled from the Materials Project database \citep{mp}, which also serves as the first group of crystals in the parent node. The LLMs are used to generate chemical formulas, and then DiffCSP diffusion model \citep{diffcsp} is used to generate 3D crystal structures, which was pretrained on the MP-20 dataset \citep{mp}. Geometric optimization are performed for the generated structures using universal ML force fields (MatterSim \citep{mattersim}). Evaluators assigned objective scores based on the 3D structure of each crystal. Chemical formulas from the parent node and individual score of each objective were provided to the LLM, which generated new formulas through crossover operations. Optimal structures are then selected via Pareto front analysis from a combined pool of generated and parent crystals.

\subsubsection{Objectives, metrics, and baselines}
In the task of designing permanent magnets with low supply chain risk, two objectives were specified: magnetic density higher than 0.2 $\text{\AA}^{-3}$ and HHI score less than 1500. The \name Co-pilot mode was deployed with iteratively refined objectives: maximizing magnetic density in the first iteration, followed by the addition of HHI score minimization in the second. During optimization, the ALIGNN model \citep{alignn} pre-trained on DFT data from the Materials Project database \citep{mp} is used as the scorer for magnetic density prediction. The HHI scores are calculated using the pymatgen package \citep{pymatgen}. The thermodynamically Stability ($E_{hull} <$ 0.1 eV/atom), Uniqueness, and Novelty (SUN) \cite{mattergen} of each generated structure was evaluated by MatterGen's method using their Alex-MP reference dataset and code \cite{mattergen}, and only SUN structures were retained. During optimization, the generated structures were optimized using ML force field (MatterSim \cite{mattersim}) due to the high computational cost, and $E_{hull}$ was obtained by MLIP energy. Finally, 200 crystal structures generated from each iteration were randomly selected and DFT verified.

MatterGen models \citep{mattergen} that target only high magnetic density (single) or both properties (joint) were used to generate 4000 structures by their conditional generation method. For MatterGen (single), the conditional generation's target is magnetic density of 0.2 $\text{\AA}^{-3}$. And the conditional generation's target of MatterGen (joint) is magnetic density of 0.2 $\text{\AA}^{-3}$ with HHI score of 1500. Only SUN structures were retained, and 200 crystal structures were randomly selected for DFT evaluation.

In the task of designing superhard materials for precision cutting, the evaluation metrics include Vickers hardness, bulk modulus, shear modulus, Pugh ratio, and energy above hull. The initial objectives of \name experiments in different modes are to maximize the bulk modulus and the shear modulus. During optimization, the ALIGNN models \citep{alignn} pre-trained on DFT data from the Materials Project database \citep{mp} are used as the scorers for prediction of bulk and shear modulus. For performance evaluation, the top 100 crystal structures from the final LLM-ranked candidates after convergence were selected for DFT calculations to obtain scores for each metric.

\textbf{Baselines}. We benchmark against both state-of-the-art domain-specific methods and general-purpose AI agents. The baselines include MatterGen \citep{mattergen} and TextGrad \citep{yuksekgonul2025optimizing}. Details of each baseline are described below:

\begin{itemize}
\item \textbf{MatterGen} is a diffusion model for inorganic material generation. The unconditional MatterGen model was pretrained on the MP-Alex-20 dataset \citep{mattergen}, which contains unlabeled crystal structures, enabling the generation of stable and novel structures. Furthermore, the adapter-equipped MatterGen model was fine-tuned on crystal structures with DFT-derived labels, thereby enabling controllable generation of crystals with desired properties.
\item \textbf{TextGrad} is an LLM-based optimization framework. Initialized with the same objective functions, TextGrad is used to design chemical formula of inorganic materials. The 3D crystal structures corresponding to chemical formulas were generated using the DiffCSP diffusion model \citep{diffcsp}, which was pretrained on the MP-20 dataset \citep{mp}.
\end{itemize}

\textbf{Hyperparameters}. Inorganic materials design follows a general outer-loop framework coupled with a domain-specific optimizer, in which LLMs act as optimizers to iteratively mutate sequences and improve the associated objective scores. We initialize the process with 1,000 random chemical formulas of crystals from Materials Project database \citep{mp} and use a batch size of 20. For agents from different modes, the optimization is run for up to ten iterations, with early stopping governed by a selection agent; all experiments support this early-stopping mechanism. When reporting both optimization scores and held-out evaluation metrics, we apply min–max normalization to place all scores on a comparable and interpretable scale. Each baseline method also has three replicates under different random seeds.

\textbf{Workflows}. Our three different levels (co-pilot, semi-pilot, and autopilot) follow the default setting, discussed in \Cref{sec:auto_level}. For all experiments, we have three replicates.

\subsubsection{High-level Goal}
\label{sup:mat_goal}
In the task of designing permanent magnets with low supply chain risk, high-level goal for \name agent is "Generate a set of chemical formulas of crystals for permanent magnet materials with high magnetic density and low supply chain risk."

In the task of designing superhard materials for precision cutting, high-level goal for \name agent is "Generate a set of chemical formulas of superhard materials for Ultra-Precision Cutting Tools."

\subsubsection{Context Information}
\label{sup:mat_context}
In two tasks, the context information for \name agent is "Please ensure that the proposed objectives are common and well-defined material properties in materials science. Please try to propose material properties that have not been considered in previous iterations but are relevant to the design goal."

\subsection{Material Property Evaluation}
\label{sup:mat_eval}
We performed density functional theory (DFT) computations employing the Vienna Ab initio Simulation Package (VASP) \citep{vasp1, vasp2} in conjunction with the projector augmented wave (PAW) approach. The calculations were implemented through atomate2 \citep{atomate2} and pymatgen \citep{pymatgen} software packages. The computational setup adhered to Materials Project \citep{mp} standards, incorporating the Perdew–Burke–Ernzerhof (PBE) functional under the generalized gradient approximation (GGA) framework \citep{pbe_gga1, pbe_gga2}. The computational procedures for various properties are described below:

(1) The total energy and energy above hull were computed through the \texttt{DoubleRelaxMaker} and \texttt{StaticMaker} modules in atomate2 \citep{atomate2} using default configurations. This protocol comprises two consecutive structural relaxations followed by a static energy calculation.

(2) Magnetic densities of the generated structures were evaluated using the \texttt{DoubleRelaxMaker} and \texttt{StaticMaker} modules in atomate2 \citep{atomate2} with standard configurations. This procedure consists of two sequential relaxations and a static calculation. We define magnetic density as the ratio of total magnetization (magnetic moment) of the unit cell to its volume.

(3) Elastic modulus were determined via the \texttt{ElasticMaker} module in atomate2 \citep{atomate2} with standard configurations. First, the structure undergoes thorough structural optimization to achieve a nearly stress-free equilibrium configuration. Next, systematic deformations are introduced to the lattice parameters, and the corresponding stress tensors are computed using DFT calculations, with simultaneous optimization of atomic positions. The resulting stress-strain relationships are then fitted using linear elastic theory to determine the complete 6$\times$6 elastic tensor. This tensor enables the calculation of averaged mechanical properties, including the Voigt and Reuss estimates for bulk and shear moduli. Vickers hardness was calculated using Tian's empirical equation \citep{tian_hardness} based on DFT-computed bulk and shear moduli.

The Herfindahl-Hirschman index (HHI) scores based on geological reserves for crystals were calculated using the \texttt{HHIModel} class from the pymatgen package \citep{pymatgen}. This compositionally-based metric, derived from geological and geopolitical data, quantifies resource-related economic factors and assesses the supply-demand risk of materials. Additionally, it measures the degree to which the constituent elements of a compound are geographically concentrated or dispersed. The HHI parameter is computed as the sum of squared market fractions ($\chi_i$) for each country, based on either production ($\mathrm{HHI}_{\mathrm{P}}$) or geological reserves ($\mathrm{HHI}_{\mathrm{R}}$) of individual elements, using United States Geological Survey (USGS) commodity statistics \citep{hhi}. For each composition, the weighted average $\mathrm{HHI}_{\mathrm{R}}$ value was calculated using the weight fraction of each element in the chemical formula. According to the U.S. Department of Justice and Federal Trade Commission, markets are classified as unconcentrated (HHI $<$ 1500), moderately concentrated (1500 $\leq$ HHI $\leq$ 2500), or highly concentrated (HHI $>$ 2500) for a given commodity. Lower HHI values are preferable, with materials having HHI scores below 1500 considered to exhibit low supply chain risk \citep{hhi}.

\section{Chemical Process Design}
\subsection{Supplementary Figures}\label{sup:SI_process_figures}

\begin{figure}[hptb]
    \centering
    \includegraphics[trim = 0 7cm 6cm 1cm, clip, width=1\linewidth]{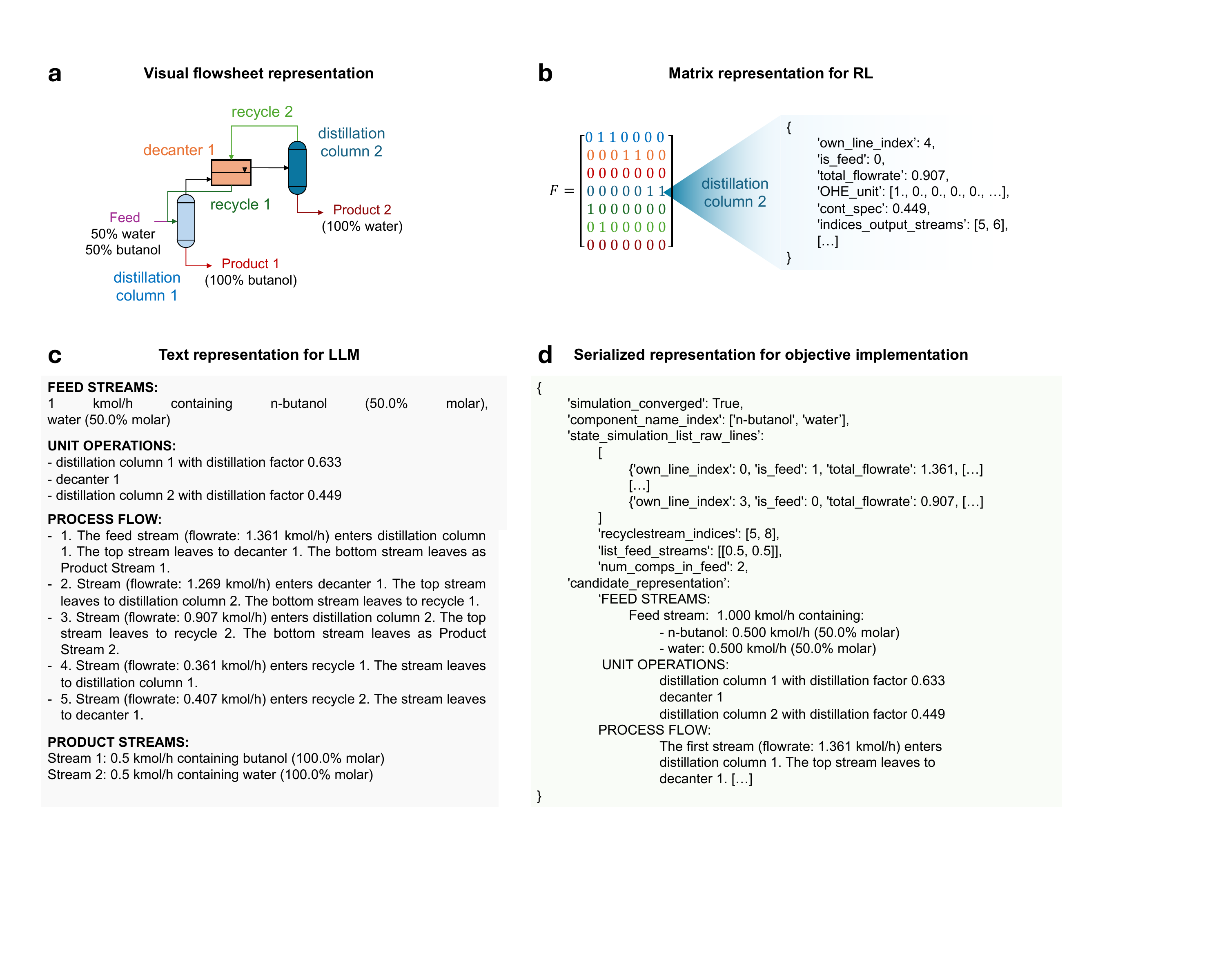}
    \caption{Different representations of an exemplary chemical process: (a) Visual representation that is created manually for illustration purposes; (b) matrix representation used within the RL optimizer~\citep{gottl2025deep}, whereas the entries correspond to connection between the unit operations and product streams (colors correspond to (a); (c) text representation that is automatically generated based on (b) for use within \name; (d) serialized process representation that is automatically generated from (b) and (c), and used to implement objectives by \name.}
    \label{supfig:SI_process_representation}
\end{figure}

\begin{figure}[hptb]
    \centering
    \includegraphics[trim = 0 4.5cm 0cm 4cm, clip, width=0.9\linewidth]{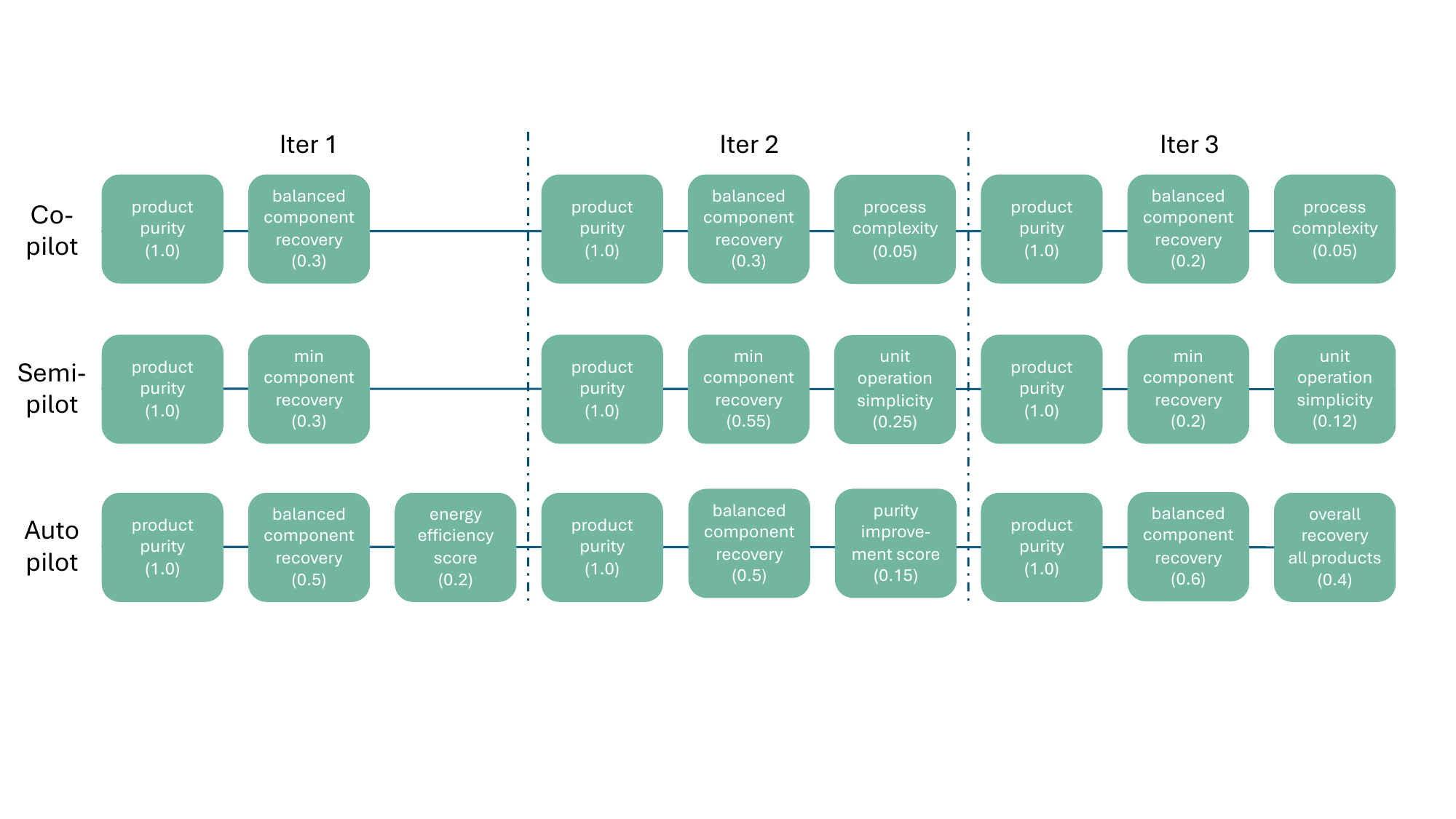}
    \caption{Objectives proposed in exemplary runs using different levels of SAGA. \name.}
    \label{supfig:SI_process_objectives_overview}
\end{figure}

\begin{figure}[hptb]
    \centering
    \includegraphics[trim = 0 0 0cm 0, clip, width=0.9\linewidth]{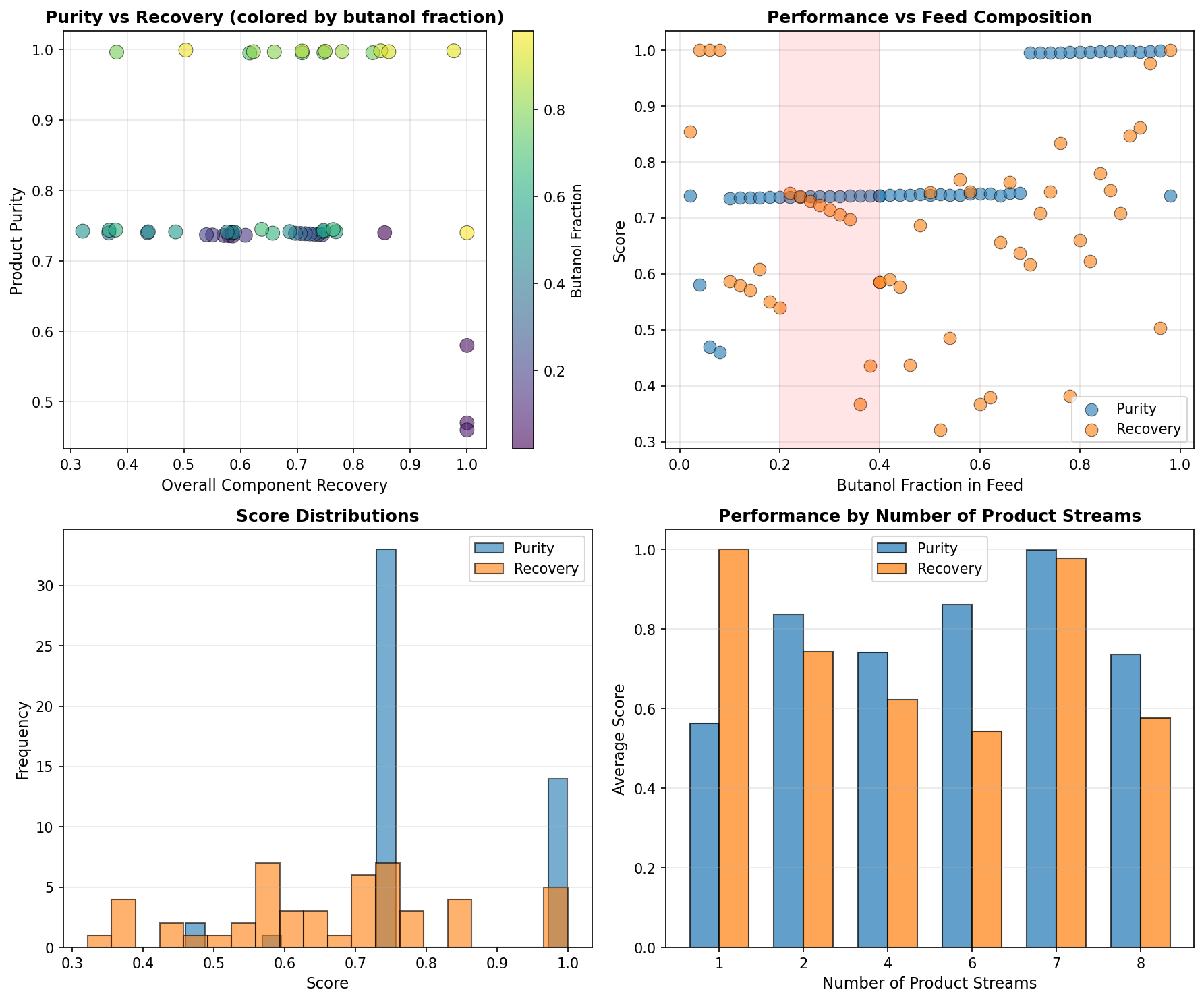}
    \caption{Exemplary plot for process candidate analysis automatically generated by \name.}
    \label{supfig:SI_process_candidate_analysis}
\end{figure}

\clearpage

\subsection{Experimental Setups}\label{sup:SI_process_exp}

\subsubsection{Objectives, metrics, and baselines}

Here we explain the experimental details of chemical process design.

\textbf{Initial objectives}. Our initial objective is maximize the product purity.

\textbf{Evaluation metrics}. We use three objectives as our evaluation metrics, which we implement based on the short-cut process simulation models in~\cite{gottl2025deep}:
\begin{itemize}
\item Product purity: average purity of the output/product streams, whereas streams are rewarded if they have a molar composition $x > 0.9$, with extra rewards for $x > \{0.95, 0.97, 0.99\}$. We choose this metric because achieving pure product streams is the main goal in designing separation processes. This metric is naturally normalized to the range from 0 to 1, and higher is better.
\item Capital costs: negative sum of costs of process unit operations. Notably, the costs for the individual unit operations are based on simple heuristics, similar to the work~\citep{gottl2025deep}. We choose this metric because low capital costs are a key goal in chemical process design. This metric is normalized to the range from 0 to 1, and higher is better (as we negate the sum).
\item Material flow intensity: negative penalty for process streams smaller than 1\% of the feed stream and for (excessive) recycle streams greater than the feed stream. We choose this metric because excessive recycle ratios and very small streams lead to practical issues in process operation and equipment design. This metric is normalized to the range from 0 to 1, and higher is better (as we negate the penalty).
\end{itemize}

\textbf{Evaluation set}. The RL agent is evaluated on a set of processes covering the discretized composition range $x\in [0,1]$ with a step size of $\Delta x = 0.02$. 
Thus, each evaluation covers a wide range of butanol/water mixtures, testing whether the RL agent can design chemical process flowsheets for varying feed compositions.

\textbf{RL optimizer}. For the optimization in the inner loop of \name, we use an RL agent based on the framework by \citet{gottl2025deep}. 
This RL agent operates on matrix representations of process flowsheets, whereas the rows and columns represent unit operations, recycles, and product streams; the entries in the matrix correspond to connections, i.e., material flows.
The flowsheet design is then formulated as a sequential planning problem by selecting unit operations and corresponding material flows.
Specifically, the design space comprises four unit operations, each with its own specifications: a distillation column, a decanter, a mixer, and a splitter. 
The design also determines the flow structure of the feed stream to these unit operations and classifies intermediate streams as inputs to additional unit operations or final output streams.
Each action is determined in a hierarchical manner including both discrete variables, e.g., selecting a distillation column as unit operation, and continuous variables, e.g., specifying the ratio of input to distillate flow rate within the distillation column.
Material flows can also include recycles, which affect pervious process states, and thus require RL agents to plan ahead~\citet{gottl2025deep}.
The reward for the agent is based on evaluating the designed processes with short-cut process models and associated objectives.

\textbf{Baselines}. We consider the RL agent~\citep{gottl2025deep} trained only on product purity as baseline, as the purity is the main objective and typically further objectives would be added manually in an iterative fashion by human experts.
We run the baseline three times under different random seeds.
Furthermore, we considered adding TextGrad as a baseline.
However, since the flowsheet design also requires determining continuous unit operations parameters, such as distillation factors and recycle ratios, to which the process (simulation) can be quite sensitive, we did not further consider purely LLM-guided process design. 
Notably, this topic is still highly underexplored due to the inherent complexity of process representations.

\textbf{Hyperparameters}. For the RL agent, we use the code base and hyperparameters from the original work in \citep{gottl2025deep}.
However, we only consider butanol/water mixtures without the possibility to add solvents.  
Since this decreases the problem complexity compared to training an RL agent to design flowsheets for different mixtures using solvents, and in order to save computational resources, we reduce the number of training batches from 10,000 to 1,000.
For the outer loop, we use the settings for the analyzer, planner, and implementer as described in the main text, and run it for three iterations.
We repeat each experiment three times under different random seeds.

\textbf{Workflows}. Our three different levels (co-pilot, semi-pilot, and autopilot) follow the default setting, discussed in \Cref{sec:auto_level}. For all experiments, we have three replicates.

\subsubsection{High-level goal}
We provide the following goal as prompt to SAGA: ``Design chemical process flowsheets for the separation of binary aezeotropic mixtures, i.e., separating a feed stream with two components into high purity streams.''

\subsubsection{Context}
In addition, we provide the following context within the prompt: ``For this task, the chemical processes will be generated by a reinforcement learning agent that will be trained to optimize your provided objective scores. Please note that the training starts from scratch in each iteration (each time new objectives are provided), so the agent has to learn the process design from scratch in each iteration and should account for the main objective, i.e., the initial objective product purity should be kept as the focus. The agent is then tested to design separation processes for a set of different feed compositions, which form the population. Note that the designed processes are evaluated with short-cut models that only converge if a physically consistent process was designed, i.e., you do not need to consider convergence issues and physical consistency, e.g., mass balances, as part of your analysis or the objectives. The designed process flowsheets should fulfill early-stage process design goals relevant for practical application and further refined process design. The central goal for this is to get very high purity streams. Other relevant goals for efficient chemical processes and their implementation in practice should also be considered. For this, you can consider including objectives related to known process design goals and existing separation processes. Note that it is important to analyze all processes in the population as they have different feed compositions. Please also note that training the reinforcement learning agent is computationally expensive and not always stable, e.g., sensitive to the selected objective weights or the combination of certain objectives. This also includes that adding too many new objectives for a new iteration might lead to unstable training, so consider using less than the maximum number of objectives, especially in early iterations.''

\subsection{Challenges in chemical process design}\label{sup:SI_process_challenges}
Using \name for chemical process design comes with several challenges including: 
\begin{enumerate}
    \item Text-based representation of chemical processes are complex, as they contain topological process information but also detailed specifications of process units, cf. \Cref{supfig:SI_process_representation} and~\cite{gottl2025deep, gao2024deep}, making analysis and implementing design objectives highly difficult. 
    Notably, Simplified Flowsheet Input-Line Entry-System (SFILES) -- inspired by SMILES for molecules -- were developed for representing chemical processes as text in a semantic, standardized way~\cite{d2006process}. SFILES have also recently been adapted to work with transformer models~\cite{vogel2023sfiles, vogel2023learning, mann2024esfiles}. Yet, they are rarely explored and usable for LLMs so far, which is why we use a simple, more intuitive text representation.
    \item Process evaluations are based on simulation models that can be computationally expensive and sensitive to (continuous) process parameters~\cite{gao2024deep}.
    \item Process design objectives can be conflicting, e.g., high purity versus capital costs.
    \item The design relies on an RL agent which is sensitive to the objectives and their weights, thus can converge to non-practical process designs.
\end{enumerate}
Therefore, we consider a simple separation process design case study, whereas our evaluation focuses on whether the \name framework is able to guide chemical process design by identifying process issues and suitable objectives.
We stress that we expect variations with respect to the objectives on the designed processes along the iterations, due to the sensitivity and partly instability of the RL agent used for optimization.

\section{Agent Processing History}

In this section, we provide concrete examples for \name procedures across different tasks.

\subsection{Antibiotic Design}
\label{subsec:antibiotic history}

Below is an abbreviated chat history from the Co-pilot mode. After finishing the first round of optimization, a human scientist inspects the proposed candidates, identifies the low synthesizability issue in the population, and suggests that the implementer should implement another objective via RDkit's SAScore.

\begin{tcolorbox}[colback={rgb,255:red,217;green,247;blue,243}, colframe={rgb,255:red,114;green,182;blue,161}, breakable, title=\text{Analyzer's, Planner's, and Implementer's output in the Co-pilot workflow of \name}]

\textbf{Inputs: Received initial population with 120 candidates. The goal is to design novel small molecules effective against  with high safety and drug-like properties.}\\

\textbf{\textcolor{myanalyzer}{Analyzer}}
\paragraph{Overview}
\begin{itemize}
    \item Iteration 1 produced 120 optimized candidates with 100\% validity.
    \item \textbf{Key outcome:} A dramatic 186$\times$ improvement in mean antibacterial activity ($0.0018 \to 0.3356$) was achieved. However, only 6.7\% (8 candidates) simultaneously satisfy all quality criteria. [\textcolor{myanalyzer}{analyzer came up with this criteria itself}]
\end{itemize}

\paragraph{Performance Analysis}
\begin{itemize}
    \item \textbf{Antibacterial Activity:} Mean $0.3356 \pm 0.0704$, up by $+0.3338$ from the initial pool. The best candidate achieves $0.5467$. 30\% of the population reaches the 70\% precision threshold ($\geq 0.37$).
    \item \textbf{Antibiotics Novelty:} Mean $0.7978 \pm 0.0231$. The optimizer maintained high novelty, with 56.7\% of candidates scoring $\geq 0.8$.
    \item \textbf{Toxicity Safety:} Mean $0.8352 \pm 0.0454$. Safety remains robust, with 84.2\% of candidates achieving high safety scores.
    \item \textbf{Drug-likeness (QED $\times$ MW):} Mean $0.6377 \pm 0.0692$, a moderate decrease of $-10.4\%$.
\end{itemize}

\paragraph{Issues and Concerns}
\begin{itemize}
    \item \textbf{Drug-likeness Erosion:} The structural exploration for activity is moving into slightly less drug-like chemical space, though 100\% of candidates still comply with Lipinski's Rule of 5.
    \item \textbf{Synthesis Concerns:} Despite purchasability targets, 100\% of candidates have 2--4 chiral centers, which may complicate synthesis for the 59.2\% of molecules that are not directly purchasable.
\end{itemize}

\noindent
\textbf{Human [after inspecting the top 10 molecules]: Some of the molecular candidates are not very synthesizable. Let's regulate that by adding a synthesizability score using RDKit SAScore.}

\paragraph{\textcolor{myplanner}{Planner}}
\begin{itemize}
    \item Maximize \texttt{antibacterial\_activity\_minimol} to ensure efficacy against the target pathogen (\textbf{Human proposed}).
    \item Maximize \texttt{antibiotics\_novelty} using Morgan fingerprints to circumvent existing resistance mechanisms (\textbf{Human proposed}).
    \item Apply a composite Drug-likeness score to ensure ADME properties, defined as:
    \[
        \text{Score} = \text{QED} \times \text{MW\_penalty}
    \]
    where the molecular weight penalty is applied for $MW > 500$:
    \[
        \text{MW\_penalty} = \frac{1}{1 + \exp((MW - 500) / 50)}
    \]
    (\textbf{Agent proposed}).
    \item Filter for known antibiotic motifs and PAINS alerts to ensure structural novelty (\textbf{Human proposed}).
    \item Maximize \texttt{local\_similarity} to Enamine REAL space to prioritize purchasable compounds (\textbf{Human proposed}).
    \item Maximize \texttt{synthesizability} score to ensure molecules' synthesizability (\textbf{Human proposed}).
\end{itemize}

\noindent
\textbf{Human: Looks good to me. Let's add the SAScore and maximize it.} \\

\textbf{\textcolor{myimplementer}{Implementer}: implementing SAScore ...}

\end{tcolorbox}

Below is another abbreviated chat history from the Semi-pilot mode. After finishing the first round of optimization,the analyzer inspects the proposed candidates and points out the over-prevalence of metabolically unstable groups and a negative correlation between activity and drug likeness. The planner receives this information and decides to implement new scorers and adjust objective weights for the next round of optimization.

\begin{tcolorbox}[colback={rgb,255:red,250;green,230;blue,222}, colframe={rgb,255:red,233;green,150;blue,117}, breakable, title=\text{Analyzer's, Planner's and Implementer's output in the Semi-pilot workflow of \name}]

\textbf{Inputs: Received initial population with 120 candidates. The goal is to design novel small molecules effective against the bacteria with high safety and drug-like properties.}\\

\textbf{\textcolor{myanalyzer}{Analyzer}}

\paragraph{Overview}
\begin{itemize}
    \item Iteration 1 demonstrates \textbf{strong progress} in candidate optimization.
    \item \textbf{Key outcome:} The primary objective (antibacterial activity) improved by $122\times$ (mean $0.0984$). 100\% of candidates passed both the motif and molecular weight filters.
    \item \textbf{Trade-offs:} While activity and novelty increased, moderate declines were observed in safety ($-4.6\%$)
\end{itemize}

\paragraph{Performance Analysis}
\begin{itemize}
    \item \textbf{Antibacterial Activity:} Mean $0.0984$. The top performing molecule achieves $0.3410$, approaching the 70\% precision threshold ($0.37$).
    \item \textbf{Antibiotics Novelty:} Mean $0.7556$ ($+0.0384$). High structural novelty is being maintained effectively.
    \item \textbf{Filters:} 100\% pass rate for both \texttt{antibiotics\_motifs\_filter} and \texttt{molecular\_weight\_filter} [\textcolor{myanalyzer}{implemented before iter1 optimization}].
    \item \textbf{Safety \& Drug-likeness:} Toxicity safety score averaged $0.739$ (moderate decline). QED drug-likeness saw a slight decline of $-0.0428$.
\end{itemize}

\paragraph{Top Candidate Profile}
\begin{itemize}
    \item \textbf{SMILES:} \texttt{COc1c(F)c(C(=O)N)c2ncsc2c1C(CO)N(CCCN)CC}
    \item \textbf{Metrics:} Activity: $0.3410$ | Novelty: $0.8108$ | Safety: $0.7394$ | QED: $0.6119$.
    \item This candidate demonstrates that balanced, high-quality molecules are achievable despite population-level trade-offs.
\end{itemize}

\noindent
\textbf{Human: Looks good, but please keep improving the antibiotic activity score, since it is still pretty low.}

\paragraph{\textcolor{myplanner}{Planner}}
\begin{itemize}
    \item \textbf{Refined Drug-likeness:} Introduced \texttt{qed\_druglikeness} to explicitly maximize the Quantitative Estimate of Drug-likeness score (Target $>0.67$).
    \item \textbf{Size Constraints:} Added \texttt{molecular\_weight\_filter} to strictly enforce the small molecule limit (150--500 Da).
    \item \textbf{Maintained Objectives:} Continued maximizing \texttt{antibacterial\_activity\_minimol} and \texttt{antibiotics\_novelty} with higher weight while monitoring safety via \texttt{toxicity\_safety\_chemprop}.
\end{itemize}

\textbf{\textcolor{myimplementer}{Implementer}: implementing the size constraint filter ...}

\end{tcolorbox}

Below is the final abbreviated chat history from the Autopilot mode. After finishing the first round of optimization, a human scientist inspects the proposed candidates along with the analysis report and points out that the overall antibiotic activity score is still quite low. The planner receives this information and decides to continue optimizing antibiotic activity score by reweighting the objective weights. It also suggests a size constraint filter to the implementer for the next round of optimization. 

\begin{tcolorbox}[colback={rgb,255:red,255;green,245;blue,200}, colframe={rgb,255:red,220;green,180;blue,60}, breakable, title=\text{Analyzer's and Planner's output in the Autopilot workflow of \name}]

\textbf{Inputs: Received initial population with 120 candidates. The goal is to design novel small molecules effective against the bacteria with high safety and drug-like properties.}\\

\textbf{\textcolor{myanalyzer}{Analyzer}}

\paragraph{Overview}
\begin{itemize}
    \item Iteration 1 achieved \textbf{dramatic progress}, with a $100\times$ improvement in mean antibacterial activity ($0.0008 \to 0.1051$) compared to iteration 0.
    \item \textbf{Structural Success:} 100\% of candidates passed both the motifs filter and the molecular weight filter.
    \item \textbf{Key Insight:} The optimizer identified specific activity-driving features: nitrile groups (present in 85\% of candidates), pyrimidine cores ($3.5\times$ enrichment), and fluorine substitution. [\textcolor{myanalyzer}{Analysis aligning with scientists' intuition on chemical scaffolds.}]
\end{itemize}

\paragraph{Performance Analysis}
\begin{itemize}
    \item \textbf{Antibacterial Activity:} Mean $0.1051$. The best candidate reached $0.271$. 23 candidates (19.2\%) achieved high precision ($\geq 0.16$), indicating the optimizer has successfully located active chemical space.
    \item \textbf{Antibiotics Novelty:} Mean $0.762$ ($+0.045$). 20 candidates scored $\geq 0.8$, ensuring exploration of distinct chemical space.
    \item \textbf{Safety \& Drug-likeness:} Both metrics saw regressions. Toxicity safety dropped to $0.751$ ($-0.071$), and QED drug-likeness dropped to $0.725$ ($-0.054$).
\end{itemize}

\paragraph{Top Candidate Profile}
\begin{itemize}
    \item \textbf{SMILES:} \texttt{ClCc1ncc(C\#N)nc1N(CCO)CCc2cc(Cl)c(C\#N)c(F)c2CN}
    \item \textbf{Metrics:} Activity: $0.2649$ | Novelty: $0.7727$ | Safety: $0.7466$ | QED: $0.6246$.
    \item This candidate exemplifies the "sweet spot": high activity with acceptable safety, despite the population-level trade-offs.
\end{itemize}

\paragraph{Issues and Concerns}
\begin{itemize}
    \item \textbf{Critical Trade-off:} There is a strong negative correlation between Activity and QED ($r=-0.370$) and a moderate negative correlation between Activity and Safety ($r=-0.281$).
    \item \textbf{Structural Conflict:} High-activity molecules in this iteration tend to be larger (Mean MW $398.6$ Da) and more complex, which improves efficacy but systematically degrades pharmaceutical properties.
    \item \textbf{Metabolic stability:} 80\% of high-activity molecules contain primary amines susceptible to rapid metabolism; morpholine rings (40\% prevalence) are also metabolically labile.
\end{itemize}

\paragraph{\textcolor{myplanner}{Planner}}
\begin{itemize}
    \item \textbf{Dynamic Scoring:} Implemented \texttt{qed\_drug\_likeness} and \texttt{molecular\_weight\_filter} during the run to counteract the observed drift toward complexity.
    \item \textbf{Pareto Optimization:} The system is currently exploring the Pareto frontier, with 52.5\% of candidates deemed Pareto-efficient [\textcolor{myplanner}{this piece of data is calculated via thresholds proposed by the planner}], balancing the conflict between potency and safety.
    \item \textbf{Metabolic stability objective}: Add a metabolic stability prediction objective. This could use CYP450 interaction prediction or structural alert-based scoring. This would prevent the optimizer from generating compounds with excellent in vitro activity but poor in vivo stability.
\end{itemize}

\textbf{\textcolor{myimplementer}{Implementer}: implementing the metabolic stability objective ...}
\end{tcolorbox}

\subsection{Nanobody Design}
\label{subsec:nanobody history}
\begin{tcolorbox}[colback={rgb,255:red,217;green,247;blue,243}, colframe={rgb,255:red,114;green,182;blue,161}, breakable, title=\text{Analyzer's, Planner's, and Implementer's output in the Co-pilot workflow of \name}]

\textbf{Inputs: Received initial population with 100 candidates. The goal is to design high-affinity nanobodies that bind to PD-L1 for cancer immunotherapy.}\\

\tcbline
\textbf{\large Iteration 1}

\textbf{\textcolor{myanalyzer}{Analyzer}}
\paragraph{Overview}
\begin{itemize}
    \item Iteration 1 generated 100 nanobody candidates with 100\% validity.
    \item \textbf{Key outcome:} Strong overall binding confidence with mean protein\_iptm of $0.809$ (best $0.918$) and mean ptm of $0.855$ (best $0.935$), placing most candidates in a regime associated with reliable complex formation.
\end{itemize}

\paragraph{Performance Analysis}
\begin{itemize}
    \item \textbf{Binding Affinity:} Mean protein\_iptm $0.809$, mean ptm $0.855$, indicating strong predicted binding.
    \item \textbf{Interface Geometry:} Mean min\_pae $1.93$ \AA, though a small number of outliers exhibit substantially higher uncertainty ($>10$ \AA). Hydrogen bonds range from 1 to 13, salt bridges from 0 to 9.
    \item \textbf{Buried Surface Area:} Mean $\Delta$SASA $\sim$894 \AA$^2$ (range 504--1263 \AA$^2$), indicating moderate-to-strong interfaces with significant spread.
    \item \textbf{Developability:} Mean liability\_score $145.8$ (best $120$), indicating multiple sequence-level liabilities per design.
\end{itemize}

\paragraph{Key Issues and Risks}
\begin{itemize}
    \item \textbf{Elevated Developability Liabilities:} High liability scores pose downstream risks for expression and manufacturability and may indirectly depress structural confidence.
    \item \textbf{Objective Tension:} Aggressively increasing interface contacts via charged or aromatic residues may conflict with developability and can destabilize loop geometry if overpacked.
\end{itemize}

\noindent
\textbf{Human [after inspecting the predicted structures]:} Many designs show low confidence, specifically in the CDR regions, despite good global metrics. Let's add: (1) pLDDT-based structural confidence metrics for both the full binder and CDR regions; (2) a CDR3 alpha-helix hard constraint; (3) ProteinMPNN-based sequence--structure compatibility scores, since ABMPNN may not generalize well to this non-standard CDR3 length.

\paragraph{\textcolor{myplanner}{Planner}}
\begin{itemize}
    \item Maximize \texttt{protein\_iptm}, and \texttt{ptm} to ensure binding affinity and complex confidence (\textbf{Agent proposed}).
    \item Minimize \texttt{min\_pae} and maximize \texttt{hbonds}, \texttt{saltbridges}, \texttt{delta\_sasa} to ensure interface quality (\textbf{Agent proposed}).
    \item Minimize \texttt{liability\_score} to maintain developability (\textbf{Agent proposed}).
    \item Maximize \texttt{binder\_plddt} (weight $=7.0$) and \texttt{cdr\_plddt} (weight $=8.0$) to ensure structural confidence of the nanobody fold and CDR binding regions (\textbf{Human proposed}).
    \item Minimize \texttt{mpnn\_score} and maximize \texttt{mpnn\_expected\_recovery} using vanilla ProteinMPNN (not ABMPNN) to evaluate sequence--structure compatibility for the non-standard CDR3 length (\textbf{Human proposed}).
    \item Maximize \texttt{cdr3\_has\_good\_alpha\_helix} as a hard constraint: CDR3 must contain a well-formed alpha helix (total helix $\geq 9$ aa, fraction $> 30\%$, helix starts within first $35\%$ of CDR3, longest continuous helix $\geq 7$ aa). Candidates failing this are rejected (\textbf{Human proposed}).
\end{itemize}

\noindent
\textbf{Human: Looks good. The alpha helix constraint is critical, making it a hard filter so that any candidate without a good CDR3 helix is rejected outright.} \\

\textbf{\textcolor{myimplementer}{Implementer}: implementing \texttt{binder\_plddt}, \texttt{cdr\_plddt}, \texttt{mpnn\_score}, \texttt{mpnn\_expected\_recovery}, and \texttt{cdr3\_has\_good\_alpha\_helix} scorers ...}

\tcbline
\textbf{\large Iteration 2}

\textbf{\textcolor{myanalyzer}{Analyzer}}
\paragraph{Overview}
\begin{itemize}
    \item Iteration 2 processed 100 candidates with no calculation failures. All candidates satisfy the hard CDR3 helix constraint (\texttt{cdr3\_has\_good\_alpha\_helix} $= 1.0$ for $100/100$).
    \item The optimization is currently in a \textbf{structure-convergence phase}: binder\_plddt is already high and stable, while cdr\_plddt is improving but remains the primary bottleneck.
\end{itemize}

\paragraph{Performance Analysis}
\begin{itemize}
    \item \textbf{Structural Confidence:} binder\_plddt: $87.37 \pm 1.51$ (uniformly high); cdr\_plddt: $74.66 \pm 3.26$ (improving but still suboptimal, $\sim$15-point gap from binder\_plddt). [\textcolor{myanalyzer}{analyzer identified the CDR pLDDT bottleneck}]
    \item \textbf{Binding Affinity:} protein\_iptm: $0.823 \pm 0.078$ (best $0.920$), min\_pae: $1.80 \pm 1.24$ \AA\ (best $0.726$ \AA).
    \item \textbf{Sequence--Structure Compatibility:} mpnn\_expected\_recovery $\approx 0.37$, mpnn\_score $\approx 1.66$---reasonable but suboptimal, indicating room for more natural, structurally consistent sequences.
\end{itemize}

\paragraph{Key Issues and Risks}
\begin{itemize}
    \item \textbf{CDR-Level Structural Uncertainty:} The dominant limitation is residual uncertainty within CDR backbones, especially CDR3 helix positioning, termination geometry, and excessive local charge density within CDR3 helices.
    \item \textbf{Saturation of Binary Constraint:} The \texttt{cdr3\_has\_good\_alpha\_helix} objective is at ceiling for all designs ($1.0$), providing no gradient for further improvement while still consuming optimization weight.
    \item \textbf{No Epitope Engagement Control:} Current objectives ensure structural quality and binding confidence, but do not verify whether the CDR regions are engaging the functionally relevant PD-L1 epitope that blocks PD-1 binding.
\end{itemize}

\noindent
\textbf{Human [after examining epitope engagement in top candidates]:} The CDR3 helix is forming correctly, but we need to ensure it's actually engaging the right epitope. PD-L1 hotspot residues at the PD-1 binding interface have been characterized---let's add Germinal-style hotspot contact metrics to verify that CDR and CDR3 residues are making functional contacts with these hotspots.

\paragraph{\textcolor{myplanner}{Planner}}
\begin{itemize}
    \item Maximize \texttt{cdr3\_helix\_amphipathicity}: quantify amphipathic character of CDR3 helix using Eisenberg hydrophobicity scale and hydrophobic moment calculation, since amphipathic helices are more effective at protein--protein interfaces (\textbf{Agent proposed}).
    \item Maximize \texttt{cdr3\_helix\_interface\_fraction}: fraction of nanobody--PD-L1 interface area contributed by CDR3 helix residues (via per-residue buried SASA); values $\geq 0.35$ indicate the helix is a dominant contributor to binding (\textbf{Agent proposed}).
    \item Maximize \texttt{cdr\_hotspot\_contacts}: unique CDR residues contacting PD-L1 hotspot region via Germinal-style calculation (hotspot expansion at $5.3$ \AA\ CA--CA $+$ key atom contacts at $6.0$ \AA) (\textbf{Human proposed}).
    \item Maximize \texttt{cdr3\_hotspot\_contacts}: unique CDR3 residues contacting PD-L1 hotspots, emphasizing engagement by the dominant paratope region (\textbf{Human proposed}).
\end{itemize}

\noindent
\textbf{Human: Good plan. Make sure the hotspot residues are defined based on the PD-1/PD-L1 interaction interface.} \\

\textbf{\textcolor{myimplementer}{Implementer}: implementing \texttt{cdr3\_helix\_amphipathicity}, \texttt{cdr3\_helix\_interface\_fraction}, \texttt{cdr\_hotspot\_contacts}, and \texttt{cdr3\_hotspot\_contacts} scorers ...}

\end{tcolorbox}

\begin{tcolorbox}[colback={rgb,255:red,250;green,230;blue,222}, colframe={rgb,255:red,233;green,150;blue,117}, breakable, title=\text{Analyzer's, Planner's, and Implementer's output in the Semi-pilot workflow of \name}]

\textbf{Inputs: Received initial population with 100 candidates. The goal is to design high-affinity nanobodies that bind to PD-L1 for cancer immunotherapy.}\\

\tcbline
\textbf{\large Iteration 1}

\textbf{\textcolor{myanalyzer}{Analyzer}}
\paragraph{Overview}
\begin{itemize}
    \item Iteration 1 produced 100 candidates targeting PD-L1 binding with 100\% validity.
    \item \textbf{Key outcome:} Binder pLDDT is decent ($84.7$--$90.4$), but CDR3 pLDDT is notably weaker ($65.3$--$74.1$). ipTM remains low on average but with strong outliers (best $0.87$).
\end{itemize}

\paragraph{Performance Analysis}
\begin{itemize}
    \item \textbf{Structural Confidence:} Binder pLDDT is decent (AF3: $84.7 \pm 3.9$; Boltz: $90.4 \pm 2.9$), but \textbf{CDR3 pLDDT is notably weaker} (AF3: $65.3 \pm 11.1$; Boltz: $74.1 \pm 11.2$), representing a $\sim$20-point gap from the binder backbone. [\textcolor{myanalyzer}{analyzer identified the CDR3 structural bottleneck}]
    \item \textbf{Interface Quality:} H-bonds $3.7 \pm 2.0$, salt bridges $2.31 \pm 1.32$, $\Delta$SASA $775 \pm 135$ \AA$^2$ (best $\sim$1206 \AA$^2$)---broadly moderate.
    \item \textbf{Binding Affinity:} ipTM remains low on average (mean $0.439$) but with strong outliers (best $0.87$), indicating that high-affinity binding is achievable within the current search space.
\end{itemize}

\paragraph{Key Issues and Risks}
\begin{itemize}
    \item \textbf{CDR3 Structural Uncertainty:} CDR3 pLDDT lags significantly behind the binder backbone, suggesting disordered or conformationally ambiguous CDR3 regions in many designs.
    \item \textbf{Low Average Interface Confidence:} Most candidates have weak predicted interface quality, though the high-performing tail demonstrates that the search space contains promising solutions.
\end{itemize}

\noindent
\textbf{Human: Prioritize CDR3 refinement to improve structural pLDDT.}

\paragraph{Planner (refined)}
\begin{itemize}
    \item Add \texttt{cdr3\_has\_good\_alpha\_helix} as a hard constraint: CDR3 must contain a well-formed alpha helix (total helix $\geq 9$ aa, fraction $> 30\%$, helix starts within first $40\%$ of CDR3, longest continuous helix $\geq 7$ aa). Candidates failing this are rejected.
    \item Add \texttt{cdr3\_helix\_rule\_margin} (minimize): quantify how comfortably each design satisfies the strict CDR3 helix rules, encouraging margin from failure boundaries.
    \item Add \texttt{cdr3\_helix\_sequence\_propensity} (maximize): sequence-based scorer that evaluates CDR3 helix-forming residue composition using Chou--Fasman propensity scales, penalizing helix breakers (Pro, Gly) within the predicted helix core.
    \item Add \texttt{mpnn\_score} (minimize) and \texttt{mpnn\_expected\_recovery} (maximize) using vanilla ProteinMPNN to evaluate sequence--structure compatibility (\textbf{Human proposed}).
    \item Increase \texttt{cdr3\_plddt} weight from $10.0 \to 12.0$ and \texttt{cdr\_plddt} weight from $8.0 \to 9.0$.
\end{itemize}

\textbf{\textcolor{myimplementer}{Implementer}: implementing \texttt{cdr3\_has\_good\_alpha\_helix}, \texttt{cdr3\_helix\_rule\_margin}, \texttt{cdr3\_helix\_sequence\_propensity}, \texttt{mpnn\_score}, and \texttt{mpnn\_expected\_recovery} scorers ...}

\tcbline
\textbf{\large Iteration 2}

\textbf{\textcolor{myanalyzer}{Analyzer}}
\paragraph{Overview}
\begin{itemize}
    \item CDR3 $\alpha$-helix pass rate: $65\%$. The helix constraint is partially enforced but not yet fully satisfied across the population.
    \item Marginal ipTM and pTM decline observed, suggesting that introducing the helix constraint created slight tension with binding affinity objectives.
\end{itemize}

\paragraph{Performance Analysis}
\begin{itemize}
    \item \textbf{Helix Quality:} Helix rule margin (lower is better) at mean $0.566$, indicating many designs sit near constraint boundaries rather than well-centered within allowed ranges.
    \item \textbf{Structural Confidence:} CDR3 pLDDT improved modestly, suggesting helix formation helps but is not yet sufficient to fully resolve CDR3 disorder.
    \item \textbf{Interface Physics:} H-bonds, salt bridges, and $\Delta$SASA declined slightly, suggesting less robust physical packing despite gains in structural regularity.
\end{itemize}

\paragraph{Key Issues and Risks}
\begin{itemize}
    \item \textbf{Incomplete Helix Enforcement:} $35\%$ of candidates still fail the CDR3 helix constraint, indicating the optimizer needs stronger selection pressure.
    \item \textbf{Binding--Structure Trade-off:} The marginal ipTM/pTM decline suggests that helix enforcement competes with interface optimization; epitope-directed objectives may help align both goals.
\end{itemize}

\noindent
\textbf{Human: CDR3 must contact the epitope hotspot residues.}

\paragraph{\textcolor{myplanner}{Planner}}
The Planner operationalizes the hotspot engagement directive: [\textcolor{myplanner}{planner converted high-level epitope requirement into concrete objectives}]
\begin{itemize}
    \item Add \texttt{cdr3\_helix\_hotspot\_contacts} (maximize): count CDR3 helix residues contacting PD-L1 hotspot region, ensuring the $\alpha$-helix directly engages the functional epitope that blocks PD-1 binding.
    \item Add \texttt{cdr3\_helix\_hotspot\_contact\_fraction} (maximize): fraction of total interface contacts contributed by CDR3 helix--hotspot interactions (via heavy-atom contacts at $4.5$ \AA\ cutoff); higher values indicate the helix is the dominant epitope-engagement element.
    \item Increase \texttt{af3\_protein\_iptm} weight to reinforce interface quality and counteract the observed regression.
    \item Increase helix constraint weights (\texttt{cdr3\_has\_good\_alpha\_helix} from $6.0 \to 8.0$, \texttt{cdr3\_helix\_rule\_margin} from $4.0 \to 6.0$) to push designs toward comfortable compliance.
\end{itemize}

\noindent
\textbf{Human: Good plan, proceed.} \\

\textbf{\textcolor{myimplementer}{Implementer}: implementing \texttt{cdr3\_helix\_hotspot\_contacts} and \texttt{cdr3\_helix\_hotspot\_contact\_fraction} scorers ...}

\tcbline
\textbf{\large Iteration 3+}

\textbf{\textcolor{myanalyzer}{Analyzer}: All key metrics met; CDR3 helix formation, epitope engagement, and structural confidence converging. Design finalized.}

\end{tcolorbox}

\begin{tcolorbox}[colback={rgb,255:red,255;green,245;blue,200}, colframe={rgb,255:red,220;green,180;blue,60}, breakable, title=\text{Analyzer's and Planner's output in the Autopilot workflow of \name}]

\textbf{Inputs: Received initial population with 100 candidates. The goal is to design high-affinity nanobodies that bind to PD-L1 for cancer immunotherapy.}\\

\tcbline
\textbf{\large Iteration 1}

\textbf{\textcolor{myanalyzer}{Analyzer}}
\paragraph{Overview}
\begin{itemize}
    \item Iteration 1 produced 100 candidates targeting PD-L1 binding with 100\% validity.
    \item \textbf{Key outcome:} ipTM remains low on average (mean $0.405$, best $0.78$). Interface interactions are moderate but with significant spread, and developability liabilities are elevated.
\end{itemize}

\paragraph{Performance Analysis}
\begin{itemize}
    \item \textbf{Binding Confidence:} protein\_iptm mean $0.405$ (best $0.78$); ptm mean $0.62$. Most candidates fall below the $0.7$ high-confidence ipTM threshold, though the tail shows promising solutions.
    \item \textbf{Interface Interactions:} H-bonds $2.7 \pm 1.5$; salt bridges $1.8 \pm 1.3$ (some candidates have $0$); $\Delta$SASA $707 \pm 130$ \AA$^2$.
    \item \textbf{Developability:} Liability score mean $168$ (best $140$, worst $220$), indicating multiple sequence liabilities not adequately penalized.
\end{itemize}

\paragraph{Key Issues and Risks}
\begin{itemize}
    \item \textbf{Overemphasis on Global Confidence:} Multiple high-weight objectives (\texttt{protein\_iptm}, \texttt{ptm}) all reward structure prediction confidence, leaving interface interactions and developability under-represented in the objective landscape. [\textcolor{myanalyzer}{analyzer identified weight imbalance as a key inefficiency}]
    \item \textbf{Developability Under-prioritized:} Liability scores remain high; the current weight ($2.0$) is insufficient to drive down risks in parallel with binding optimization.
    \item \textbf{Weak Electrostatic Complementarity:} Some candidates have $0$ salt bridges, indicating limited charge-driven interactions at the interface.
\end{itemize}

\paragraph{\textcolor{myplanner}{Planner}}
Based on the analysis, the Planner proposes weight rebalancing and two new objectives:
\begin{itemize}
    \item \textbf{De-prioritize global confidence:} Decrease \texttt{ptm} weight from $3.0 \to 1.0$, as global fold confidence is secondary to interface quality for protein--protein interactions.
    \item \textbf{Strengthen developability pressure:} Increase \texttt{liability\_score} weight from $2.0 \to 3.0$ to drive down liabilities without compromising binding quality.
    \item \textbf{Promote interface quality:} Increase \texttt{saltbridges} weight from $1.0 \to 2.0$ to discourage zero-electrostatics solutions; increase \texttt{delta\_sasa} weight from $1.0 \to 2.0$ to promote larger, more interactive interfaces.
    \item \textbf{Target conserved epitope:} Add \texttt{epitope\_conservation\_score} (maximize) to measure conservation of contacted PD-L1 epitope positions using MSA-derived Shannon entropy, encouraging binding to evolutionarily conserved residues to reduce escape risk.
    \item \textbf{Assess geometric fit:} Add \texttt{interface\_shape\_complementarity} (maximize) to compute the Lawrence \& Colman $S_c$ statistic ($0$--$1$ scale; $S_c \geq 0.7$ indicates good geometric surface complementarity between nanobody and target).
\end{itemize}

\textbf{\textcolor{myimplementer}{Implementer}: implementing \texttt{epitope\_conservation\_score} and \texttt{interface\_shape\_complementarity} scorers; adjusting objective weights ...}

\tcbline
\textbf{\large Iteration 2}

\textbf{\textcolor{myanalyzer}{Analyzer}}
\paragraph{Overview}
\begin{itemize}
    \item Developability improved meaningfully: liability score dropped by $-14.7$ with tighter spread. Salt bridges improved ($+0.4$). However, H-bonds declined ($-0.3$) and $\Delta$SASA declined ($-20$ \AA$^2$).
    \item A trade-off tension emerged: gains in developability and electrostatics coincided with regressions in hydrogen bonding and buried surface area.
\end{itemize}

\paragraph{Performance Analysis}
\begin{itemize}
    \item \textbf{Binding Confidence:} protein\_iptm essentially flat ($-0.001$); ptm marginal regression. Best performers remain strong (best ipTM $\sim 0.79$).
    \item \textbf{Developability:} Liability score improved ($-14.7$) with much tighter distribution ($153 \pm 3.4$ vs $168 \pm 10.6$), validating the weight increase.
    \item \textbf{Interface Regression:} H-bonds declined ($-0.3$) and $\Delta$SASA declined ($-20$ \AA$^2$), suggesting the search is trading interfacial packing for sequence cleanliness.
    \item \textbf{Electrostatics Gain:} Salt bridges improved ($+0.4$), confirming the weight adjustment worked.
\end{itemize}

\paragraph{Key Issues and Risks}
\begin{itemize}
    \item \textbf{Trade-off Tension:} Improvements in liability and salt bridges came at the cost of H-bonds and buried area, hinting the search is nudging toward electrostatics and sequence cleanliness while sacrificing interface depth.
    \item \textbf{Early Convergence Risk:} Liability score and binding confidence distributions narrowed, suggesting the population may be converging without maximizing interface quality.
    \item \textbf{Confidence--Interaction Imbalance:} Confidence-oriented objectives still dominate the weight budget, leaving limited pressure for H-bond and buried-area recovery.
\end{itemize}

\paragraph{\textcolor{myplanner}{Planner}}
\begin{itemize}
    \item \textbf{Rebalance confidence vs.\ interactions:} Decrease \texttt{protein\_iptm} weight from $3.0 \to 2.0$ to free optimization budget for interface-quality objectives.
    \item \textbf{Recover interface packing:} Increase \texttt{hbonds} weight from $3.0 \to 4.0$ and \texttt{delta\_sasa} weight from $2.0 \to 3.0$ to prioritize dense, well-packed interfaces.
    \item \textbf{Sharpen interface precision:} Increase \texttt{min\_pae} weight from $3.0 \to 4.0$ to penalize loose interfaces more aggressively.
    \item \textbf{Ensure functional epitope engagement:} Add \texttt{cdr3\_hotspot\_contacts} (maximize) to count CDR3 residues contacting PD-L1 hotspot region via Germinal-style calculation (hotspot expansion at $5.3$ \AA\ CA--CA $+$ key atom contacts at $6.0$ \AA), ensuring the dominant paratope region engages the PD-1 binding interface.
    \item \textbf{Improve interface polar satisfaction:} Add \texttt{unsatisfied\_polar\_penalty} (minimize) to penalize buried polar atoms at the interface that lack hydrogen bond partners, counteracting the observed H-bond regression.
\end{itemize}

\textbf{\textcolor{myimplementer}{Implementer}: implementing \texttt{cdr3\_hotspot\_contacts} and \texttt{unsatisfied\_polar\_penalty} scorers; adjusting objective weights ...}

\tcbline
\textbf{\large Iteration 3+}

\textbf{\textcolor{myanalyzer}{Analyzer}: Interface metrics recovering; epitope conservation and shape complementarity improving. Weight adjustments and hotspot engagement optimization continue. Optimization proceeds autonomously ...}

\end{tcolorbox}

\subsection{Functional DNA Sequence Design}
\label{subsec:dna history}

\begin{tcolorbox}[colback={rgb,255:red,217;green,247;blue,243}, colframe={rgb,255:red,114;green,182;blue,161}, breakable, title=Analyzer's output in the Co-pilot workflow of \name]
\textbf{Inputs: Received initial population with 5000 candidates (random DNA sequences with lengh 200 bp). The inputs are same for three modes.}\\

\textbf{Iteration 1 \textcolor{myanalyzer}{(Analyzer):}}\\

\paragraph{Overview}
\begin{itemize}
    \item Iteration~1 reduced 5000 inputs to a selected population of 282 candidates aimed at 200~bp HepG2-specific enhancers.
\end{itemize}

\paragraph{Performance Analysis}
\begin{itemize}
    \item Clear progress was observed across all objectives relative to the previous pool.
    \item The population-level specificity margin showed a rough but consistent improvement.
    \item \textbf{Tails:} Some candidates still exhibit high off-target activity.
\end{itemize}

\paragraph{Issues and Concerns}
\begin{itemize}
    \item The objectives do not include motif enrichment information, which is critical and should be prioritized.
    \item The objectives do not include stability information, which is critical and should be prioritized.
    \item Off-target means remain positive, implying that many sequences still drive expression in K562/SKNSH; specificity is not yet consistently strong across the set.
    \item Variance contraction, especially in off-target scores, may signal narrowing exploration and potential diversity loss in later iterations.
\end{itemize}

\noindent
\textbf{Human: Problems of original issues and recommendations:} No explicit consideration of transcription-factor-related criteria.

\paragraph{Strategic Recommendations (after human modification)}
\begin{itemize}
    \item Add objectives that enrich motif information. Avoid overly complicated solutions, as this is a serious problem (\textbf{Human proposed}).
    \item Add objectives that improve sequence stability. Avoid overly complicated solutions, as this is a serious problem (\textbf{Human proposed}).
    \item Add a composite specificity objective to sharpen selection pressure: maximize specificity margin,
    \[
        \text{HepG2} - \alpha \cdot \max(\text{K562}, \text{SKNSH}),
    \]
    starting with $\alpha = 0.5$--$1.0$, or alternatively
    \[
        \text{HepG2} - \alpha \cdot (\text{K562} + \text{SKNSH}) / 2
    \]
    (\textbf{Agent proposed}).
    \item If a weighted scheme is used, slightly increase emphasis on HepG2 until the mean reaches $\geq 1.1$--$1.2$, then tighten off-target penalties in the following round (\textbf{Agent proposed}).
    \item Maintain all three base objectives; do not remove any off-target objective at this stage. Consider logging the composite margin for monitoring even if it is not optimized directly (\textbf{Agent proposed}).
\end{itemize}

\textbf{Iteration 2 \textcolor{myanalyzer}{(Analyzer):}}\\
\paragraph{Overview}
\begin{itemize}
    \item Iteration 2 produced 193 candidates...
    \item \textbf{Key outcome:} A large gain in HepG2 specificity margin, with small regressions in motif enrichment and stability penalty...
\end{itemize}

\paragraph{Performance Analysis}
\begin{itemize}
    \item \textbf{HepG2 specificity margin:} Mean $0.9367 \pm 1.2182$, up by $+0.4264$ from the last iteration. The best candidate achieves $7.8889$ (an extreme outlier; $\sim$5.7 SD above the mean), while the worst is slightly negative ($-0.0849$). Increased variance suggests broader exploration and the emergence of highly specific designs.
    \item \textbf{Hepatocyte motif enrichment:} Mean $0.9322 \pm 0.0682$, down by $-0.0212$. The best remains $1.0000$, while the worst value of $0.7132$ indicates a tail of candidates with weaker hepatocyte motif presence. Despite the dip, overall enrichment remains high.
    \item \textbf{DNA stability penalty:} Mean $0.0449 \pm 0.0407$, slightly worse by $+0.0061$. The best is $0.0000$, while the worst value of $0.2050$ indicates a minority of sequences with notable GC imbalance or long homopolymers. Overall, the penalty remains low for most candidates.
\end{itemize}

\paragraph{Issues and Concerns}
\begin{itemize}
    \item \textbf{Trade-off drift:} The strong gain in specificity margin coincides with modest erosion in motif enrichment and stability, suggesting that the optimizer is prioritizing the composite margin at the expense of biological motif features and manufacturability.
    \item \textbf{Tail risks:} The presence of candidates with negative margin and higher stability penalties (up to $0.2050$) could dilute overall set quality.
    \item \textbf{Variance:} Elevated variance in the margin and increased spread in motif enrichment imply that the search has not yet converged and is still exploring widely.
\end{itemize}

\paragraph{Human: We still need to consider biology-driven optimization.}
\paragraph{Strategic Recommendations}
\begin{itemize}
    \item Add objectives that enrich motif information. Avoid overly complicated solutions, as this is a serious issue \textbf{(Human proposed)}.
    \item Add objectives that improve sequence stability. Avoid overly complicated solutions, as this is a serious issue \textbf{(Human proposed)}.
    \item Introduce a hinge-style objective or thresholded penalty for \texttt{hepatocyte\_motif\_enrichment\_score} below $\sim$0.93--0.95 to prevent further regression while preserving high-specificity gains \textbf{(Human proposed)}.
    \item In addition to (or instead of) the composite margin, add objectives to maximize \texttt{HepG2\_MPRA} and minimize $\max(\texttt{K562\_MPRA}, \texttt{SKNSH\_MPRA})$ (or minimize K562 and SKNSH separately). This clarifies trade-offs and helps avoid solutions that game the margin via non-motif features \textbf{(Human proposed)}.
    \item Add a soft constraint or hinge penalty when \texttt{dna\_stability\_gc\_homopolymer\_penalty} $> 0.08$ to suppress the worst cases without excessively shifting the mean \textbf{(Agent proposed)}.
\end{itemize}

\textbf{After optimization, selection agent or human expert believe the optimization can stop, we reach the criteria.}
\end{tcolorbox}

\begin{tcolorbox}[colback={rgb,255:red,250;green,230;blue,222}, colframe={rgb,255:red,233;green,150;blue,117}, breakable, title=Planner's output in the Semi-pilot workflow of \name]
\paragraph{Initial Objectives \textcolor{myplanner}{(Planner):}}
\begin{enumerate}
    \item \texttt{dna\_hepg2\_enhancer\_MPRA\_expression} 
    (type = ``candidate-wise'', weight = None, maximize)

    \noindent
    \textbf{Description:} HepG2 MPRA expression score (value range: $-\infty$ to $+\infty$). MPRA (Massively Parallel Reporter Assay) quantifies gene expression driven by DNA sequences using massively parallel sequencing to measure how effectively different sequences drive reporter gene expression. This score evaluates DNA enhancer sequences based on their predicted expression levels in HepG2 cells (human hepatocellular carcinoma cell line). Higher scores indicate stronger enhancer activity and greater gene expression in liver-like cellular contexts, while lower scores suggest weaker enhancer activity or potential silencing effects. This metric is essential for designing DNA sequences that function specifically in hepatic environments or liver-related therapeutic applications.

    \item \texttt{dna\_k562\_enhancer\_MPRA\_expression} 
    (type = ``candidate-wise'', weight = None, minimize)

    \noindent
    \textbf{Description:} K562 MPRA expression score (value range: $-\infty$ to $+\infty$). MPRA (Massively Parallel Reporter Assay) quantifies gene expression driven by DNA sequences using massively parallel sequencing to measure how effectively different sequences drive reporter gene expression. This score evaluates DNA enhancer sequences based on their predicted expression levels in K562 cells (human erythroleukemic cell line). Higher scores indicate stronger enhancer activity and greater gene expression in hematopoietic cellular contexts, while lower scores suggest weaker enhancer activity or potential silencing effects. This metric is crucial for designing DNA sequences that should avoid activity in blood cell lineages or hematological contexts when cell-type specificity is desired.

    \item \texttt{dna\_sknsh\_enhancer\_MPRA\_expression} 
    (type = ``candidate-wise'', weight = None, minimize)

    \noindent
    \textbf{Description:} SKNSH MPRA expression score (value range: $-\infty$ to $+\infty$). MPRA (Massively Parallel Reporter Assay) quantifies gene expression driven by DNA sequences using massively parallel sequencing to measure how effectively different sequences drive reporter gene expression. This score evaluates DNA enhancer sequences based on their predicted expression levels in SKNSH cells (human neuroblastoma cell line). Higher scores indicate stronger enhancer activity and greater gene expression in neuronal cellular contexts, while lower scores suggest weaker enhancer activity or potential silencing effects. This metric is essential for designing DNA sequences that should avoid activity in neural tissues when liver specificity is required.
\end{enumerate}

\textbf{Human: Looks good.}

\paragraph{Iteration 2 Objectives with Human Feedback \textcolor{myplanner}{(Planner):}}

\begin{enumerate}
    \item \texttt{hepg2\_specificity\_composite} 
    (type = ``candidate-wise'', weight = None, maximize)

    \noindent
    \textbf{Description:} Composite HepG2-specific activity score coupling on- and off-target MPRA predictions. The value range is $(-\infty, +\infty)$. Higher scores are better, indicating strong predicted HepG2 activity while penalizing any positive off-target activity in K562 and SKNSH. Off-target scores $\leq 0$ incur no penalty.

    \item \texttt{liver\_motif\_enrichment\_sumlogodds} 
    (type = ``candidate-wise'', weight = None, maximize)

    \noindent
    \textbf{Description:} Motif enrichment score for key hepatocyte transcription factors based on PWM scanning. For each candidate sequence $s$, both strands are scanned against the following JASPAR PWMs: HNF4A (MA0114.5), HNF1A (MA0046.3), CEBPA (MA0102.4), FOXA1 (MA0148.4), FOXA2 (MA0047.3), RXRA (MA0511.2), CEBPB (MA0466.3), and ONECUT1/HNF6 (MA0799.1). A uniform background is assumed ($A = C = G = T = 0.25$). All motif hits with log-odds scores greater than or equal to the threshold corresponding to a $p$-value $\leq 10^{-4}$ (e.g., computed via \texttt{MOODS} or \texttt{gimme-motifs}) are retained. The score is defined as
    \[
        \text{score} = \sum_{\text{hits}} \left( \text{log\_odds}_{\text{hit}} - \text{log\_odds}_{\text{threshold, motif}} \right).
    \]
    The value range is $[0, +\infty)$. Higher scores indicate a greater number and/or stronger liver transcription factor binding sites, which are associated with enhancer activity in HepG2.

    \item \texttt{dna\_sequence\_stability\_penalty} 
    (type = ``candidate-wise'', weight = None, minimize)

    \noindent
    \textbf{Description:} A simple stability-oriented penalty combining GC balance and homopolymer runs. For a 200~bp sequence $s$, let
    \[
        \text{GC\_fraction} = \frac{\#(G \text{ or } C)}{200}, \quad
        \text{gc\_dev} = \left| \text{GC\_fraction} - 0.45 \right|.
    \]
    Let $L$ denote the longest homopolymer length across $\{A,T,C,G\}$, and define
    \[
        \text{homopolymer\_excess} = \max(0, L - 5).
    \]
    The penalty score is computed as
    \[
        \text{score} = 1.0 \cdot \text{gc\_dev} + 0.2 \cdot \text{homopolymer\_excess}.
    \]
    The value range is $[0, +\infty)$. Lower scores are better, reflecting GC content near 45\% and avoidance of long homopolymers ($>5$), which improves sequence stability and synthesis or handling robustness.
\end{enumerate}
\textbf{Human: Looks good.}

\paragraph{Iteration 3 Objectives with Human Feedback \textcolor{myplanner}{(Planner):}}
\begin{enumerate}
    \item \texttt{hepg2\_specificity\_composite\_tau0p1} 
    (type = ``candidate-wise'', weight = None, maximize)

    \noindent
    \textbf{Description:} Composite HepG2-specific activity score coupling on- and off-target MPRA predictions with a small noise threshold. For a candidate sequence $s$, 
    and define $\mathrm{ReLU}(x) = \max(0, x)$. With $\tau = 0.10$, $\alpha = 1.3$, and $\beta = 1.3$, the score is computed as
    \[
        \text{score} = H - \alpha \cdot \mathrm{ReLU}(K - \tau) - \beta \cdot \mathrm{ReLU}(N - \tau).
    \]
    The value range is $(-\infty, +\infty)$. Higher scores are better, indicating strong predicted HepG2 activity while ignoring negligible off-target noise ($\leq 0.10$) and penalizing only meaningful positive off-target activity.

    \item \texttt{liver\_motif\_enrichment\_sumlogodds\_p1e3} 
    (type = ``candidate-wise'', weight = None, maximize)

    \noindent
    \textbf{Description:} Motif enrichment score for hepatocyte transcription factors based on PWM scanning with a more permissive threshold to reduce zero inflation. For each candidate sequence $s$, both strands are scanned against the following JASPAR PWMs: HNF4A (MA0114.5), HNF1A (MA0046.3), CEBPA (MA0102.4), FOXA1 (MA0148.4), FOXA2 (MA0047.3), RXRA (MA0511.2), CEBPB (MA0466.3), and ONECUT1/HNF6 (MA0799.1). A uniform background is assumed ($A = C = G = T = 0.25$). All motif hits with log-odds scores greater than or equal to the motif-specific threshold corresponding to a $p$-value $\leq 10^{-3}$ (e.g., computed via \texttt{MOODS} or \texttt{gimme-motifs}) are retained. The score is defined as
    \[
        \text{score} = \sum_{\text{hits}} \left( \text{log\_odds}_{\text{hit}} - \text{log\_odds}_{\text{threshold, motif}} \right).
    \]
    The value range is $[0, +\infty)$. Higher scores indicate a greater number and/or stronger hepatocyte transcription factor binding sites, promoting robust HepG2 enhancer architecture.

    \item \texttt{dna\_sequence\_stability\_violation\_hinge} 
    (type = ``candidate-wise'', weight = None, minimize)

    \noindent
    \textbf{Description:} Hinge-style penalty focusing on unstable outliers based on GC balance and homopolymer runs. For a 200~bp sequence $s$, let
    \[
        \text{GC\_fraction} = \frac{\#G + \#C}{200}, \quad
        \text{gc\_dev} = \left| \text{GC\_fraction} - 0.45 \right|.
    \]
    Let $L$ denote the longest homopolymer length across $\{A, T, C, G\}$, and define
    \[
        \text{homopolymer\_excess} = \max(0, L - 5).
    \]
    Define the base penalty
    \[
        P = \text{gc\_dev} + 0.2 \cdot \text{homopolymer\_excess}.
    \]
    The final score is computed as
    \[
        \text{score} = \max(0, P - 0.15).
    \]
    The value range is $[0, +\infty)$. Lower scores are better: a score of $0$ indicates acceptable stability, while positive values target sequences with substantial GC imbalance or long homopolymers.
\end{enumerate}
\textbf{Human: Looks good. After optimization, selection agent or human expert believe the optimization can stop, we reach the criteria.}

\end{tcolorbox}

\begin{tcolorbox}[colback={rgb,255:red,255;green,245;blue,200}, colframe={rgb,255:red,220;green,180;blue,60}, breakable, title=Planner's output in the Autopilot workflow of \name with analyzer's feedback]
\paragraph{Initial Objectives \textcolor{myplanner}{(Planner):}}
\begin{enumerate}
    \item \texttt{dna\_hepg2\_enhancer\_MPRA\_expression} 
    (type = ``candidate-wise'', weight = None, maximize)

    \noindent
    \textbf{Description:} HepG2 MPRA expression score (value range: $-\infty$ to $+\infty$). MPRA (Massively Parallel Reporter Assay) quantifies gene expression driven by DNA sequences using massively parallel sequencing to measure how effectively different sequences drive reporter gene expression. This score evaluates DNA enhancer sequences based on their predicted expression levels in HepG2 cells (human hepatocellular carcinoma cell line). Higher scores indicate stronger enhancer activity and greater gene expression in liver-like cellular contexts, while lower scores suggest weaker enhancer activity or potential silencing effects. This metric is essential for designing DNA sequences that function specifically in hepatic environments or liver-related therapeutic applications.

    \item \texttt{dna\_k562\_enhancer\_MPRA\_expression} 
    (type = ``candidate-wise'', weight = None, minimize)

    \noindent
    \textbf{Description:} K562 MPRA expression score (value range: $-\infty$ to $+\infty$). MPRA (Massively Parallel Reporter Assay) quantifies gene expression driven by DNA sequences using massively parallel sequencing to measure how effectively different sequences drive reporter gene expression. This score evaluates DNA enhancer sequences based on their predicted expression levels in K562 cells (human erythroleukemic cell line). Higher scores indicate stronger enhancer activity and greater gene expression in hematopoietic cellular contexts, while lower scores suggest weaker enhancer activity or potential silencing effects. This metric is crucial for designing DNA sequences that should avoid activity in blood cell lineages or hematological contexts when cell-type specificity is desired.

    \item \texttt{dna\_sknsh\_enhancer\_MPRA\_expression} 
    (type = ``candidate-wise'', weight = None, minimize)

    \noindent
    \textbf{Description:} SKNSH MPRA expression score (value range: $-\infty$ to $+\infty$). MPRA (Massively Parallel Reporter Assay) quantifies gene expression driven by DNA sequences using massively parallel sequencing to measure how effectively different sequences drive reporter gene expression. This score evaluates DNA enhancer sequences based on their predicted expression levels in SKNSH cells (human neuroblastoma cell line). Higher scores indicate stronger enhancer activity and greater gene expression in neuronal cellular contexts, while lower scores suggest weaker enhancer activity or potential silencing effects. This metric is essential for designing DNA sequences that should avoid activity in neural tissues when liver specificity is required.
\end{enumerate}

\textbf{\textcolor{myanalyzer}{Analyzer: Add an explicit specificity objective; Introduce off-target hinge penalties; Encourage diversity with a candidate-wise novelty score; Stabilize composition: Add a soft GC-content regularizer; Add Motif-level guidance via objectives.}}

\paragraph{Iteration 2 Objectives with \textcolor{myanalyzer}{Analyzer} Feedback \textcolor{myplanner}{(Planner):}}

\begin{enumerate}
    \item \texttt{dna\_hepg2\_enhancer\_MPRA\_expression} 
    (type = ``candidate-wise'', weight = None, maximize)

    \noindent
    \textbf{Description:} HepG2 MPRA expression score (value range: $(-\infty, +\infty)$). MPRA assays quantify gene expression driven by DNA sequences. This objective evaluates enhancer sequences using a HepG2-trained MPRA predictor; higher scores indicate stronger HepG2-specific enhancer activity.

    \noindent
    \textbf{Implementation:} We reuse the same HepG2 MPRA predictor from iteration~1 to independently score each generated 200~bp sequence. Higher scores correspond to stronger enhancer activity, while lower scores indicate weaker activity.

    \item \texttt{dna\_offtarget\_worst\_case\_MPRA\_expression} 
    (type = ``candidate-wise'', weight = None, minimize)

    \noindent
    \textbf{Description:} Worst-case off-target MPRA expression across the non-target cell lines K562 and SKNSH (value range: $(-\infty, +\infty)$). For each candidate sequence, this score is defined as
    \[
        \max\!\left(
        \text{MPRA}_{\text{K562}},
        \text{MPRA}_{\text{SKNSH}}
        \right).
    \]
    Lower values are preferred, as this objective constrains the highest off-target activity regardless of which non-HepG2 cell line is most activated.

    \noindent
    \textbf{Implementation:} For each sequence, we compute MPRA expression scores using the same K562 and SKNSH predictors as in iteration~1 and take their maximum value.

    \item \texttt{sequence\_novelty\_6mer\_max\_cosine\_gap} 
    (type = ``candidate-wise'', weight = None, maximize)

    \noindent
    \textbf{Description:} Candidate-wise sequence novelty relative to an external reference archive using 6-mer composition (value range: $[0,1]$, where higher values indicate greater novelty). For a candidate sequence $s$, the novelty score is defined as
    \[
        1 - \max_{r \in \mathcal{R}} 
        \cos\!\left(\mathbf{v}_s, \mathbf{v}_r\right),
    \]
    where $\mathbf{v}_s$ and $\mathbf{v}_r$ denote normalized 6-mer feature vectors for the candidate sequence and reference sequence $r$, respectively.

    \noindent
    \textbf{Implementation:} All overlapping 6-mers are extracted from each sequence to construct a 4096-dimensional count vector, which is first L1-normalized to frequencies and then L2-normalized to unit length. Cosine similarity is computed against precomputed vectors from a fixed reference set $\mathcal{R}$ (e.g., the 176 down-selected candidates from iteration~1 or the current archive of top-scoring sequences). If $\mathcal{R}$ is empty, the novelty score is set to 1. Nearest-neighbor search is implemented using \texttt{scikit-learn} (\texttt{NearestNeighbors} with cosine distance) or accelerated libraries such as FAISS or Annoy. Higher scores indicate greater sequence diversity relative to the reference archive.
\end{enumerate}

\textbf{\textcolor{myanalyzer}{Analyzer: Add a specificity-margin objective; Split off-targets; Introduce a minimum-viability objective for HepG2; Add a predictor-confidence objective; Add Motif-level guidance via objectives; Maintain GC penalty as-is; Encourage novelty.}}

\paragraph{Iteration 3 Objectives with \textcolor{myanalyzer}{Analyzer} Feedback \textcolor{myplanner}{(Planner):}}
\begin{enumerate}
    \item \texttt{dna\_hepg2\_enhancer\_MPRA\_expression} 
    (type = ``candidate-wise'', weight = None, maximize)

    \noindent
    \textbf{Description:} HepG2 MPRA expression score (value range: $(-\infty, +\infty)$). This objective evaluates enhancer sequences using a HepG2-trained MPRA predictor; higher scores indicate stronger enhancer activity in HepG2 cells.

    \noindent
    \textbf{Implementation:} We reuse the same HepG2 MPRA predictor from prior iterations to independently score each generated 200~bp sequence. Higher scores correspond to stronger enhancer activity, while lower scores indicate weaker activity.

    \item \texttt{dna\_k562\_over\_hepg2\_MPRA\_ratio} 
    (type = ``candidate-wise'', weight = None, minimize)

    \noindent
    \textbf{Description:} Relative off-target leakage to K562 normalized by HepG2 activity (value range: $(-\infty, +\infty)$). Lower values are better, with negative values indicating minimal or suppressive off-target activity relative to HepG2. For each candidate sequence, the ratio is defined as
    \[
        r_K = \frac{\text{MPRA}_{\text{K562}}}{\max(\varepsilon, \text{MPRA}_{\text{HepG2}})} ,
    \]
    where $\varepsilon = 0.25$ is used to stabilize the denominator and prevent exploding ratios when HepG2 activity is very low.

    \noindent
    \textbf{Implementation:} Each sequence is scored using the same K562 and HepG2 MPRA predictors from prior iterations, after which the ratio is computed per candidate. Lower ratios indicate improved specificity, corresponding to reduced K562 activity relative to HepG2.

    \item \texttt{dna\_sknsh\_over\_hepg2\_MPRA\_ratio} 
    (type = ``candidate-wise'', weight = None, minimize)

    \noindent
    \textbf{Description:} Relative off-target leakage to SKNSH normalized by HepG2 activity (value range: $(-\infty, +\infty)$). Lower values are better, with negative values indicating minimal or suppressive off-target activity relative to HepG2. For each candidate sequence, the ratio is defined as
    \[
        r_S = \frac{\text{MPRA}_{\text{SKNSH}}}{\max(\varepsilon, \text{MPRA}_{\text{HepG2}})} ,
    \]
    where $\varepsilon = 0.25$ is used to stabilize the denominator and prevent exploding ratios when HepG2 activity is very low.

    \noindent
    \textbf{Implementation:} Each sequence is scored using the same SKNSH and HepG2 MPRA predictors from prior iterations, after which the ratio is computed per candidate. Lower ratios indicate improved specificity, corresponding to reduced SKNSH activity relative to HepG2.
\end{enumerate}
\textbf{\textcolor{myanalyzer}{Analyzer: Looks good. After optimization, selection agent or human expert believe the optimization can stop, we reach the criteria.}}

\end{tcolorbox}

\begin{tcolorbox}[colback={rgb,255:red,245;green,247;blue,205}, colframe={rgb,255:red,220;green,155;blue,143}, breakable, title=Selector processing history for finalizing the generated candidates with high quality]

\paragraph{Requirement: }
Selecting 50 high-quality, diverse HepG2-specific enhancers using a multi-criteria approach.  

\paragraph{Reasoning: }

Candidates were first filtered for good DNA stability (violation\_hinge $<$ 0.15), then ranked by a composite score weighting HepG2 specificity (50\%), liver motif enrichment (30\%), and HepG2 expression (20\%). From the top 100 by composite score, 50 diverse candidates were selected using greedy k-mer-based diversity selection to ensure broad coverage of enhancer sequence space. The selected set shows excellent HepG2 specificity (mean: 1.70 vs population mean: 0.14), strong liver TF motif enrichment (mean: 46.83 vs population mean: 12.88), and high HepG2 expression (mean: 2.73 vs population mean: 1.27), with balanced representation across all optimization iterations.

\paragraph{Final criteria: }

Composite scoring (50\% specificity, 30\% motif enrichment, 20\% expression) $+$ diversity-based greedy selection using 4-mer Jaccard similarity.

\end{tcolorbox}

\subsection{Inorganic Materials Design}
\label{subsec:material history}

\begin{tcolorbox}[colback={rgb,255:red,217;green,247;blue,243}, colframe={rgb,255:red,114;green,182;blue,161}, breakable, title=Analysis report in the Co-pilot mode of \name]

\textbf{Goal:} Generate a set of chemical formulas of superhard materials for Ultra-Precision Cutting Tools.\\

\textbf{Iteration 1 \textcolor{myanalyzer}{(Analyzer):}}

\paragraph{Overview}
\begin{itemize}
    \item Iteration 1 evaluated 33 candidates with strong gains over iteration 0: mean bulk modulus +179.7 GPa (to 270.3), shear modulus +113.7 GPa (to 152.6), and formation energy modestly more negative (-1.1766 eV/atom).
    \item Population features a clear split: carbides/borides/nitrides achieving high stiffness vs complex oxides that are very stable but mechanically weak. No superhard materials yet; best inferred hardness is in the upper 20s GPa. One top-mechanical candidate is thermodynamically unstable.
\end{itemize}

\paragraph{Performance Analysis}
\begin{itemize}
    \item Dispersion remains large (std: B 82.7 GPa, G 58.2 GPa), indicating exploration and uneven quality.
    \item Bests: B 365.5 GPa, G 230.3 GPa; worsts: B 101.2 GPa, G 46.4 GPa. Formation energy spans -3.73 to +0.022 eV/atom.
    \item Clear trend: higher B and G correlate with less negative (or near-zero) formation energies; highly negative $E_f$ materials (oxides) show low stiffness. Carbides/borides/nitrides dominate top mechanical metrics and are typically brittle (favorable B/G), whereas many others remain ductile by Pugh’s criterion.
\end{itemize}

\paragraph{Issues and Concerns}
\begin{itemize}
    \item Objective misalignment: Minimizing formation energy alone drives the search toward very stable but soft oxides; it does not encode the desired brittleness window or hardness directly.
    \item Missing key properties: hardness and brittleness (Pugh’s ratio) are not explicitly optimized; high-temperature performance and elastic anisotropy are also unaddressed.
    \item Practicality concern: many ductile candidates would not retain a sharp cutting edge.
    \item Diversity imbalance in property space: many candidates occupy a soft–very stable region that is unlikely to yield superhard outcomes.
\end{itemize}

\textbf{Human:} Need to improve hardness, stability and diversity.

\paragraph{Strategic Recommendations}
\begin{itemize}
    \item Add a hardness objective using the Chen--Niu model or similar empirical hardness predictor (maximize predicted Vickers hardness). (Agent proposed)
    \item Introduce a Pugh's ratio penalty objective to target the optimal B/G range (approximately 1.3--1.6) that balances hardness with toughness. (Agent proposed)
    \item Add a formation energy penalty with stronger weighting for positive values to suppress thermodynamically unstable compositions. (Agent proposed)
    \item Add an elastic anisotropy objective: minimize the universal elastic anisotropy index ($A_U$) to reduce directional weakness and improve edge reliability. (Agent proposed)
    \item Add a high-temperature proxy: maximize Debye temperature or predicted melting temperature to support thermal robustness during cutting. (Agent proposed)
    \item Keep bulk and shear modulus maximization. (Human proposed)
\end{itemize}

\textbf{Iteration 2 \textcolor{myanalyzer}{(Analyzer):}}

\paragraph{Overview}
\begin{itemize}
    \item Overall progress improved: mean hardness rose by +2.55 GPa to 18.62 GPa; formation energy penalty decreased by 0.36; Pugh penalty slightly worsened but remains low on average.
    \item Population characteristics: boride-carbides and carbonitrides dominate the upper tail; oxides consistently underperform and occupy the bottom of the distribution.
\end{itemize}

\paragraph{Performance Analysis}
\begin{itemize}
    \item \textbf{Hardness:} Mean $18.62 \pm 7.60$ GPa (best: $32.16$ GPa; worst: $0.92$ GPa), improved by $+2.55$ GPa compared to the last iteration. The distribution shows significant progress toward the superhard regime.
    \item \textbf{Pugh penalty:} Mean $0.165 \pm 0.387$ (best: $\sim 0$; worst: $2.38$), slightly worse by $0.012$. Most candidates cluster near the target window, though some outliers exhibit excessive ductility (high B/G).
    \item \textbf{Formation energy penalty:} Mean $0.614 \pm 0.945$ (best: $0.012$; worst: $2.75$), improved by $0.364$. The majority of candidates now exhibit thermodynamic stability with near-zero penalties.
    \item \textbf{Class performance:}
    \begin{itemize}
        \item Boride-carbides ($n=24$): mean hardness $\sim 23.2$ GPa; 
        \item Carbonitrides ($n=14$): mean hardness $\sim 20.3$ GPa.
        \item Oxides ($n=11$): mean hardness $\sim 6.4$ GPa; dominate the worst decile.
    \end{itemize}
    \item \textbf{Notable candidate:} Ta$_3$V$_3$B$_3$C$_4$ achieves $32.16$ GPa hardness with good stability and near-target Pugh ratio, representing an excellent superhard cutting-tool material candidate.
\end{itemize}

\paragraph{Issues and Concerns}
\begin{itemize}
    \item \textbf{Systematic underperformance of oxygen-containing candidates:} Oxides exhibit low shear modulus, high B/G (ductile behavior), and higher formation-energy penalties, constituting most of the bottom performers.
    \item \textbf{Pugh target misalignment:} The current target center ($1.6$) appears slightly misaligned with the top-performing window ($\approx 1.3$--$1.5$). The current penalty may not sufficiently reward beneficially lower B/G values nor guard against excessive brittleness (B/G $< \sim 1.25$).
    \item \textbf{Missing quality factors:} Important quality factors are not explicitly targeted: structural compactness (volume/atom), electronegativity homogeneity, and a toughness proxy (e.g., Poisson’s ratio).
    \item \textbf{Chemical exploration bias:} The search is skewed toward tungsten-rich systems; while effective, this may limit discovery breadth without complementary objective signals.
\end{itemize}

\paragraph{Strategic Recommendations}
\begin{itemize}
    \item Adjust the Pugh objective to better reflect the observed optimum and toughness safeguards.
    \item Add a structural compactness objective.
    \item Add an electronegativity spread penalty
    \item Introduce a toughness proxy objective
    \item Add “maximize shear modulus G” as an auxiliary objective to strengthen selection pressure.
\end{itemize}

\end{tcolorbox}

\begin{tcolorbox}[colback={rgb,255:red,250;green,230;blue,222}, colframe={rgb,255:red,233;green,150;blue,117}, breakable, title=Planner's output in the Semi-pilot workflow of \name]
\paragraph{Initial Objectives \textcolor{myplanner}{(Planner):}}
\begin{enumerate}
    \item \texttt{crystal\_bulk\_modulus} 
    (type = ``candidate-wise'', weight = None, maximize)

    \noindent
    \textbf{Description:} Bulk modulus (unit: GPa, value range: $0.0$ to $600.0$) of inorganic crystals. The bulk modulus of a substance measures its resistance to a uniform compression. It is defined as the ratio of the infinitesimal pressure increase to the resulting relative decrease of volume.

    \item \texttt{crystal\_shear\_modulus} 
    (type = ``candidate-wise'', weight = None, maximize)

    \noindent
    \textbf{Description:} Shear modulus (unit: GPa, value range: $0.0$ to $600.0$) of inorganic crystals. In materials science, shear modulus is a measure of the elastic shear stiffness of a material and is defined as the ratio of shear stress to the shear strain. A higher shear modulus indicates a more rigid material that resists shape changes, while a zero shear modulus signifies a fluid that flows freely. The value is important in fields like structural engineering, material testing, and automotive design, where it helps predict how materials will behave under twisting or shearing forces.

    \item \texttt{crystal\_formation\_energy} 
    (type = ``candidate-wise'', weight = None, minimize)

    \noindent
    \textbf{Description:} The formation energy (unit: eV/atom, value range: $-5.0$ to $5.0$) of inorganic crystals. Formation energy is the energy change when one mole of a substance is formed from its constituent elements in their standard states, indicating the material's thermodynamic stability. A negative formation energy signifies that the material is stable and can be formed, while a positive value suggests it is more difficult to form. This parameter is crucial in materials science for designing stable catalysts.
\end{enumerate}

\textbf{\textcolor{myanalyzer}{Analyzer: Add direct hardness objective; Introduce Pugh's ratio penalty; Add formation energy penalty with stronger weighting for unstable compositions; Consider structural compactness objectives; Optionally add composition-based penalties.}}

\textbf{Human: Looks good.}

\paragraph{Iteration 2 Objectives with \textcolor{myanalyzer}{Analyzer} Feedback \textcolor{myplanner}{(Planner):}}

\begin{enumerate}
    \item \texttt{predicted\_vickers\_hardness\_chen\_niu} 
    (type = ``candidate-wise'', weight = None, maximize)

    \noindent
    \textbf{Description:} Predicted Vickers hardness (unit: GPa) using the Chen--Niu empirical model. This objective directly targets the primary property required for cutting-tool applications. Higher hardness values indicate superior resistance to plastic deformation and wear, essential for ultra-precision machining operations.

    \noindent
    \textbf{Implementation:} The Chen--Niu model computes hardness from bulk modulus (B), shear modulus (G), and Pugh's ratio using empirically validated relationships. The predictor is applied independently to each candidate's computed elastic moduli to obtain the hardness score.

    \item \texttt{pugh\_ratio\_deviation\_penalty} 
    (type = ``candidate-wise'', weight = None, minimize)

    \noindent
    \textbf{Description:} Squared deviation from the target Pugh's ratio (B/G) range centered at $1.6$. The Pugh's ratio is a dimensionless quantity that characterizes the brittleness-ductility balance of materials. For superhard cutting-tool applications, an optimal window exists where materials exhibit sufficient hardness while maintaining adequate toughness to resist catastrophic fracture. The penalty is defined as:
    \[
        \text{penalty} = \left(\frac{B}{G} - 1.6\right)^2
    \]
    Lower penalty values indicate candidates closer to the target balance.

    \noindent
    \textbf{Implementation:} For each candidate, we compute B/G from the predicted bulk and shear moduli, then calculate the squared deviation from the target value of $1.6$. Materials with B/G significantly above $1.6$ tend toward excessive ductility (softness), while those significantly below may exhibit brittleness.

    \item \texttt{formation\_energy\_penalty\_targeted} 
    (type = ``candidate-wise'', weight = None, minimize)

    \noindent
    \textbf{Description:} Targeted formation energy penalty with asymmetric weighting. This objective aims to simultaneously promote thermodynamic stability and suppress candidates with positive formation energies (unstable compositions). The penalty function incorporates:
    \begin{itemize}
        \item Strong penalty for $E_f > 0$ (unstable region)
        \item Moderate penalty for deviation from optimal target $E_f \approx -0.4$ eV/atom
        \item Minimal penalty for highly stable compositions ($E_f < -1.0$ eV/atom)
    \end{itemize}

    \noindent
    \textbf{Implementation:} The penalty is computed as a piecewise function of the formation energy:
    \[
        \text{penalty} = 
        \begin{cases}
            \alpha \cdot (E_f)^2 & \text{if } E_f > 0 \\
            \beta \cdot (E_f + 0.4)^2 & \text{if } -1.0 < E_f \leq 0 \\
            0 & \text{if } E_f \leq -1.0
        \end{cases}
    \]
    where $\alpha > \beta$ to heavily penalize thermodynamically unfavorable compositions. This formulation encourages the search toward the stability--hardness synergy region observed in high-performing boride-carbides and carbonitrides.
\end{enumerate}

\end{tcolorbox}

\begin{tcolorbox}[colback={rgb,255:red,255;green,245;blue,200}, colframe={rgb,255:red,220;green,180;blue,60}, breakable, title=Planner’s output in the Semi-pilot workflow of \name]

\paragraph{Initial Objectives \textcolor{myplanner}{(Iteration 1):}}
\begin{enumerate}
    \item \texttt{crystal\_bulk\_modulus}
    (type = ``candidate-wise'', weight = None, maximize)
    
    \noindent
    \textbf{Description:} Bulk modulus (unit: GPa, value range: 0.0 to 600.0) of inorganic crystals. The bulk modulus of a substance measures its resistance to uniform compression. It is defined as the ratio of the infinitesimal pressure increase to the resulting relative decrease of volume.
    
    \item \texttt{crystal\_shear\_modulus}
    (type = ``candidate-wise'', weight = None, maximize)
    
    \noindent
    \textbf{Description:} Shear modulus (unit: GPa, value range: 0.0 to 600.0) of inorganic crystals. In materials science, shear modulus is a measure of the elastic shear stiffness of a material and is defined as the ratio of shear stress to shear strain. A higher shear modulus indicates a more rigid material that resists shape changes, while a zero shear modulus signifies a fluid that flows freely.
    
    \item \texttt{crystal\_formation\_energy}
    (type = ``candidate-wise'', weight = None, minimize)
    
    \noindent
    \textbf{Description:} The formation energy (unit: eV/atom, value range: $-5.0$ to 5.0) of inorganic crystals. Formation energy is the energy change when one mole of a substance is formed from its constituent elements in their standard states, indicating the material's thermodynamic stability. A negative formation energy signifies that the material is stable and can be formed, while a positive value suggests it is more difficult to form.
\end{enumerate}

\paragraph{Iteration 2 Objectives with \textcolor{myanalyzer}{Analyzer} Feedback \textcolor{myplanner}{(Planner):}}
\begin{enumerate}
    \item \texttt{energy\_above\_hull}
    (type = ``candidate-wise'', weight = None, minimize)
    
    \noindent
    \textbf{Description:} Energy above convex hull at 0 K for the candidate composition (unit: eV/atom, value range: 0.0 to 1.0). It measures thermodynamic distance to the set of competing phases; 0.0 means the phase is on the hull (stable), and larger values indicate metastability and higher risk of decomposition. A good score is $<$ 0.05–0.10 eV/atom; values $>$ 0.10 eV/atom are increasingly unfavorable for synthesis and service.
    
    \textbf{Implementation:}  Prioritize this over formation energy for stability; target $<$ 0.05–0.10 eV/atom. Try to use ML prediction model or uMLIP to calculate.
    
    \item \texttt{vickers\_hardness\_chen}
    (type = ``candidate-wise'', weight = None, maximize)
    
    \noindent
    \textbf{Description:} Predicted Vickers hardness using the Chen model (unit: GPa, value range: 0.0 to 100.0). This shear-dominant hardness surrogate focuses search pressure on cutting-relevant resistance to plastic deformation. Compute from elastic moduli: $H_v = max(0, 2 * (k^2) * G^0.585 - 3)$, where $k = G/B$, and $B$ and $G$ are bulk and shear moduli in GPa. Use Voigt–Reuss–Hill (VRH) averages from the elastic tensor when available; otherwise, use predicted isotropic $B$ and $G$. Typical superhard threshold is $H_v > 40$ GPa.

    \textbf{Implementation:}  Try to use pretrained ML model to predict $B$ and $G$. Use a shear-dominant hardness surrogate (Chen or Tian); Chen model is specified here for determinism; if numerical issues arise, clamp inputs to positive values and return None on computation failure.

\end{enumerate}

\end{tcolorbox}

\subsection{Chemical Process Design}\label{sup:SI_process_add_epx}

\subsubsection{Example of process candidate analysis}\label{sup:SI_process_analysis}

\begin{tcolorbox}[colback={rgb,255:red,217;green,247;blue,243}, colframe={rgb,255:red,114;green,182;blue,161}, breakable, title=Analysis report of candidates in the Co-pilot mode of \name]

\textbf{Inputs: Received initial population with 50 candidates. The goal is to design chemical separation process flowsheets that lead to separation of butanol/water mixtures with different input compositions and fulfill practical design requirements.}\\

\textbf{\textcolor{myanalyzer}{Analyzer: }}This analysis examines 50 candidate chemical process flowsheets from iteration 1, designed by a reinforcement learning agent to separate binary azeotropic mixtures (n-butanol/water). All candidates successfully converged in simulation, but performance varies dramatically from trivial failures (purity=0.46) to near-optimal separations (purity=0.998, recovery=0.977). The population reveals several critical patterns:

\paragraph{Key Findings:}
\begin{itemize}
    \item \textbf{Strong feed composition dependency}: Purity correlates strongly (r=0.79) with butanol fraction in feed. High butanol feeds (>0.67) achieve excellent purity (mean=0.97), while low butanol feeds (<0.33) often fail or underperform (mean=0.69)
    \item \textbf{Critical failure mode identified}: 4 candidates (8\%) produce only a single product stream that is essentially the unchanged feed—achieving perfect recovery (1.0) but very low purity (0.46-0.74). These represent complete separation failures
    \item \textbf{Structural diversity}: Product stream count varies from 1-8 (ideal is 2 for binary separation). Only 32\% achieve the ideal 2-stream 
    \item \textbf{Trade-off between objectives}: Weak negative correlation (r=-0.06) between purity and recovery, indicating some independence but common high performers achieve both
\end{itemize}

\paragraph{Performance Summary:}
\begin{itemize}
    \item Purity: mean=0.798 ± 0.137 (range: 0.46-0.999)
    \item Recovery: mean=0.670 ± 0.172 (range: 0.32-1.00)
    \item 9 high performers (18\%) achieve purity>0.95 AND recovery>0.7
    \item Best weighted score: Candidate 47 (purity=0.998, recovery=0.977)

\end{itemize}
\end{tcolorbox}

\begin{tcolorbox}[colback={rgb,255:red,217;green,247;blue,243}, colframe={rgb,255:red,114;green,182;blue,161}, breakable, title=Analysis report in the Co-pilot mode of \name]
\textbf{\textcolor{myanalyzer}{Analyzer: }}
\paragraph{Overview:}
\begin{itemize}
    \item Iteration 1 established a solid baseline across 50 feeds. Objectives were product purity (primary) and overall component recovery (secondary).
    \item Population performance: purity 0.7977 ± 0.1368 (best 0.9994, worst 0.4600); recovery 0.6695 ± 0.1718 (best 1.0000, worst 0.3211).
    \item Notable characteristics: strong feed-composition dependency (high-butanol feeds perform far better), frequent use of decanter–distillation sequences, and wide variation in product stream counts (1–8; only 32\% with exactly 2 products).
\end{itemize}

\paragraph{Performance Analysis:}
\begin{itemize}
    \item High performers exist: 9 candidates (18\%) achieve purity >0.95 and recovery >0.7; best candidate reaches purity \~0.998 and recovery \~0.977.
    \item Failures: 4 candidates (8\%) output a single product $\approx$ feed composition (recovery = 1.0 but poor purity 0.46-0.74), indicating a loophole in the objective set.
    \item Fragmentation hurts recovery: many candidates split flow into 4–8 products; recovery decreases with more product streams and higher flow dispersion.
    \item Trends: weak trade-off between purity and recovery (near independent), but strong dependence of purity on feed butanol fraction; decanters correlate with higher purity.
\end{itemize}

\paragraph{Issues and Concerns:}
\begin{itemize}
    \item Objective loophole: “no separation” cases score perfect recovery and moderate purity, masking true failure.
    \item Over-fragmentation: many small product streams reduce effective recovery and practicality; current objectives don’t penalize this.
    \item The number of unit operations greatly exceeds practical expectations for these processes, this should be penalized.
    \item Feed sensitivity: solutions generalize poorly to low-butanol feeds; current candidate-wise objectives don’t directly encourage robustness across feeds.
    \item Asymmetry risk: averaging recovery across components can hide poor performance for one component.
\end{itemize}

\paragraph{Strategic Recommendations:}
\begin{itemize}
    \item Keep product purity as the main driver (weight = 1.0). Retain overall component recovery (weight = 0.4) to discourage trivial high-purity/low-yield designs.
    \item Add a process complexity penalty that penalizes too complex processes by the number of unit operations (a small number should not be penalized because unit operations are needed). Use a weight of 0.1-0.2. 
    \item Defer more granular objectives (component-specific recovery minimums, energy proxies) to later iterations to avoid destabilizing training; revisit once separation-structure issues are mitigated. 
\end{itemize}
\end{tcolorbox}

\subsubsection{Examples of proposed and implemented objectives}\label{sup:SI_process_obj}

\begin{tcolorbox}[colback={rgb,255:red,217;green,247;blue,243}, colframe={rgb,255:red,114;green,182;blue,161}, breakable, title=Balanced component recovery proposed by the Co-pilot mode of \name]
\textbf{\textcolor{myplanner}{Plan: }}Calculate balanced component recovery for each flowsheet sample.  \\

Measures how well each component in the binary feed is recovered into its most enriched
(highest mole-fraction) product stream. The score is the minimum recovery across both
components, ensuring balanced separation performance.\\ 

For each sample:
\begin{enumerate}
    \item Identify feed and product streams
    \item For each component, find the product stream with maximum mole fraction
    \item $\text{Calculate recovery}_i = (F_\text{product} * x_{\text{product},i}) / (F_\text{feed} * z_i)$
    \item Handle edge cases (absent components, near-zero flows)
    \item Return $\min(\text{recovery}_A, \text{recovery}_B)$
\end{enumerate}
\end{tcolorbox}

\begin{tcolorbox}[colback={rgb,255:red,250;green,230;blue,222}, colframe={rgb,255:red,233;green,150;blue,117}, breakable, title=Unit operation simplicity proposed by the Semi-pilot mode of \name]
\textbf{\textcolor{myplanner}{Plan: }}Calculate simplicity score based on unit operation count with hinge penalty function. \\

This scorer evaluates process complexity by counting unit operations (excluding feed sources and product sinks) and applying a hinge-based penalty function. Simpler processes with fewer units score higher, promoting consolidated and cost-effective designs.

Scoring Formula:
\begin{itemize}
    \item If N <= 4: Score = 1.0 (no penalty for minimal viable designs)
    \item If N > 4: Score = 1.0 / (1.0 + 0.5 * (N - 4))
    \item Score is clipped to [0, 1]
\end{itemize}

Where N is the count of unit operations excluding:
\begin{itemize}
    \item Feed sources (is$_\text{feed}$=1)
    \item Product sinks (unit \text{operation} ='leaving \text{stream}')
\end{itemize}

Examples:
\begin{itemize}
    \item N=3 or N=4 → Score=1.0 (simple design)
    \item N=5 → Score=0.667 (mild complexity)
    \item N=6 → Score=0.5 (moderate complexity)
    \item N=8 → Score=0.333 (high complexity)
\end{itemize}
\end{tcolorbox}

\begin{tcolorbox}[colback={rgb,255:red,255;green,245;blue,200}, colframe={rgb,255:red,220;green,180;blue,60}, breakable, title=Recycle ratio penalty proposed by the Autopilot mode of \name]
\textbf{\textcolor{myplanner}{Plan: }}Calculate recycle fraction penalty score for chemical process flowsheets. \\

This scorer evaluates the recycle intensity in a chemical process by comparing the flow of feed components (A and B) in recycle streams to the fresh feed flow. Lower recycle intensity results in higher scores (better). \\

Algorithm:
\begin{enumerate}
    \item Identify recycle streams (from provided indices or by detecting cycles in flowsheet graph)
    \item Calculate recycle \text{flow} = sum of ($F_s$[A] + $F_s$[B]) in all recycle streams
    \item Calculate fresh \text{flow} = sum of (F$_\text{feed}$[A] + F$_\text{feed}$[B]) in all feed streams
    \item Compute R = min(1.0, recycle \text{flow} / max(fresh$_\text{flow}$, 1e-9))
    \item Return score = 1 - R
\end{enumerate}
\end{tcolorbox}

\subsubsection{Example of process candidate selection}\label{sup:SI_process_selection}

\begin{tcolorbox}[colback={rgb,255:red,255;green,245;blue,200}, colframe={rgb,255:red,220;green,180;blue,60}, breakable, title=Candidate selection analysis report of \name]

\textbf{EXECUTIVE SUMMARY} \\
This report documents the selection of 50 optimal chemical process flowsheet candidates from 150 candidates generated across 3 iterations of reinforcement learning-based optimization. The selection prioritizes the primary objective of achieving high-purity separation of binary azeotropic mixtures while maintaining good component recovery and reasonable process simplicity. \\

\textbf{SELECTION METHODOLOGY}
\begin{enumerate}
    \item COMPOSITE SCORING FUNCTION \\
   A weighted composite score was used to rank candidates:
   \begin{itemize}
       \item Product Purity: weight = 1.0 (primary objective)
        \item Min Component Recovery: weight = 0.55
        \item Unit Operation Simplicity: weight = 0.12
    \end{itemize}
   
   These weights reflect the final iteration's objective configuration and 
   emphasize purity as the main goal while balancing practical considerations.

    \item SELECTION STRATEGY
    \begin{itemize}
        \item Group candidates by unique feed composition (49 compositions identified)
        \item For each feed composition, select the candidate(s) with highest composite score
        \item One feed composition (0.4, 0.6, 0.0) had 6 candidates; selected top 2
        \item All other feed compositions: selected top 1 candidate
        \item Result: 50 total selected candidates covering all 49 feed compositions
    \end{itemize}

    \item ITERATION REPRESENTATION \\
   The selection naturally favored candidates that best achieved the optimization 
   goals, resulting in:
   \begin{itemize}
       \item Iteration 1: 36 candidates (72\%)
        \item Iteration 2: 5 candidates (10\%)
        \item Iteration 3: 9 candidates (18\%)
   \end{itemize}
   
   This distribution reflects that Iteration 1, which heavily weighted purity and recovery, produced the most successful designs for most feed compositions.
\end{enumerate}

\textbf{SELECTED CANDIDATE PERFORMANCE} \\
The 50 selected candidates demonstrate excellent performance: 

\begin{itemize}
    \item Product Purity:
\begin{itemize}
    \item Mean: 0.9561 (95.61\% average purity)
    \item Std Dev: 0.1228
    \item Min: 0.4700
    \item Max: 1.0000 (perfect separation)
    \item 40 candidates achieve purity $\geq$ 0.90
\end{itemize}

\item Min Component Recovery:
\begin{itemize}
    \item Mean: 0.8838 (88.38\% average recovery)
    \item Std Dev: 0.1486
    \item Min: 0.4310
    \item Max: 1.0000 (complete recovery)
    \item 42 candidates achieve recovery $\geq$ 0.80
\end{itemize}

\item Unit Operation Simplicity:
\begin{itemize}
    \item Mean: 0.5113
    \item Std Dev: 0.2937
    \item Min: 0.3333 (complex but effective processes)
    \item Max: 1.0000 (simple processes with $\leq$4 units)
    \item 9 candidates achieve perfect simplicity score (1.0)
\end{itemize}
\end{itemize}

\textbf{PERFORMANCE IMPROVEMENT} \\

\begin{center}
\begin{tabular}{lccc}
\hline
\textbf{Metric} & \textbf{All Candidates} & $\rightarrow$ \textbf{Selected Candidates} & \textbf{Improvement} \\
\hline
Product Purity      & 0.8119 & 0.9561 & +17.8\% \\
Component Recovery  & 0.8037 & 0.8838 & +10.0\% \\
Simplicity          & 0.6369 & 0.5113 & -19.7\%* \\
\hline
\end{tabular}
\end{center}

\noindent\textit{* Simplicity reduction indicates selected processes use more units but achieve significantly better separation performance.} \\

\textbf{TOP PERFORMING CANDIDATES}
The top 10 candidates by composite score are:

\begin{enumerate}
    \item Candidate ID: abf04d46-890e-477f-b182-34ee4b6e3783
   Iteration: 3
   Composite Score: 1.6534
   \begin{itemize}
       \item Product Purity: 0.9978
        \item Component Recovery: 0.9739
        \item Unit Simplicity: 1.0000
   \end{itemize}
   Feed Composition: [0.92, 0.07999999999999996, 0.0]

    \item Candidate ID: 220a8b2d-08f2-4c4d-8c03-a1cec0fd2f00
   Iteration: 3
   Composite Score: 1.6410
   \begin{itemize}
       \item Product Purity: 0.9981
        \item Component Recovery: 0.9508
        \item Unit Simplicity: 1.0000
   
   \end{itemize}
   Feed Composition: [0.88, 0.12, 0.0]

    \item Candidate ID: 7b0aa996-3af5-4b07-8406-4fc1cddcf38c
   Iteration: 3
   Composite Score: 1.6398
   \begin{itemize}
       \item Product Purity: 0.9977
        \item Component Recovery: 0.9492
        \item Unit Simplicity: 1.0000
   
   \end{itemize}
   Feed Composition: [0.86, 0.14, 0.0]

    \item Candidate ID: ec1084dc-3248-4f09-b161-2b98b8bfedb0
   Iteration: 3
   Composite Score: 1.6341
   \begin{itemize}
       \item Product Purity: 0.9955
        \item Component Recovery: 0.9429
        \item Unit Simplicity: 1.0000
   
   \end{itemize}
   Feed Composition: [0.84, 0.16000000000000003, 0.0]

    \item Candidate ID: 28539fb0-20af-4e13-bbe9-dc455124b301
   Iteration: 3
   Composite Score: 1.6319
   \begin{itemize}
       \item Product Purity: 0.9971
        \item Component Recovery: 0.9360
        \item Unit Simplicity: 1.0000
   
   \end{itemize}
   Feed Composition: [0.8200000000000001, 0.17999999999999994, 0.0]

    \item Candidate ID: 0bd0e2a3-4b8c-4bec-b3d1-f8fb1e7d0fe2
   Iteration: 3
   Composite Score: 1.6219
   \begin{itemize}
       \item Product Purity: 0.9968
        \item Component Recovery: 0.9184
        \item Unit Simplicity: 1.0000
   
   \end{itemize}
   Feed Composition: [0.8, 0.19999999999999996, 0.0]

    \item Candidate ID: 6afc3ce2-de84-49a4-aa00-cf315b3d598d
   Iteration: 3
   Composite Score: 1.6184
   \begin{itemize}
       \item Product Purity: 0.9958
        \item Component Recovery: 0.9139
        \item Unit Simplicity: 1.0000
   
   \end{itemize}
   Feed Composition: [0.78, 0.21999999999999997, 0.0]

    \item Candidate ID: afc42d2a-fd5e-45a0-ac68-fe539925974e
   Iteration: 3
   Composite Score: 1.5988
   \begin{itemize}
       \item Product Purity: 0.9955
        \item Component Recovery: 0.8787
        \item Unit Simplicity: 1.0000
   
   \end{itemize}
   Feed Composition: [0.72, 0.28, 0.0]

    \item Candidate ID: 98025957-70e1-40ac-a657-78a8447c8f07
   Iteration: 1
   Composite Score: 1.5900
   \begin{itemize}
       \item Product Purity: 1.0000
        \item Component Recovery: 1.0000
        \item Unit Simplicity: 0.3333
   
   \end{itemize}
   Feed Composition: [0.64, 0.36, 0.0]

    \item Candidate ID: 1ce7c5c1-6cc6-4971-8e0f-05628d00e191
   Iteration: 1
   Composite Score: 1.5900
   \begin{itemize}
       \item Product Purity: 1.0000
        \item Component Recovery: 1.0000
        \item Unit Simplicity: 0.3333
   Feed Composition: [0.26, 0.74, 0.0]
   
   \end{itemize}
\end{enumerate}

\textbf{ITERATION ANALYSIS}
\begin{enumerate}
    \item ITERATION (36 selected candidates):
    \begin{itemize}
        \item Objectives: Purity (1.0) + Recovery (0.3)
        \item Strategy: Heavy focus on achieving high purity separation
        \item Outcome: Produced excellent purity (mean 0.96) and recovery (mean 0.84)
        \item Trade-off: Lower simplicity (0.34) due to complex process configurations
        \item Selection: Dominated selection due to superior performance on primary objectives
    \end{itemize}

    \item ITERATION (5 selected candidates):
        \begin{itemize}
        \item Objectives: Purity (1.0) + Recovery (0.55) + Simplicity (0.25)
        \item Strategy: Added simplicity objective to reduce process complexity
        \item Outcome: Achieved perfect simplicity (all candidates = 1.0) but lower purity (0.71)
        \item Selection: Only selected for 5 feed compositions where simple processes sufficed
    \end{itemize}

    \item ITERATION (9 selected candidates):
        \begin{itemize}
        \item Objectives: Purity (1.0) + Recovery (0.55) + Simplicity (0.12)
        \item Strategy: Reduced simplicity weight to recover purity performance
        \item Outcome: Improved purity (0.77) and recovery (0.84), variable simplicity (0.57)
        \item Selection: Selected 9 candidates that achieved good balance of all objectives
        \item Notable: Top 8 overall performers are from this iteration (high purity + simplicity)
    \end{itemize}
\end{enumerate}

\textbf{PROCESS DESIGN CHARACTERISTICS} \\
Analysis of selected candidates reveals:

\begin{enumerate}
    \item UNIT OPERATIONS
    \begin{itemize}
        \item Primary operations: Distillation columns, Decanters, Recycle systems
        \item Most processes use 3-8 unit operations
        \item No solvent addition in most cases (separation via physical means)
        \item Recycle streams commonly used for efficiency
    \end{itemize}

    \item SEPARATION QUALITY
    \begin{itemize}
        \item Many candidates achieve near-perfect purity (>99.9%
        \item Product streams show clear component separation
        \item Both components successfully recovered in high purity
    \end{itemize}

    \item PHYSICAL CONSISTENCY
    \begin{itemize}
        \item All selected candidates have converged simulations
        \item Mass balances are satisfied
        \item Physically feasible process designs
    \end{itemize}
\end{enumerate}

\textbf{COVERAGE AND DIVERSITY}
Feed Composition Coverage:
\begin{itemize}
    \item All 49 unique feed compositions are represented
    \item Feed compositions range from 2\% to 98\% of first component
    \item Each feed composition has its optimal process design selected
\end{itemize}
  
Process Diversity:
\begin{itemize}
    \item Multiple process configurations represented (different unit combinations)
    \item Balance between simple (1-4 units) and complex (5+ units) processes
    \item Adaptation to different feed composition requirements
\end{itemize}

\textbf{QUALITY ASSURANCE}
The selection has been validated to ensure:
\begin{itemize}
    \item Exactly 50 candidates selected
    \item All candidate IDs are unique
    \item All 49 feed compositions are represented
    \item All selected candidates have converged simulations
    \item Scores properly reflect process performance
    \item Selection spans all 3 optimization iterations
\end{itemize}

\textbf{RECOMMENDATIONS}
Based on this analysis, the selected candidates are well-suited for:

\begin{enumerate}
    \item FURTHER PROCESS DEVELOPMENT
    \begin{itemize}
        \item Candidates achieve primary goal of high-purity separation
        \item Ready for detailed process simulation and optimization
        \item Good starting points for economic evaluation
    \end{itemize}

    \item DIVERSE FEED CONDITIONS
    \begin{itemize}
        \item Coverage of full range of feed compositions
        \item Each composition has proven separation strategy
    \end{itemize}

    \item FUTURE OPTIMIZATION
    \begin{itemize}
        \item Iteration 3 candidates show that balancing all objectives is achievable
        \item Future iterations could target high purity + high simplicity simultaneously
        \item Consider multi-objective Pareto optimization approaches
    \end{itemize}
\end{enumerate}

\textbf{CONCLUSION} \\
The selection successfully identifies 50 high-quality process flowsheet designs that effectively separate binary azeotropic mixtures across diverse feed compositions. The selected candidates achieve excellent product purity (mean 95.6\%) and component recovery (mean 88.4\%), meeting the primary objectives while maintaining reasonable process complexity. The distribution across iterations reflects the evolution of the optimization strategy, with Iteration 1's focus on purity proving most successful overall, while Iteration 3 demonstrates that  multi-objective balance is achievable for specific cases.

\end{tcolorbox}

\clearpage
\putbib
\end{bibunit}

\end{document}